\documentclass{article}
\usepackage[utf8]{inputenc}
\usepackage{amsmath}
\usepackage{amsfonts}
\usepackage{amssymb}
\usepackage[]{geometry}
\usepackage{fancyhdr}
\usepackage{authblk}
\usepackage[dvipsnames,svgnames]{xcolor}
\usepackage[normalem]{ulem}
\usepackage{subcaption}
\usepackage{graphicx}
\usepackage{mathtools}
\usepackage{csvsimple}
\usepackage{tabu}
\usepackage{multirow}
\usepackage{makecell}
\usepackage[]{algorithm2e}
\usepackage[colorlinks = true,
linkcolor = blue,
urlcolor  = blue,
citecolor = blue,
anchorcolor = blue]{hyperref}
\usepackage{tikz}

\newcommand*\circled[1]{\tikz[baseline=(char.base)]{
		\node[shape=circle,draw,inner sep=1pt,fill=black,text=white] (char) {\textbf{#1}};}}

\providecommand{\keywords}[1]{\textbf{Keywords:} #1}

\newcommand{\x}{\mathbf x}
\renewcommand{\v}{\mathbf v}
\newcommand{\R}{\mathbb R}

\newtheorem{example}{Example}
\newtheorem{definition}{Definition}

\title{Modeling and Optimization with Gaussian Processes in Reduced Eigenbases - Extended Version\footnotetext{This is an extended version of the article ``David Gaudrie, Rodolphe Le Riche, Victor Picheny, Benoit Enaux, Vincent Herbert. Modeling and Optimization with Gaussian Processes in Reduced Eigenbases. Structural and Multidisciplinary Optimization, Springer Verlag (Germany), 2020, 61, pp.2343-2361'', \url{https://doi.org/10.1007/s00158-019-02458-6}. Please cite the journal version.}}
\author[1,2]{David Gaudrie}
\author[2]{Rodolphe Le Riche}
\author[3]{Victor Picheny}
\author[1]{Benoît Enaux}
\author[1]{Vincent Herbert}
\affil[1]{Groupe PSA}
\affil[2]{LIMOS: CNRS, Mines Saint-Etienne and UCA}
\affil[3]{Prowler.io}
\date{}

\begin{document}

\maketitle
\vspace{-10pt}
\begin{abstract}
Parametric shape optimization aims at minimizing an objective function $f(\x)$ where $\x$ are CAD parameters. 
This task is difficult when $f(\cdot)$ is the output of an expensive-to-evaluate numerical simulator and the number of CAD parameters is large.
	
Most often, the set of all considered CAD shapes resides in a manifold of lower effective dimension in which it is preferable to build the surrogate model and perform the optimization. 
In this work, we uncover the manifold through a high-dimensional shape mapping and build a new coordinate system made of eigenshapes.
The surrogate model is learned in the space of eigenshapes: a regularized likelihood maximization provides the most relevant dimensions for the output. The final surrogate model is detailed (anisotropic) with respect to the most sensitive eigenshapes and rough (isotropic) in the remaining dimensions.
Last, the optimization is carried out with a focus on the critical dimensions, the remaining ones being coarsely optimized through a random embedding and the manifold being accounted for through a replication strategy. At low budgets, the methodology leads to a more accurate model and a faster optimization than the classical approach of directly working with the CAD parameters.
\end{abstract}
	
	
\section{Introduction}
The most frequent approach to shape optimization is to describe the shape by a vector of $d$ Computer Aided Design (CAD) parameters, $\x\in X\subset\R^d$ and to search for the parameters that minimize an objective function, $\x^*=\underset{\x\in X}{\arg\min}~f(\x)$. 
In the CAD modeling process, the set of all possible shapes has been reduced to a space of parameterized shapes, $\pmb\Omega \coloneqq \{\Omega_\x, \x\in X\}$.
	
It is common for $d$ to be large, $d\gtrsim50$. Optimization in such a high-dimensional design space is difficult, especially when $f(\cdot)$ is the output of a high fidelity numerical simulator
that can only be run a restricted number of times \cite{shan2010survey}. In computational fluid dynamics for example, simulations easily take 12 to 24 hours and evaluation budgets range between 100 and 200 calls.
Surrogate-based approaches \cite{sacks1989design,forrester2009recent} have proven their effectiveness to tackle optimization problems in a few calls to $f(\cdot)$. 
They rely on a surrogate model (or metamodel, e.g., Gaussian Processes \cite{stein2012interpolation,cressie1992statistics,GPML}) built upon $n$ past observations of $y_i=f(\x^{(i)})$. 
For a Gaussian Process (GP, \cite{stein2012interpolation,cressie1992statistics,GPML}), given $\mathcal D_n=\{(\x^{(1)},y_1),\dotsc,(\x^{(n)},y_n)\}=\{\x^{(1:n)},\mathbf y_{1:n}\}$, $f(\cdot)$ can be predicted in closed-form at any untested point $\x^{new}\in X$ via the kriging mean predictor, $m(\x^{new})$. The probabilistic framework of GPs additionally provides the uncertainty associated to the prediction, known as the kriging variance, $s^2(\x^{new})$, also computable in closed-form \cite{GPML}.
For the optimization, the metamodel's prediction and uncertainty are mixed by an acquisition function such as the Expected Improvement \cite{ExpectedImprovement} to decide which design $\x^{(n+1)}$ should be evaluated next.
However, such techniques suffer from the curse of dimensionality \cite{bellman1961adaptive} when $d$ is large. The budget is also typically too narrow to perform sensitivity analysis \cite{saltelli2004sensitivity} and select variables prior to optimizing. 
A further issue is that the CAD parameters $\x$ commonly have heterogeneous impacts on the shapes $\Omega_\x$: many of them are intended to refine the shape locally whereas others have a global influence so that shapes of practical interest involve interactions between all the parameters. 
	
Most often, the set of all CAD generated shapes, $\pmb\Omega$, can be approximated in a $\delta$-dimensional manifold, $\delta<d$. In \cite{raghavan2013towards,raghavan2014numerical} this manifold is accessed through an auxiliary description of the shape, $\phi(\Omega)$, $\phi$ being either its characteristic function or the signed distance to its contour. The authors aim at minimizing an objective function using diffuse approximation and gradient-based techniques, while staying on the manifold of admissible shapes.
Active Shape Models \cite{cootes1995active} provide another way to handle shapes in which the contour is discretized \cite{stegmann2002brief,wang2012kernel}.

Building a surrogate model in reduced dimension can be performed in different ways. The simplest is to restrict the metamodel to the most influential variables. 
But typical evaluation budgets are too narrow to find these variables before the optimization. 
Moreover, correlations might exist among the original dimensions (here CAD parameters) so that a selection of few variables may not constitute a valid reduced order description and meta-variables may be more appropriate. 
In \cite{wu2019developed}, the high-dimensional input space is circumvented by decomposing the model into a series of low-dimensional models after an ANOVA procedure. 
In \cite{bouhlel2016improving}, a kriging model is built in the space of the first Partial Least Squares axes for emphasizing the most relevant directions.
Related approaches for dimensionality reduction inside GPs consist in a projection of the input $\x$ on a lower dimensional hyperplane spanned by orthogonal vectors. These vectors are determined in different manners, e.g. by searching the active space in \cite{constantine2014active,li2019surrogate}, or during the hyper-parameters estimation in \cite{tripathy2016gaussian}. 
A more detailed bibliography of dimension reduction in GPs is conducted in Section \ref{section:GP_in_eigenbasis}.

For optimization purposes, the modes of discretized shapes \cite{stegmann2002brief} are integrated in a surrogate model in \cite{li2018data}. 
In \cite{cinquegrana2018investigation}, the optimization is carried out on the most relevant modes using evolutionary algorithms combined with an adaptive adjustment of the bounds of the design space, also employed in \cite{shan2004space}.
	
\begin{figure}[!ht]
	\begin{tabular}{m{0.5\textwidth}|m{0.5\textwidth}}
		\begin{minipage}{\linewidth}
			\centering
			\circled{1} Sample inputs $\x$ and apply $\phi(\x)$.
			\includegraphics[width=0.6\textwidth]{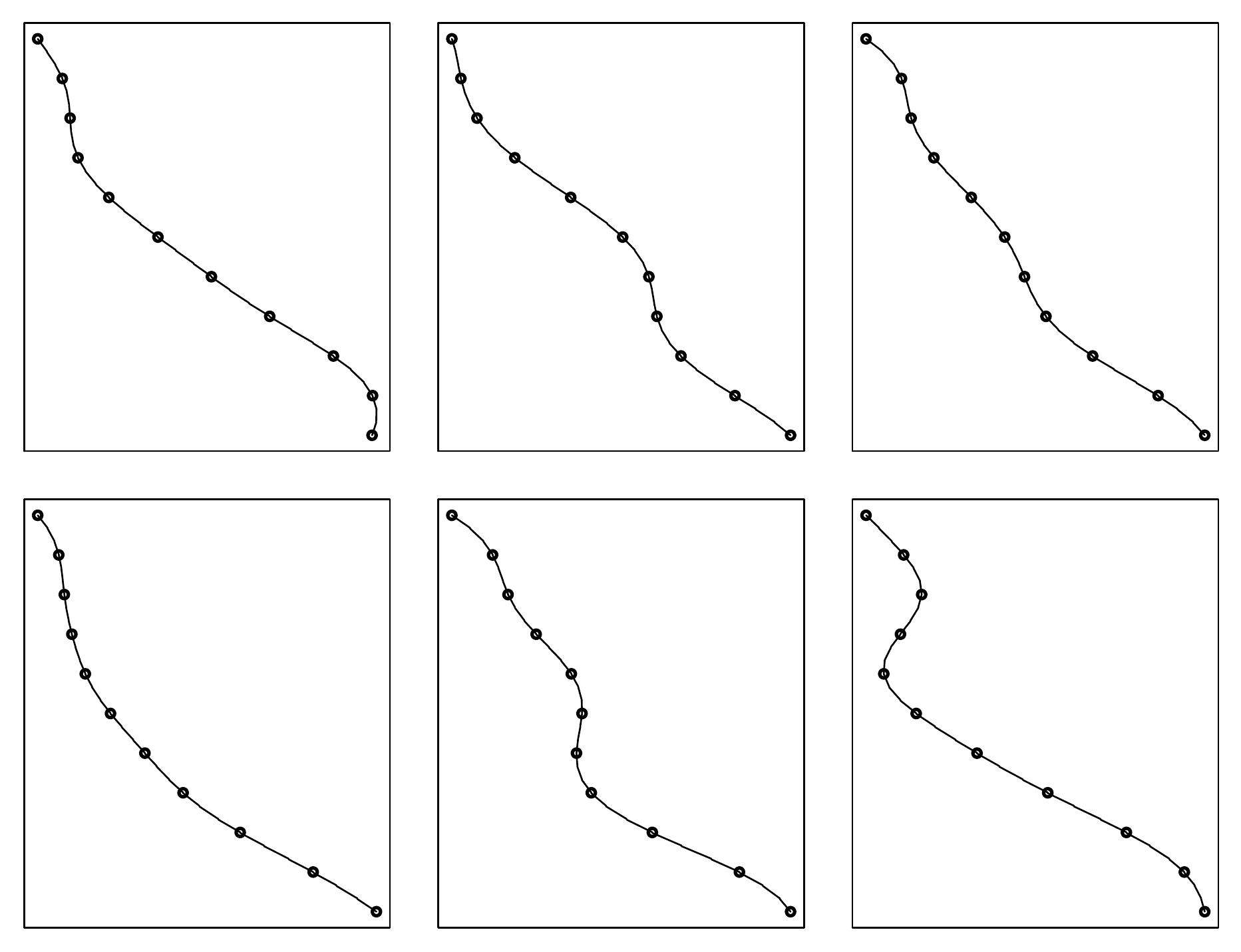}
		\end{minipage}
		
		&
		
		\begin{minipage}{\linewidth}
			\centering
			\circled{2} PCA: $\x$ mapped to $\pmb\alpha$\\in $\{\mathbf v^1,\dotsc,\mathbf v^D\}$ basis.\\
			\includegraphics[width=0.6\textwidth]{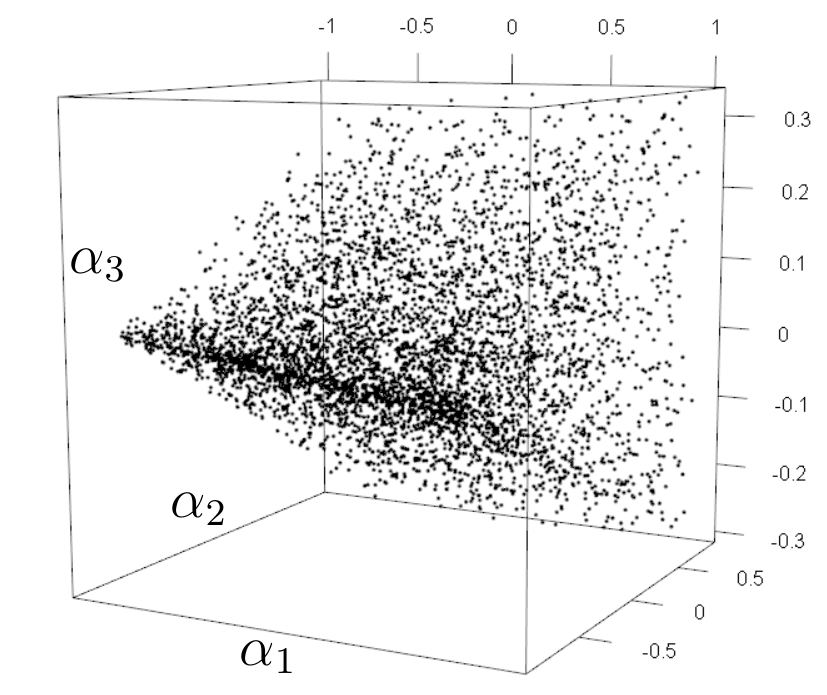}
		\end{minipage}
		
		\\\hline
		
		\begin{minipage}{\linewidth}
			\centering
			\circled{3} Determine the active and inactive eigendimensions for the output: $\pmb\alpha=[\pmb\alpha^a,\pmb\alpha^{\overline{a}}]$.
		\end{minipage}
		
		&
		
		\begin{minipage}{\linewidth}
			\centering\vspace{0.1cm}
			\circled{4} Build an additive GP with two different resolutions: $Y(\pmb\alpha)=\beta+Y^a(\pmb\alpha^a)+Y^{\overline{a}}(\pmb\alpha^{\overline{a}})$.
			\includegraphics[width=0.4\textwidth]{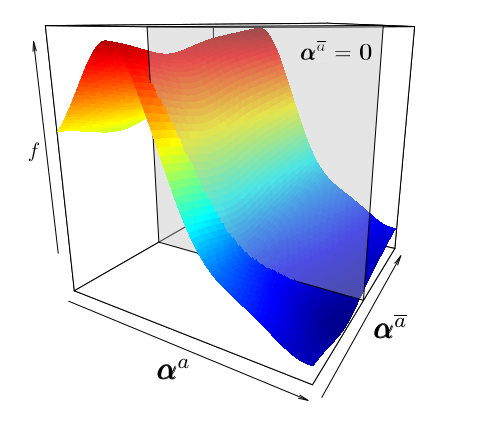}
		\end{minipage}
		
		\\\hline
		
		\begin{minipage}{\linewidth}
			\centering\vspace{0.1cm}
			\circled{5} Optimization in $\pmb\alpha^a$ space $\bigoplus$ random embedding in $\pmb\alpha^{\overline{a}}$ space $\Rightarrow\pmb\alpha^{(n+1)^*}$.
			\includegraphics[width=0.96\textwidth]{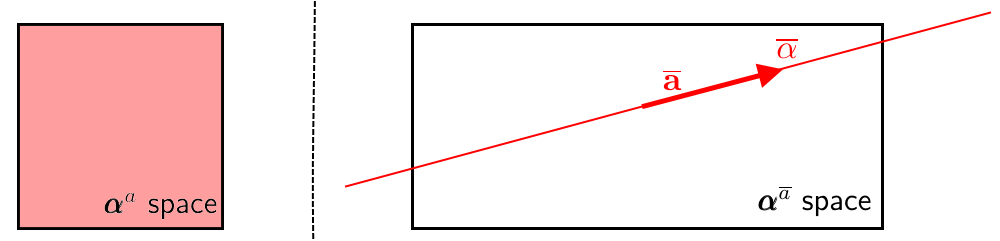}
		\end{minipage}
		
		&
		
		\begin{minipage}{\linewidth}
			\centering
			\circled{6} Solve the pre-image problem: $\pmb\alpha^{(n+1)^*}\Rightarrow\x^{(n+1)}$;\\update the GP with $(\pmb\alpha(\x^{(n+1)}),f(\x^{(n+1)}))$ and another point if replication used.
		\end{minipage}

	\end{tabular}
	\caption{Summary of the proposed method. Steps 3-6 are iterated during the optimization process.}
	\label{fig:summary}
\end{figure}

Following the same route, in Section \ref{sec:CAD_to_eigenbasis}, 
we retrieve a shape manifold with dimension $\delta<d$. 
Our approach is based on a Principal Component Analysis (PCA, \cite{wall2003singular}) of shapes described in an ad hoc manner in the same vein as \cite{cinquegrana2018investigation,li2018data} but it provides a new investigation of the best way to characterize shapes. 
Section \ref{section:GP_in_eigenbasis} is devoted to the construction of a kriging surrogate model in reduced dimension. 
Contrarily to \cite{li2018data,li2019surrogate}, the least important dimensions are still accounted for. A regularized likelihood approach is employed for dimension selection, instead of the linear PLS method \cite{bouhlel2016improving}. 
In Section \ref{section:optim_in_eigenbasis}, we employ the metamodel to perform global optimization \cite{jones1998efficient} via the maximization of the Expected Improvement \cite{ExpectedImprovement}. A reduction of the space dimension is achieved through a random embedding technique \cite{wang2013bayesian} and a pre-image problem is solved to keep the correspondence between the eigenshapes and the CAD parameters. The proposed method is summarized in Figure \ref{fig:summary}.

\subsection*{Main notations}

\begin{table}[!ht]
	\begin{tabular}{ll}
		$\mathcal A$ & Manifold of $\pmb\alpha$'s for which $\exists\x\in X$: $\mathbf V^\top(\phi(\x)-\overline{\pmb\phi})=\pmb\alpha$.\\
		$\mathcal A_N$ & Empirical manifold of $\pmb\alpha$'s which are the coordinates of the $\phi(\x^{(i)})$'s in the eigenbasis.\\
		$\pmb\alpha$ & Coordinates of a design in the eigenshape basis.\\
		$\pmb\alpha^a$ & Active components of $\pmb\alpha$.\\
		$\pmb\alpha^{\overline{a}}$ & Inactive components of $\pmb\alpha$.\\
		$d$ & Number of (CAD) parameters.\\
		$d'$ & True effective dimension.\\
		$\delta$ & Number of chosen/selected components for dimension reduction.\\
		$D$ & Dimension of the high-dimensional shape representation.\\
		$n$ & Number of evaluated designs.\\
		$N$ & Number of shapes in the $\pmb\Phi$ database.\\
		$\Omega_{\x}$ & Shape induced by the $\x$ parameterization.\\
		$\phi(\cdot)$ & High-dimensional shape mapping, $\phi: X \mapsto\Phi$\\
		$\Phi$ & Space of shape discretizations, $\Phi\subset\R^D$.\\
		$\pmb\phi$ & High-dimensional shape representation of one design ($\pmb\phi\in\R^D$).\\
		$\overline{\pmb\phi}$ & Mean shape in the $\pmb\Phi$ database.\\
		$\pmb\Phi$ & Shape database ($N\times D$ matrix whose $i$-th row is $\phi(\x^{(i)})$).\\
		$\mathbf V$ & $D\times D$ matrix whose columns $(\v^1,\dotsc,\v^D)$ are the eigenvectors of\\ & the covariance matrix of $\pmb\Phi$. They are the vectors of the orthonormal $\mathcal V$ basis.\\
		$\x$ & Design vector in the space of CAD parameters, $\x\in X$.\\
		$X$ & Original search space (of CAD parameters), $X\subset\R^d$.
	\end{tabular}
\end{table}
	
\section{From CAD description to shape eigenbasis}
\label{sec:CAD_to_eigenbasis}
	
CAD parameters are usually set up by engineers to automate shape generation. These parameters may be Bézier or Spline control points which locally readjust the shape. Other CAD parameters, such as the overall width or the length of a component, have a more global impact on the shape. While these parameters are intuitive to a designer, they are not chosen to achieve any specific mathematical property and in particular do not let themselves interpret to reduce dimensionality.

In order to define a better behaved description of the shapes that will help in reducing dimensionality, we exploit the fact that the time to generate a shape $\Omega_\x$ is negligible in comparison with the evaluation time of $f(\x)$. 
	
In the spirit of kernel methods \cite{vapnik1995nature,scholkopf1997kernel}, we analyze the designs $\x$ in a high-dimensional feature space $\Phi\subset\R^D$, $D\gg d$ (potentially infinite dimensional) that is defined via a mapping $\phi(\x)$, $\phi:X\rightarrow\Phi$. 
With an appropriate $\phi(\cdot)$, it is possible to distinguish a lower dimensional manifold embedded in $\Phi$. 
As we deal with shapes, natural candidates for $\phi(\cdot)$ are shape representations.

This paper is motivated by parametric shape optimization problems. 
However, the approaches developed for metamodeling and optimization are generic and extend to any situation where a pre-existing collection of designs $\{\x^{(1)},\dotsc,\x^{(N)}\}$ and a fast auxiliary mapping $\phi(\x)$ exist. $\phi(\x)=\x$ is a possible case.
If $\x$ are parameters that generate a signal, another example would be $\phi(\x)$, the discretized times series.

\subsection{Shape representations}
In the literature, shapes have been described in different ways.
First, the \emph{characteristic function} of a shape $\Omega_\x$ \cite{raghavan2013towards} is \begin{equation}
	\chi_{\Omega_\x}(\mathbf s)=
	\left\{
	\begin{aligned}
	1\text{ if }\mathbf s\in{\Omega_\x}\\
	0\text{ if }\mathbf s\notin{\Omega_\x}\\
	\end{aligned}
	\right.		
	\end{equation}
where $\mathbf s\in\R^2$ or $\R^3$ is the spatial coordinate. $\chi$ is computed at some relevant locations (e.g. on a grid) $\mathbb S=\{\mathbf s^{(1)},\dotsc,\mathbf s^{(D)}\}$ and is cast as a $D$-dimensional vector of of 0's or 1's depending on whether the $\mathbf s^{(i)}$'s are inside or outside the shape.

Second, the \emph{signed distance to the contour} $\partial\Omega_\x$ \cite{raghavan2014numerical} is 

\begin{equation}
\mathbb D_{\Omega_\x}(\mathbf s)=\varepsilon(\mathbf s)\underset{\mathbf y\in\partial{\Omega_\x}}{\min}\Vert\mathbf s-\mathbf y\Vert_2\text{, where }\varepsilon(\mathbf s)=\left\{
\begin{aligned}
1\text{ if }\mathbf s\in{\Omega_\x}\\
-1\text{ if }\mathbf s\notin{\Omega_\x}\\
\end{aligned}
\right.		
\end{equation}
and is also computed at some relevant locations (e.g. on a grid) $\mathbb S$, transformed into a vector with $D$ components.
 
\begin{figure}[h!]
	\centering
	\includegraphics[width=0.12\textwidth]{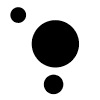}
	\includegraphics[width=0.12\textwidth]{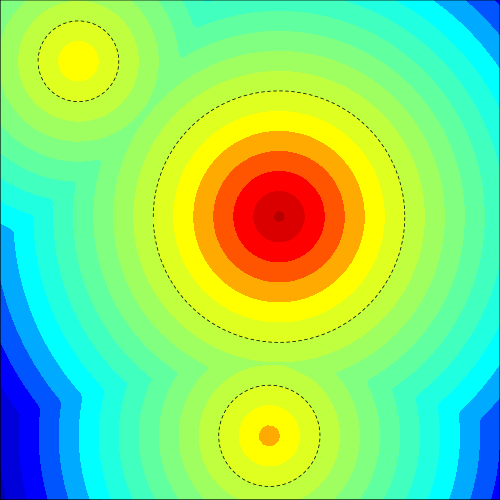}
	\includegraphics[width=0.12\textwidth]{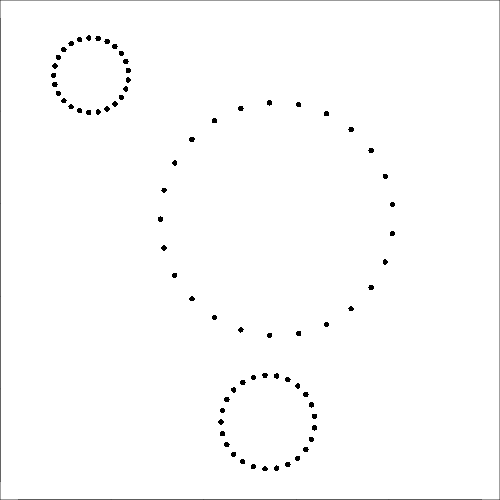}\\
	\includegraphics[width=0.12\textwidth]{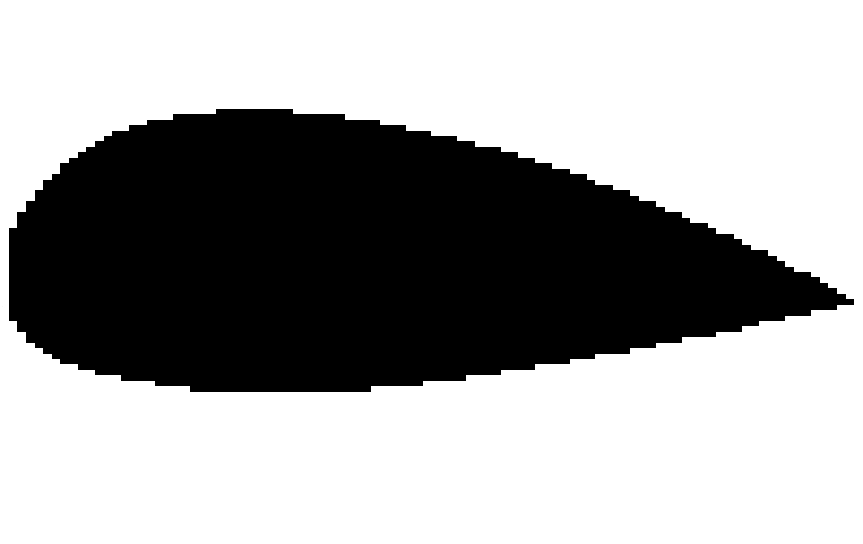}
	\includegraphics[width=0.12\textwidth]{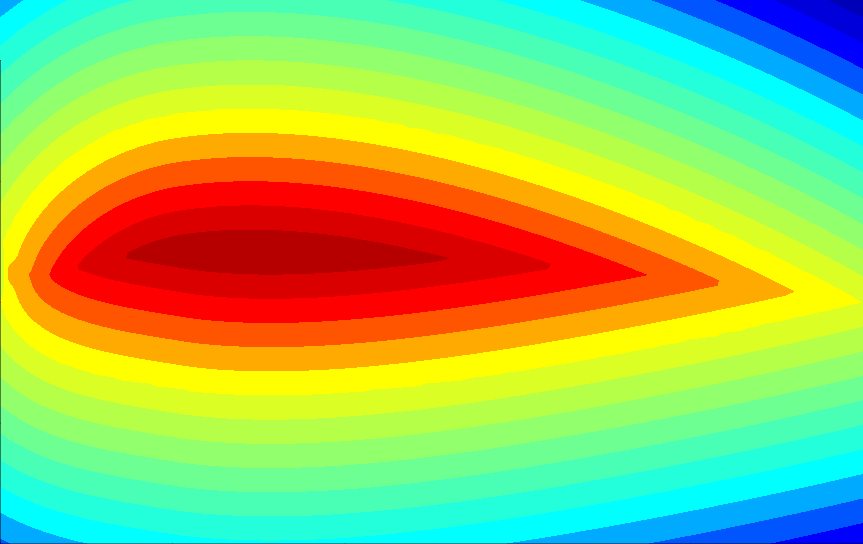}
	\includegraphics[width=0.12\textwidth]{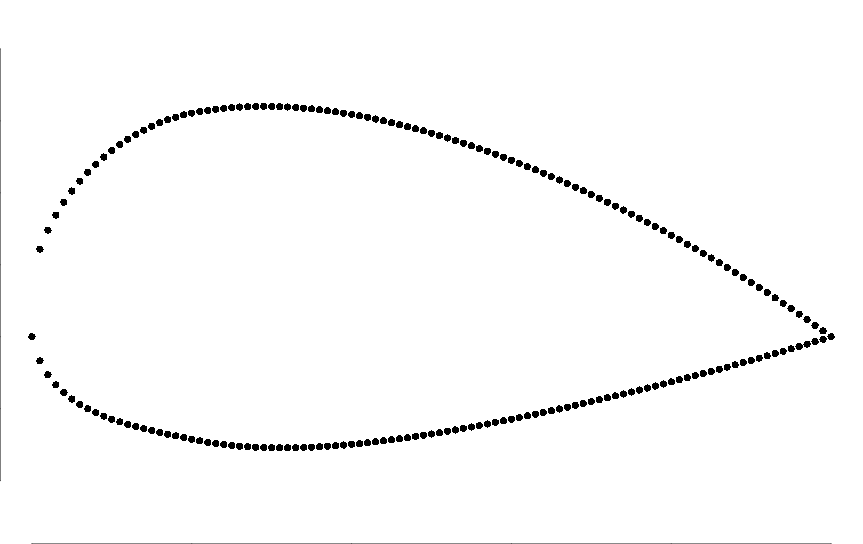}
	\caption{Shape representations for a design consisting of three circles (top) and for a NACA airfoil (bottom). The representations are the characteristic function (left), the signed distance to the contour (center), and the contour discretization(right).}
	\label{fig:shape_representation}
\end{figure}

Finally, the Point Distribution Model \cite{cootes1995active,stegmann2002brief} where $\partial\Omega_\x$ is discretized at $D/k$ locations $\mathbf s^{(i)}\in\partial\Omega_\x\subset\R^k$ ($k=2$ or 3), also leads to a $D$-dimensional representation of $\Omega_\x$ where $\mathcal D_{\Omega_\x}=({\mathbf s^{(1)}}^\top,\dotsc,{\mathbf s^{(D/k)}}^\top)^\top\in\R^D$. 
For different shapes $\Omega$ and $\Omega'$, $\mathbb S$ has to be the same for $\chi$ and $\mathbb D$, and the \emph{discretizations} $\{{\mathbf s^{(1)}}^\top,\dotsc,{\mathbf s^{(D/k)}}^\top\}$ of $\Omega$ and $\Omega'$ need to be consistent for $\mathcal D$.
Figure \ref{fig:shape_representation} illustrates these shape representations for two different designs. The first one consists of three circles parameterized by their centers and radii. 
The second design is a NACA airfoil which depends on three parameters. 
These shapes are described by the mappings $\phi(\x)\in\R^D$ with $\phi(\x)=\chi_{\Omega_\x}(\mathbb S),\mathbb D_{\Omega_\x}(\mathbb S)$ and $\mathcal D_{\Omega_\x}$, respectively. 
Specifying another design with parameters $\x'$ generally leads to $\phi(\x)\ne\phi(\x')$.

\subsection{PCA to retrieve the effective shape dimension}
\label{section:PCA}
During the step \circled{1} of our method (see Figure \ref{fig:summary}), a large number $(N)$ of plausible designs $\x^{(i)}\in X$ is mapped to $\Phi\subset\R^D$ and build the matrix $\pmb\Phi\in\R^{N\times D}$ which contains the $\phi(\x^{(i)})\in\R^D$ in rows and whose column-wise mean is $\overline{\pmb\phi}\in\R^D$.
In the absence of a set of relevant $\x^{(i)}$'s, these designs can be sampled from an a priori distribution, typically a uniform distribution.
Next (step \circled{2} in Figure \ref{fig:summary}), we perform a Principal Component Analysis (PCA) on $\pmb\Phi$: correlations are sought between the $\phi(\x)_j$'s, $j=1,\dotsc,D$.
The eigenvectors of the empirical covariance matrix 
$\mathbf C_{\pmb\Phi}:=\frac1N(\pmb\Phi-\mathbf1_N\overline{\pmb\phi}^\top)^\top(\pmb\Phi-\mathbf1_N\overline{\pmb\phi}^\top)$,
written $\v^j\in\R^D$, form an ordered orthonormal basis of $\Phi$ with decreasing importance as measured by the PCA eigenvalues $\lambda_j$, $j=1,\dotsc,D$.
They correspond to orthonormal directions in $\Phi$ that explain the most the dispersion of the high-dimensional representations of the shapes, $\phi(\x^{(i)})$. 
Any design $\x$ can now be expressed in the eigenbasis $\mathcal V:=\{\v^1,\dotsc,\v^D\}$ since
\begin{equation}
\phi(\x)=\overline{\pmb\phi}+\sum_{j=1}^{D}\alpha_j\v^j
\label{eq:sumav}
\end{equation}
where $(\alpha_1,\dotsc,\alpha_D)^\top=:\pmb\alpha=\mathbf V^\top(\phi(\x)-\overline{\pmb\phi})$ are the coordinates in $\mathcal V$  (principal components), and $\mathbf V:=(\v^1,\dotsc,\v^D)\in\R^{D\times D}$ is the matrix of eigenvectors (principal axes). 
$\alpha_j$ is the deviation from the mean shape $\overline{\pmb\phi}$, in the direction of the eigenvector $\v^j$. 
The $\pmb\alpha^{(i)}$'s form a manifold $\mathcal A_N:=\{\pmb\alpha^{(1)},\dotsc,\pmb\alpha^{(N)}\}$ which approximates the true $\pmb\alpha$ manifold, $\mathcal A:=\{\pmb\alpha\in\R^D:\exists\x\in X$, $\pmb\alpha=\mathbf V^\top(\phi(\x)-\overline{\pmb\phi}) \}$. 
Even though $\mathcal A_N\subset\R^D$, it is often a manifold of lower dimension, $\delta\ll D$, as we will soon see (Section \ref{sec:experiments_reduction}).

\subsubsection*{Link with kernel PCA}
$N$ designs $\x^{(i)}\in\R^d$ have been mapped to a high-dimensional feature space $\Phi\subset\R^D$ in which PCA was carried out. 
This is precisely the task that is performed in Kernel PCA \cite{scholkopf1997kernel}, 
a nonlinear dimension reduction technique (contrarily to PCA which seeks linear directions in $\R^d$). KPCA aims at finding a linear description of the data in a feature space $\Phi$, by applying a PCA to nonlinearly mapped $\phi(\x^{(i)})\in\Phi$. 
The difference with our approach is that the mapping $\phi(\cdot)$ as well as the feature space $\Phi$ are usually unknown in KPCA, since $\phi(\x)$ may live in a very high-dimensional or even infinite dimensional space in which dot products cannot be computed efficiently. 
Instead, dot products are computed using designs in the original space $X$ via a \emph{kernel}
which should not be mistaken with the kernel of GPs,
$k_\phi:X\times X\rightarrow\R$, $k_\phi(\x,\x')=\langle\phi(\x),\phi(\x')\rangle_{\Phi}$ (this is called the ``kernel-trick'' \cite{vapnik1995nature,scholkopf1997kernel}).
The eigencomponents of the points after mapping, $\alpha^{(i)}_j = {\v^j}^\top (\phi(\x^{(i)})-\overline{\pmb\phi})$, can be recovered from the eigenanalysis of the $N\times N$ Gram matrix 
$\mathbf K$ with $K_{ij}=k_\phi(\x^{(i)},\x^{(j)})$ (see \cite{scholkopf1997kernel,wang2012kernel} for algebraic details). 
Finding which original variables in $\x$ correspond to a given $\v^j$ is not straightforward and requires the resolution of a pre-image problem \cite{mika1999kernel,wang2012kernel}.

Having a shape-related and computable $\phi(\cdot)$ avoids these ruses and makes the principal axes $\v^j$ directly meaningful. It is further possible to give the expression of the equivalent kernel in our approach, in terms of the mapping $\phi(\cdot)$, from the polarization identity.
By definition of the (centered) high dimensional mapping to $\Phi$, $\x\mapsto\phi(\x)-\overline{\pmb\phi}$,
\begin{align*}\Vert(\phi(\x)-\overline{\pmb\phi})-(\phi(\x')-\overline{\pmb\phi})\Vert^2_{\R^D}=\langle(\phi(\x)-\overline{\pmb\phi})-(\phi(\x')-\overline{\pmb\phi}),(\phi(\x)-\overline{\pmb\phi})-(\phi(\x')-\overline{\pmb\phi})\rangle_{\R^D}\\
=\Vert(\phi(\x)-\overline{\pmb\phi})\Vert^2_{\R^D}+\Vert(\phi(\x')-\overline{\pmb\phi})\Vert^2_{\R^D}-2\underset{k_\phi}{\underbrace{\langle(\phi(\x)-\overline{\pmb\phi}),(\phi(\x')-\overline{\pmb\phi})\rangle_{\R^D}}}
\end{align*}
hence, 
\begin{equation}
k_{\phi}(\x,\x')=\frac12(\Vert\phi(\x)-\overline{\pmb\phi}\Vert^2_{\R^D}+\Vert\phi(\x')-\overline{\pmb\phi}\Vert^2_{\R^D}-\Vert\phi(\x)-\phi(\x')\Vert^2_{\R^D})
\end{equation}
Logically, $k_{\phi}(\cdot,\cdot)$, a similarity measure between designs, is negatively proportional to the distance between the shape representations. 
Because of the size of the eigenanalyses to be performed, kernel PCA is advantageous over a mapping followed by a PCA when $D > N$, i.e. when the shapes have a very high resolution, and vice versa. In the current work where $\phi(\cdot)$ is known and $D$ is smaller than 1000, we will follow the mapping plus PCA approach.

\subsection{Experiments}
\label{sec:experiments_reduction}
In this section, all the parametric design problems used in the experiments throughout this paper are introduced and discussed in terms of significant dimensions.  
Unless stated otherwise, the database $\pmb\Phi$ is made of $N=5000$ designs sampled uniformly in $X$.
We start with 3 test cases of known intrinsic dimension, which will be complemented by 4 other test cases. The metamodeling and the optimization will be addressed later in Sections \ref{section:GP_in_eigenbasis} and \ref{section:optim_in_eigenbasis}.
\subsubsection{Retrieval of true dimensionality}
\begin{figure}[h!]
		\centering
		\includegraphics[width=0.8\textwidth]{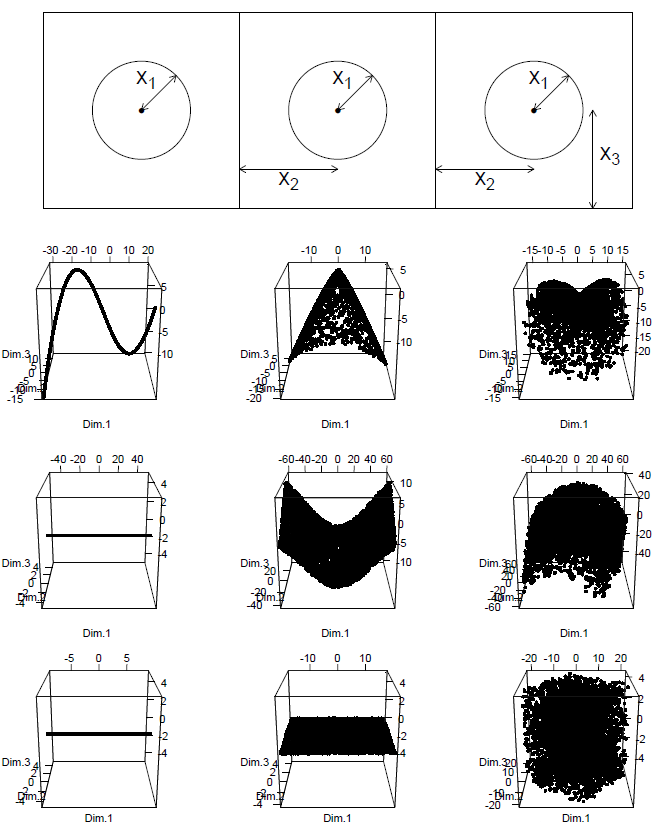}
		\caption{Example \ref{ex:un_cercle}: three first eigencomponents of the $\pmb\alpha^{(i)}$'s for three parametric test cases (columns) with low effective dimension equal to 1 (left), 2 (center) and 3 (right). The rows correspond to different $\phi(\cdot)$'s which are the characteristic function (top), the signed distance to the contour (middle) and the discretization of the contour (bottom).
		}
		\label{fig:manifold}
	\end{figure}

\begin{figure}[h!]
		\centering
		\includegraphics[width=0.5\textwidth]{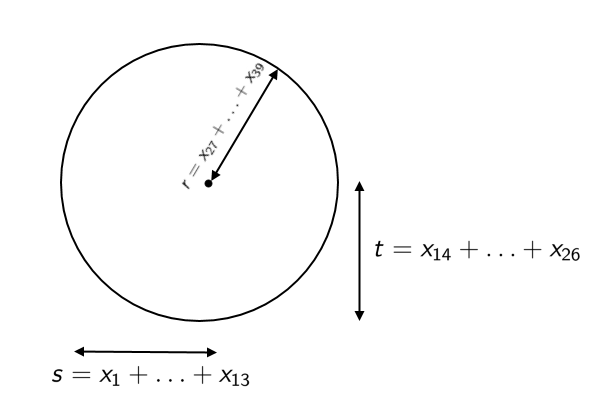}
		\caption{Second example: an over-parameterized circle.}
		\label{fig:cercle_surparametre}
	\end{figure}

\begin{figure}[h!]
		\centering
		\includegraphics[width=0.7\textwidth]{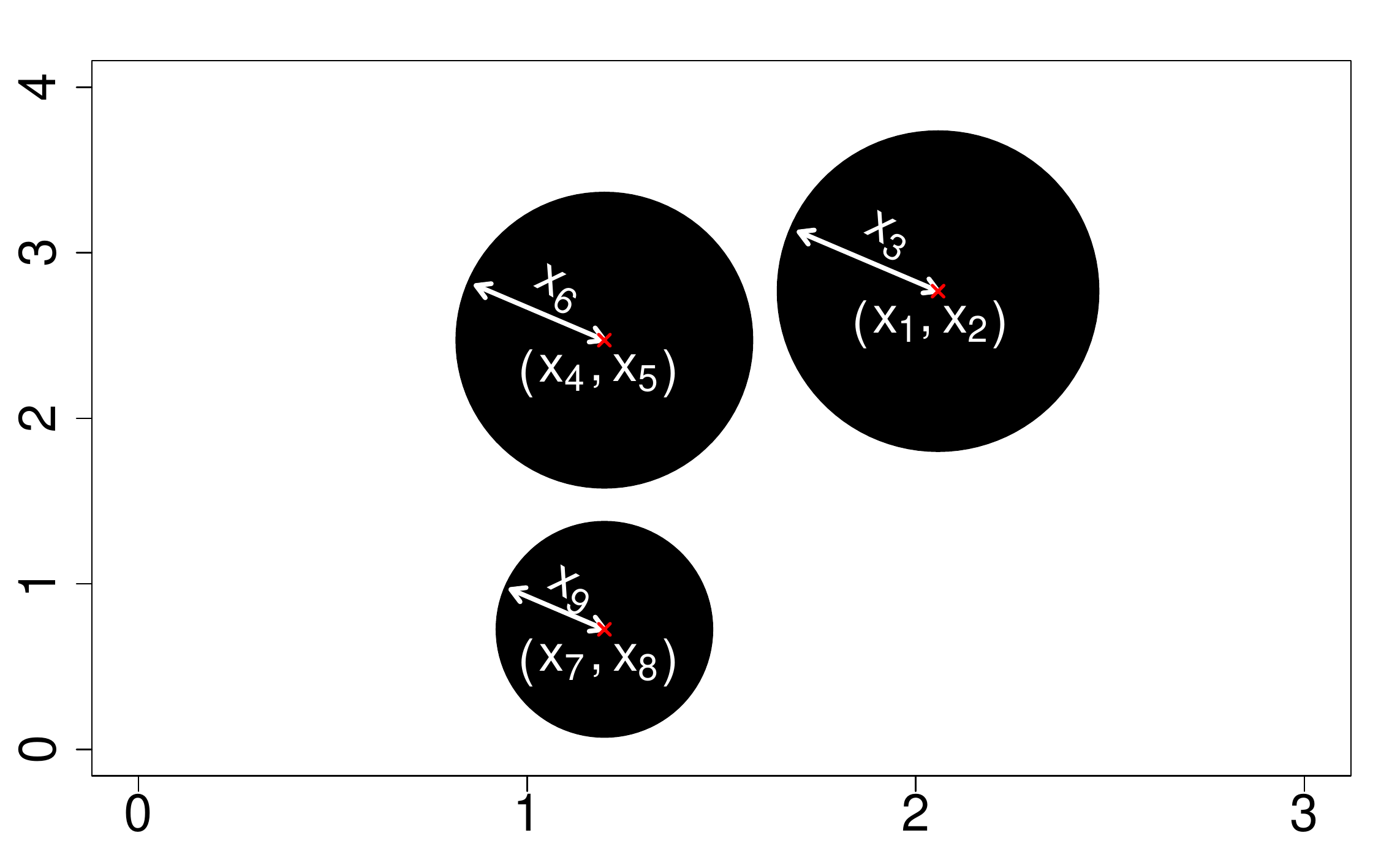}
		\caption{Third example: three circles with varying centers and radii.}
		\label{fig:3cercles}
	\end{figure}

In this part, we generate shapes of known low intrinsic dimension. 
In the Example \ref{ex:un_cercle} (cf. Figure \ref{fig:manifold}), the shapes are circular holes of varying centers and radii, therefore described by 1, 2 or 3 parameters. 
In the Example \ref{ex:cercle_surparametre} (cf. Figure \ref{fig:cercle_surparametre}), they are also circular holes but whose center positions and radii are described by sums\footnote{other algebraic operations such as multiplications have also led to the same conclusions.} of parts of the 39 parameters. 
Last, in the Example \ref{ex:3cercles} (cf. Figure \ref{fig:3cercles}), the shapes are made of three non overlapping circles with parameterized centers and radii. 
PCAs were then carried out on the $\pmb\Phi$'s associated to the three mappings (characteristic function, signed contour distance and contour discretization). In each example, the 10 first PCA eigenvalues $\lambda_j$ are reported. The $\pmb\alpha$'s manifolds, $\mathcal A_N\subset\R^D$, are plotted in the first three dimensions as well as the first eigenvectors in the $\Phi$ space.

\begin{example}
	A hole in $\R^2$ parameterized by its radius ($d=1$), its radius and the $x$-coordinate of its center ($d=2$), or its radius and the $x$ and $y$ coordinates of its center ($d=3$).
	\label{ex:un_cercle}
\end{example}

	\clearpage

\newgeometry{bottom=4cm,top=4cm}
\begin{table}[!ht]
	\centering
	\makebox[\textwidth][c]{
		\begin{tabu}{|c|c|c|c|c|c|c|}
			\hline
			& \multicolumn{2}{c|}{Characteristic function} & \multicolumn{2}{c|}{Signed Distance} & \multicolumn{2}{c|}{Discretization}\\\hline
			$j$ & Eigenvalue & Cumulative percentage & Eigenvalue & Cumulative percentage & Eigenvalue & Cumulative percentage\\\hline
			1 & 324.63 & 63.09 & 840.14 & 100 & 25.20 & 100\\
			2 & 75.98 & 77.86 & 0 & 100 & 0 & 100\\
			3 & 32.69 & 84.21 & 0 & 100 & 0 & 100\\
			4 & 18.20 & 87.75 & 0 & 100 & 0 & 100\\
			5 & 11.48 & 89.98 & 0 & 100 & 0 & 100\\
			6 & 8.12 & 91.56 & 0 & 100 & 0 & 100\\
			7 & 5.92 & 92.71 & 0 & 100 & 0 & 100\\
			8 & 4.45 & 93.57 & 0 & 100 & 0 & 100\\
			9 & 3.50 & 94.25 & 0 & 100 & 0 & 100\\
			10 & 2.79 & 94.80 & 0 & 100 & 0 & 100\\\hline
		\end{tabu}
	}
	\caption{10 first PCA eigenvalues for the different $\phi(\cdot)$'s, circle with $d=1$ parameter.}
	\label{tab:eigenvalues_d1}
\end{table}

\begin{table}[!ht]
	\centering
	\makebox[\textwidth][c]{
		\begin{tabu}{|c|c|c|c|c|c|c|}
			\hline
			& \multicolumn{2}{c|}{Characteristic function} & \multicolumn{2}{c|}{Signed Distance} & \multicolumn{2}{c|}{Discretization}\\\hline
			$j$ & Eigenvalue & Cumulative percentage & Eigenvalue & Cumulative percentage & Eigenvalue & Cumulative percentage\\\hline
			1 & 60.90 & 26.50 & 1332.17 & 80.41 & 100.82 & 94.14\\
			2 & 44.63 & 45.93 & 294.07 & 98.15 & 6.27 & 100\\
			3 & 26.70 & 57.55 & 25.48 & 99.69 & 0 & 100\\
			4 & 20.62 & 66.52 & 3.88 & 99.93 & 0 & 100\\
			5 & 9.48 & 70.65 & 0.81 & 99.97 & 0 & 100\\
			6 & 4.87 & 72.77 & 0.24 & 99.99 & 0 & 100\\
			7 & 3.97 & 74.49 & 0.09 & 99.99 & 0 & 100\\
			8 & 3.74 & 76.12 & 0.04 & 100 & 0 & 100\\
			9 & 3.25 & 77.54 & 0.02 & 100 & 0 & 100\\
			10 & 3.11 & 78.89 & 0.01 & 100 & 0 & 100\\\hline
		\end{tabu}
	}
	\caption{10 first PCA eigenvalues for the different $\phi(\cdot)$'s, circle with $d=2$ parameters.}
	\label{tab:eigenvalues_d2}
\end{table}

\begin{table}[!ht]
	\centering
	\makebox[\textwidth][c]{
		\begin{tabu}{|c|c|c|c|c|c|c|}
			\hline
			& \multicolumn{2}{c|}{Characteristic function} & \multicolumn{2}{c|}{Signed Distance} & \multicolumn{2}{c|}{Discretization}\\\hline
			$j$ & Eigenvalue & Cumulative percentage & Eigenvalue & Cumulative percentage & Eigenvalue & Cumulative percentage\\\hline
			1 & 26.48 & 10.12 & 1045.26 & 42.42 & 82.13 & 48.51\\
			2 & 25.82 & 19.98 & 1037.44 & 84.53 & 80.82 & 96.26\\
			3 & 20.58 & 27.84 & 300.14 & 96.71 & 6.34 & 100\\
			4 & 19.38 & 35.24 & 33.83 & 98.08 & 0 & 100\\
			5 & 15.65 & 41.22 & 18.49 & 98.83 & 0 & 100\\
			6 & 11.36 & 45.56 & 14.40 & 99.42 & 0 & 100\\
			7 & 11.20 & 49.84 & 3.78 & 99.57 & 0 & 100\\
			8 & 11.05 & 54.06 & 3.64 & 99.72 & 0 & 100\\
			9 & 7.52 & 56.93 & 1.58 & 99.78 & 0 & 100\\
			10 & 7.21 & 59.69 & 1.55 & 99.84 & 0 & 100\\\hline
		\end{tabu}
	}
	\caption{10 first PCA eigenvalues for the different $\phi(\cdot)$'s, circle with $d=3$ parameters.}
	\label{tab:eigenvalues_d3}
\end{table}\restoregeometry\clearpage

\begin{figure}[h!]
	\centering
	\includegraphics[width=0.6\textwidth]{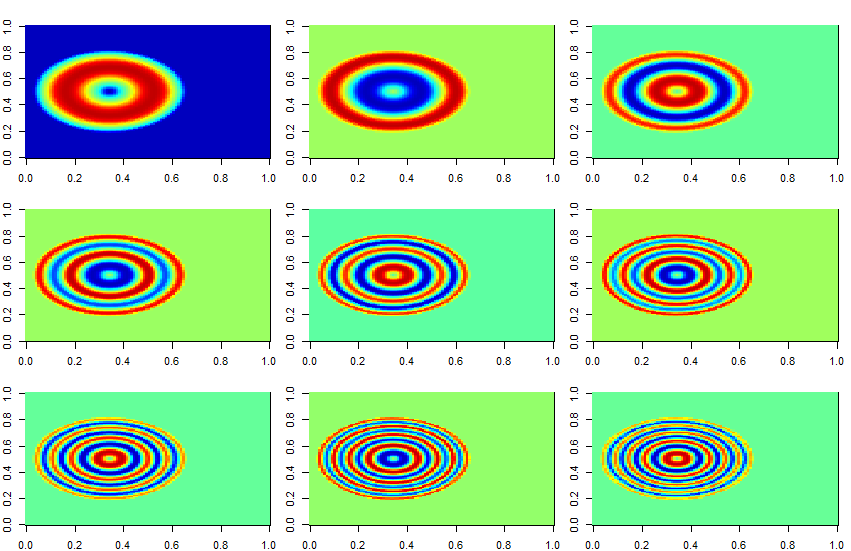}
	\caption{Example \ref{ex:un_cercle}, circle with $d=1$ parameter, 9 first eigenvectors (left to right and top to bottom) when $\phi(\cdot)$ = characteristic function.}
	\label{fig:ex1_d1_image}
\end{figure}

\begin{figure}[h!]
\centering
\includegraphics[width=0.48\textwidth]{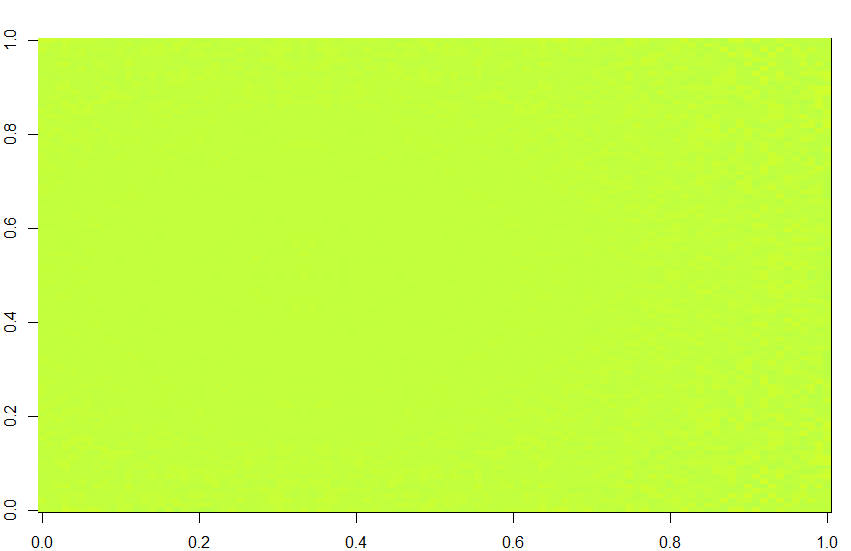}
\includegraphics[width=0.35\textwidth]{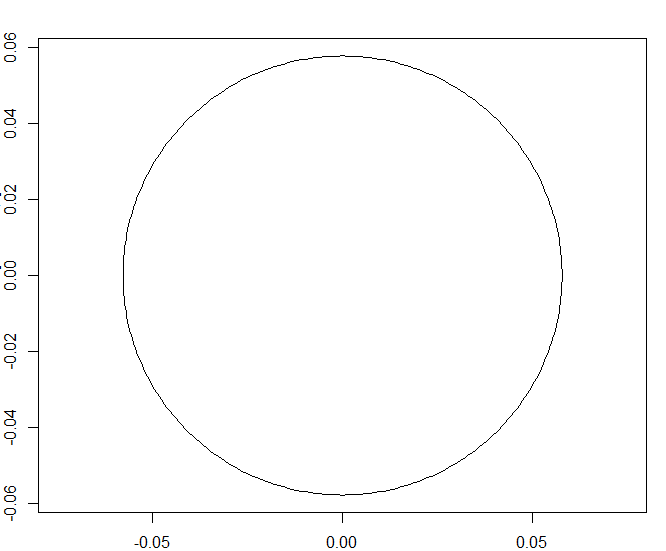}
\caption{Example \ref{ex:un_cercle}, circle with $d=1$ parameter, first eigenvector when $\phi(\cdot)$ = signed distance (left) and when $\phi(\cdot)$ = contour discretization (right).}
\label{fig:ex1_d1_signed_distance}
\end{figure}

\begin{figure}[h!]
	\centering
	\includegraphics[width=0.6\textwidth]{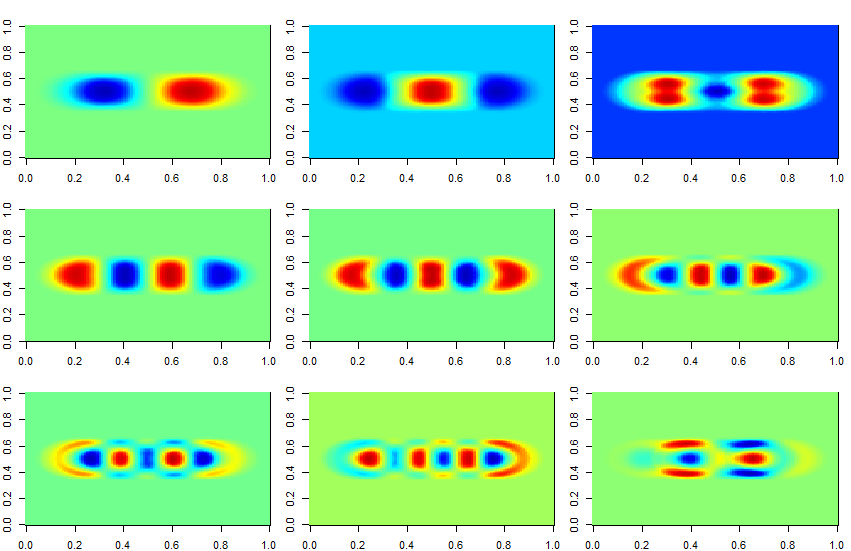}
	\caption{Example \ref{ex:un_cercle}, circle with $d=2$ parameters, 9 first eigenvectors (left to right and top to bottom) when $\phi(\cdot)$ = characteristic function.}
	\label{fig:ex1_d2_image}
\end{figure}

\begin{figure}[h!]
	\centering
	\includegraphics[width=0.6\textwidth]{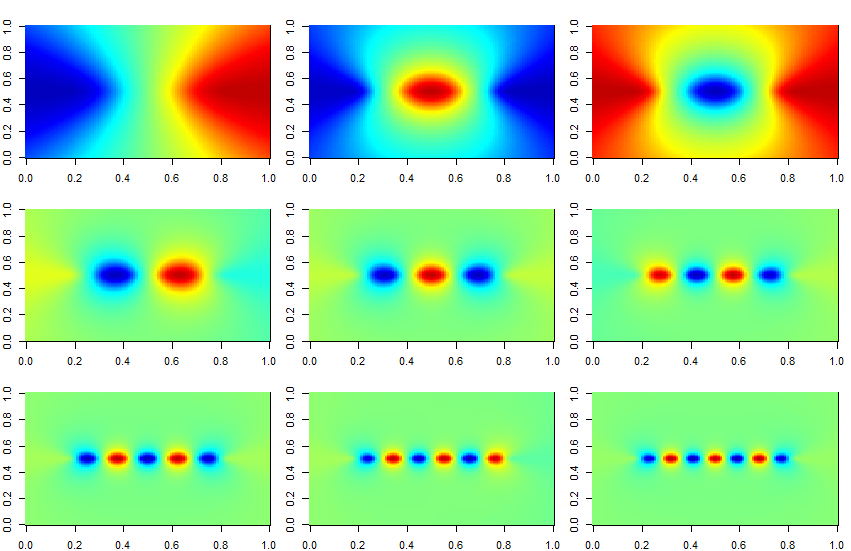}
	\caption{Example \ref{ex:un_cercle}, circle with $d=2$ parameters, 9 first eigenvectors (left to right and top to bottom) when $\phi(\cdot)$ = signed distance.}
	\label{fig:ex1_d2_signed_distance}
\end{figure}

\begin{figure}[h!]
	\centering
	\includegraphics[width=0.4\textwidth]{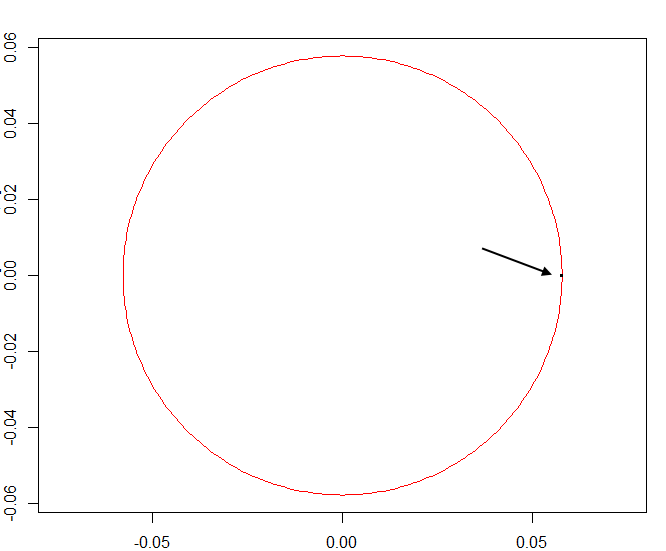}
	\caption{Example \ref{ex:un_cercle}, circle with $d=2$ parameters, 2 first eigenvectors (black and red) when $\phi(\cdot)$ = contour discretization.}
	\label{fig:ex1_d2_pdm}
\end{figure}

\begin{figure}[h!]
\centering
\includegraphics[width=0.6\textwidth]{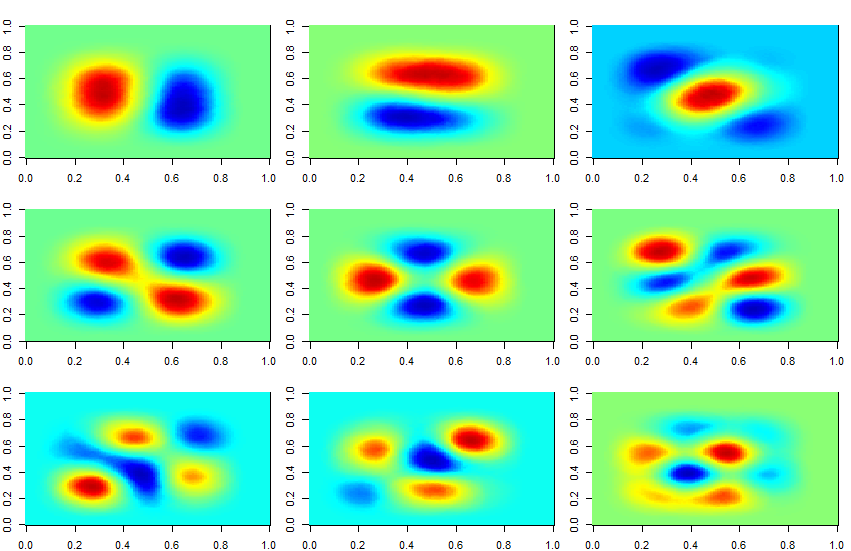}
\caption{Example \ref{ex:un_cercle}, circle with $d=3$ parameters, 9 first eigenvectors (left to right and top to bottom) when $\phi(\cdot)$ = characteristic function.}
\label{fig:ex1_d3_image}
\end{figure}

\begin{figure}[h!]
\centering
\includegraphics[width=0.6\textwidth]{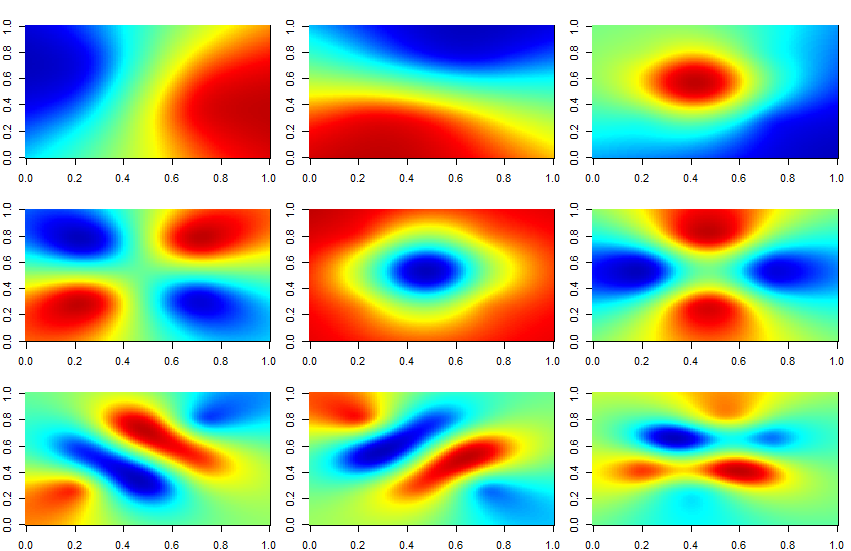}
\caption{Example \ref{ex:un_cercle}, circle with $d=3$ parameters, 9 first eigenvectors (left to right and top to bottom) when $\phi(\cdot)$ = signed distance.}
\label{fig:ex1_d3_signed_distance}
\end{figure}

\begin{figure}[h!]
\centering
\includegraphics[width=0.4\textwidth]{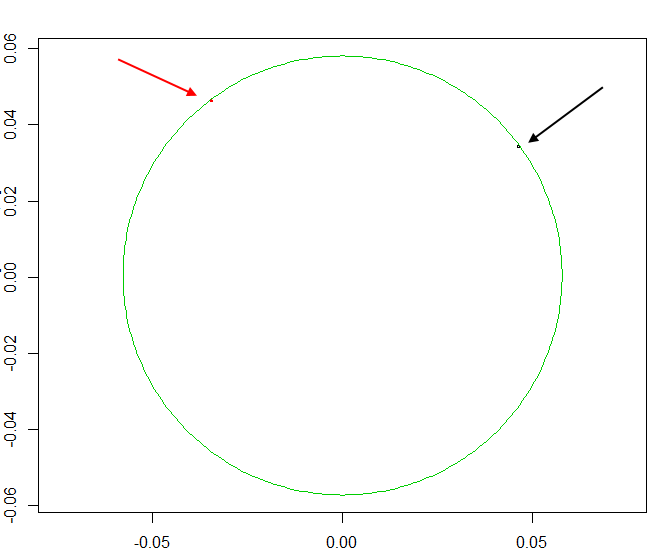}
\caption{Example \ref{ex:un_cercle}, circle with $d=3$ parameters, 3 first eigenvectors (black, red, green) when $\phi(\cdot)$ = contour discretization.}
\label{fig:ex1_d3_pdm}
\end{figure}

Figures \ref{fig:ex1_d1_image}-\ref{fig:ex1_d3_pdm} show the 9 first eigenvectors (if they have strictly positive eigenvalue) in the 3 cases of Example \ref{ex:un_cercle} with the three $\phi(\cdot)$'s.

\clearpage

A property of PCA is that a linear combination of the eigenvectors given in Equation (\ref{eq:sumav}) enables to retrieve any $\phi(\x^{(i)})$.
Some of the eigenvectors are easy to interpret: in Figure \ref{fig:ex1_d1_signed_distance} left (signed distance), the eigenvector is constant because the average shape is a map (an image) whose level lines are perfect circles so that adding a constant to it changes the radius of the null contour line; in Figure \ref{fig:ex1_d2_pdm} where the mapping is a contour discretization, the first eigenvector (as well as the second in Figure \ref{fig:ex1_d3_pdm}) is a non-centered point that allows horizontal (and vertical) translations. The second (third in Figure \ref{fig:ex1_d3_pdm}) eigenvector is a circle which dilates or compresses the hole.
As is seen in Tables \ref{tab:eigenvalues_d2} and \ref{tab:eigenvalues_d3}, more eigenvectors are necessary for the characteristic function and for the signed distance than for the contour discretization.
Contrarily to the characteristic function and the signed contour, when the mapping $\phi(\cdot)$ is the contour discretization, the first eigenvectors look like shapes on their own and therefore we will call them \emph{eigenshapes}. This does not mean however that all of them are valid shapes, as was seen in Figures \ref{fig:ex1_d2_pdm} and \ref{fig:ex1_d3_pdm} with the point vectors.
In fact, most $\v^j$'s are ``non-physical'' in the sense that there may not exist one design $\x$ such that $\phi(\x)=\v^j$, see for instance Figure \ref{fig:naca22_eigenshapes} where the eigenshapes do not correspond to a valid $\x$ from $\v^3$ on. 
In the case of the characteristic function, even though $\phi(\x)\in\{0,1\}^D$, the eigenvectors are real-valued (see Figure \ref{fig:ex1_d1_image} for instance).

\begin{example}
	An over-parameterized hole in $\R^2$: the horizontal position of its center is $s:=\sum_{j=1}^{13}x_j$, the vertical position of its center is $t:=\sum_{j=14}^{26}x_j$ and its radius is $r:=\sum_{j=27}^{39}x_j$, as shown in Figure \ref{fig:cercle_surparametre}. To increase the complexity of the problem, $x_1$, $x_{14}$ and $x_{27}$ are of a magnitude larger than the other $x_j$'s: the circle mainly depends on these 3 parameters. 
	\label{ex:cercle_surparametre}
\end{example}

	\clearpage

\newgeometry{bottom=4cm,top=3cm}
\begin{figure}[h!]
	\centering
	\includegraphics[width=0.32\textwidth]{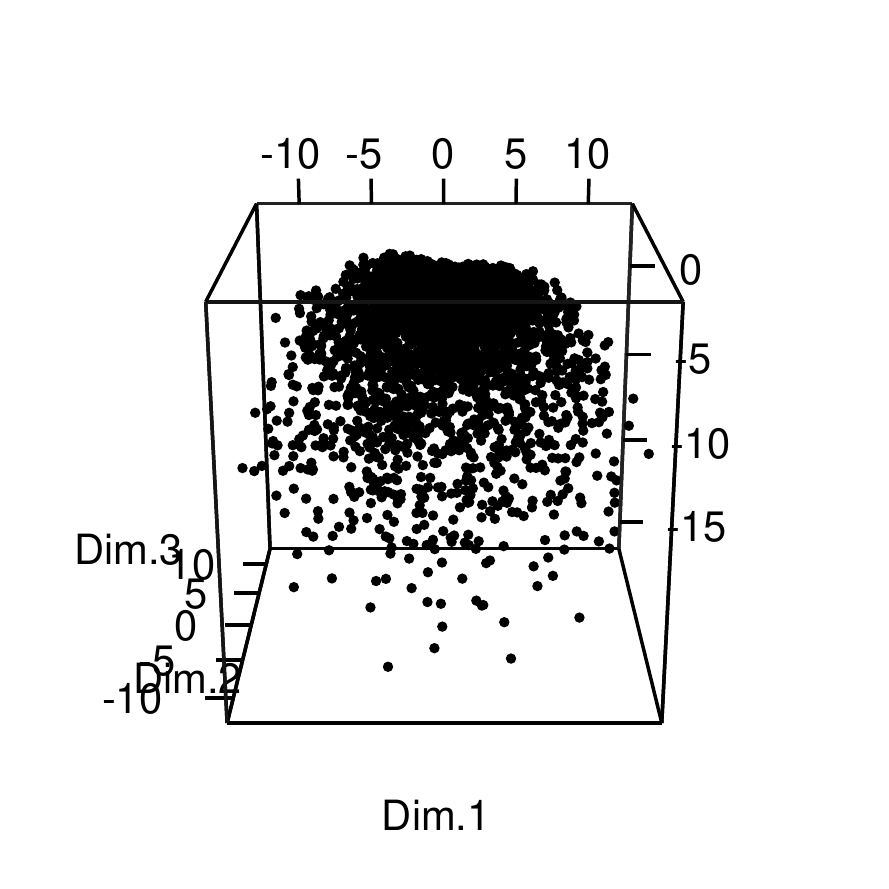}
	\includegraphics[width=0.32\textwidth]{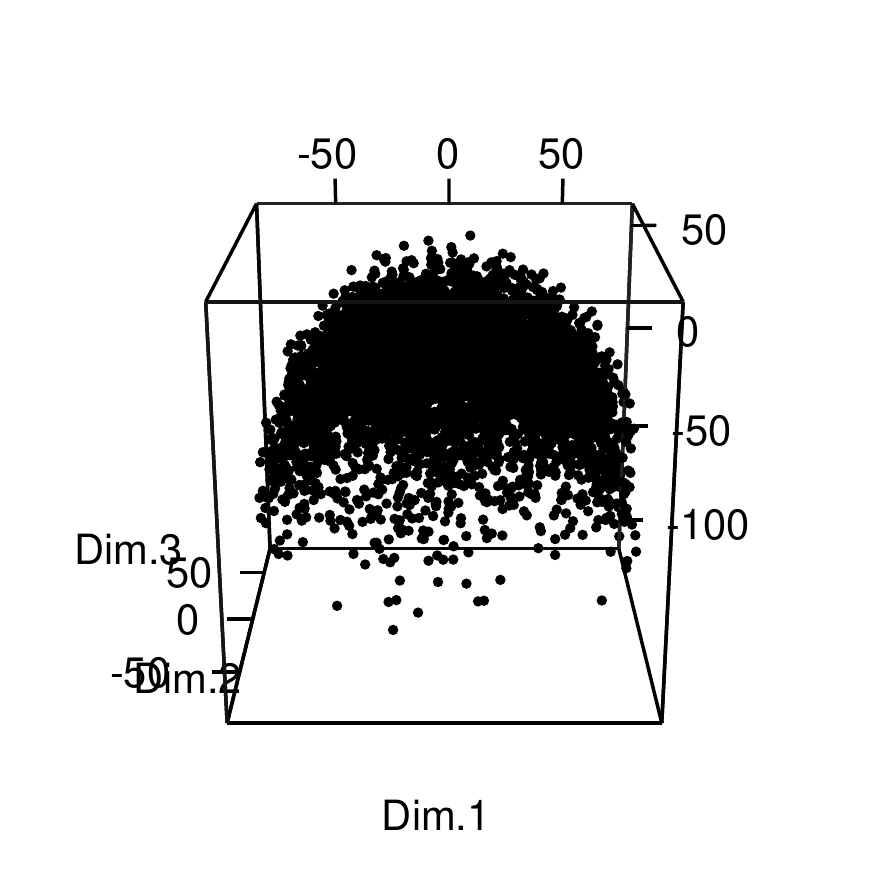}
	\includegraphics[width=0.32\textwidth]{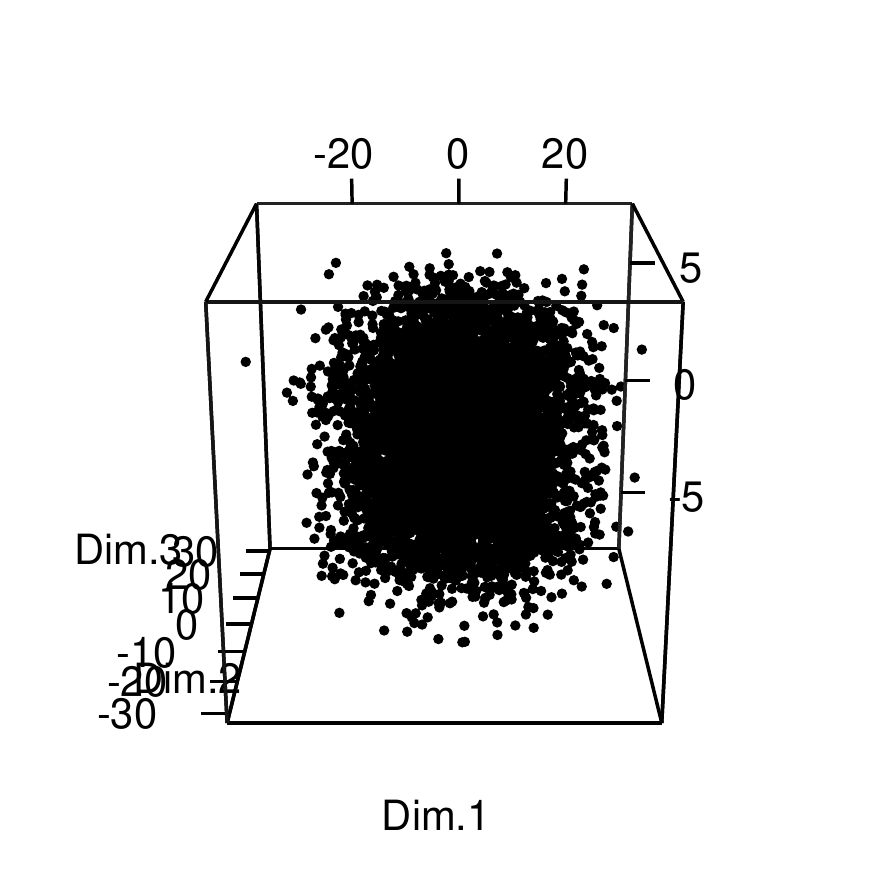}\\
	\includegraphics[width=0.32\textwidth]{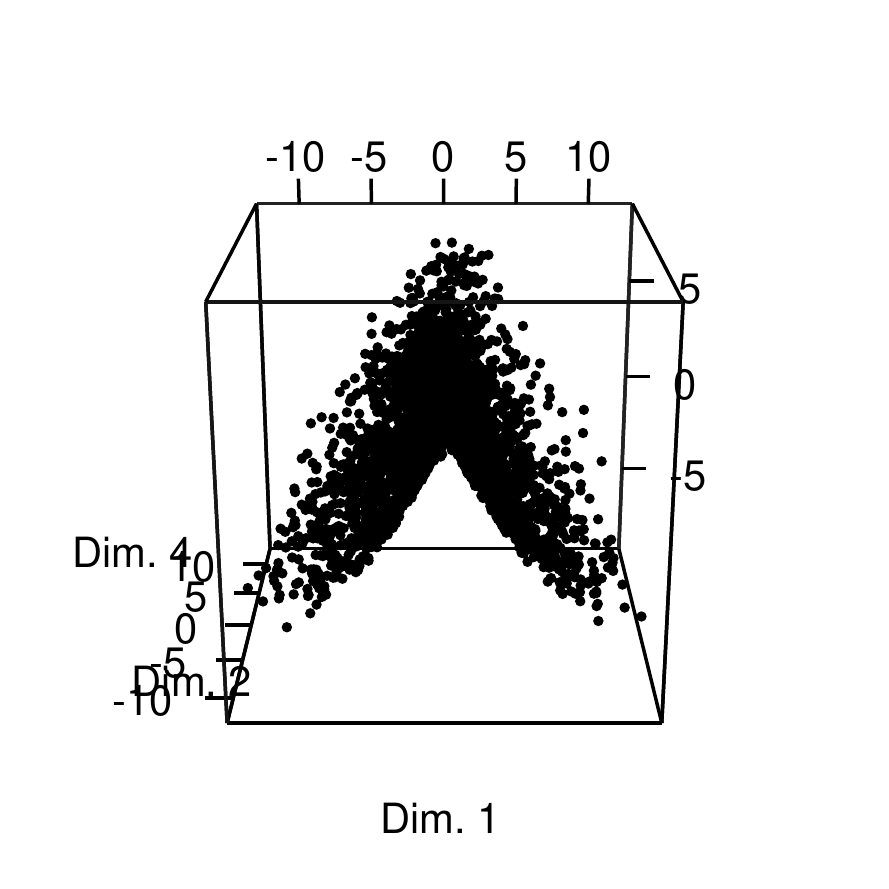}
	\includegraphics[width=0.32\textwidth]{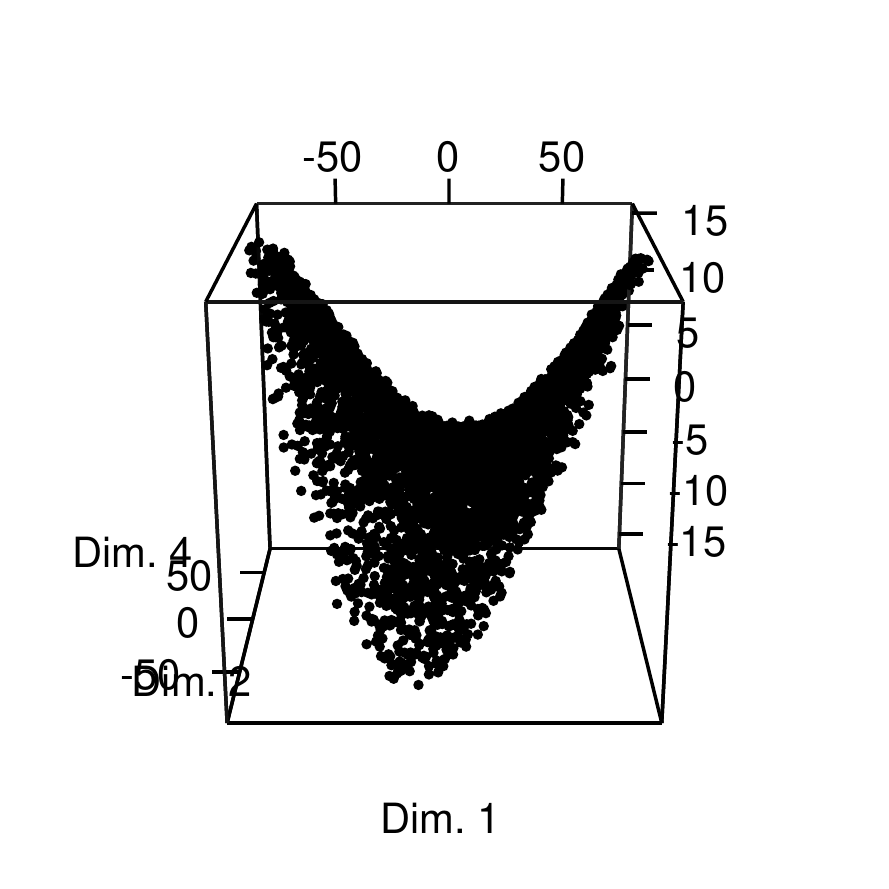}
	\includegraphics[width=0.32\textwidth]{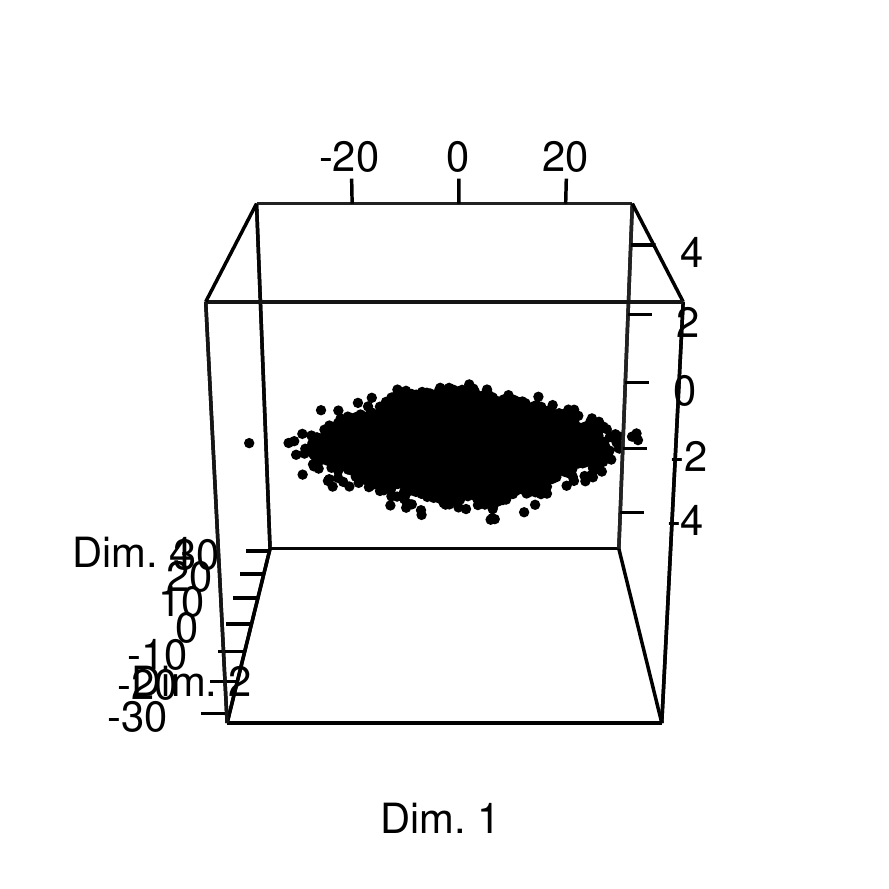}
	\caption{Four first eigencomponents of the $\pmb\alpha^{(i)}$'s in the Example \ref{ex:cercle_surparametre}, for the three different shape representations $\phi(\cdot)$. Left: characteristic function, middle: signed distance to the contour, right: discretization of the contour. 
The manifolds are shown in the $\{\v^1,\v^2,\v^3\}$ (top), and $\{\v^1,\v^2,\v^4\}$ bases (bottom). 
As can be seen from the two-dimensional surface in the $\{\v^1,\v^2,\v^4\}$ space when $\phi(\cdot)=\mathcal D$ (bottom right), the true dimension (3) is retrieved with the contour discretization. Note also that the associated manifold is convex.
	\label{fig:manifold_cercle_surparametre}
}
\end{figure}

\begin{table}[!ht]
	\centering
	\makebox[\textwidth][c]{
		\begin{tabu}{|c|c|c|c|c|c|c|}
			\hline
			& \multicolumn{2}{c|}{Characteristic function} & \multicolumn{2}{c|}{Signed Distance} & \multicolumn{2}{c|}{Discretization}\\\hline
			$j$ & Eigenvalue & Cumulative percentage & Eigenvalue & Cumulative percentage & Eigenvalue & Cumulative percentage\\\hline
			1 & 9.24 & 9.48 & 1238.53 & 40.24 & 109.04 & 49.23\\
			2 & 8.97 & 18.69 & 1210.72 & 79.57 & 104.69 & 96.50\\
			3 & 8.76 & 27.68 & 516.05 & 96.33 & 7.75 & 100\\
			4 & 5.95 & 33.79 & 39.70 & 97.62 & 0 & 100\\
			5 & 5.28 & 39.21 & 24.47 & 98.42 & 0 & 100\\
			6 & 3.93 & 43.25 & 21.83 & 99.13 & 0 & 100\\
			7 & 3.59 & 46.93 & 6.10 & 99.33 & 0 & 100\\
			8 & 3.36 & 50.38 & 6.03 & 99.52 & 0 & 100\\
			9 & 2.90 & 53.35 & 3.27 & 99.63 & 0 & 100\\
			10 & 2.80 & 56.23 & 3.12 & 99.73 & 0 & 100\\\hline
		\end{tabu}
	}
	\caption{10 first PCA eigenvalues for the different $\phi(\cdot)$'s, over-parameterized circle with $d=39$ parameters, with real dimension $d=3$.}
	\label{tab:eigenvalues_d39}
\end{table}\restoregeometry\clearpage

\begin{figure}[h!]
	\centering
	\includegraphics[width=0.6\textwidth]{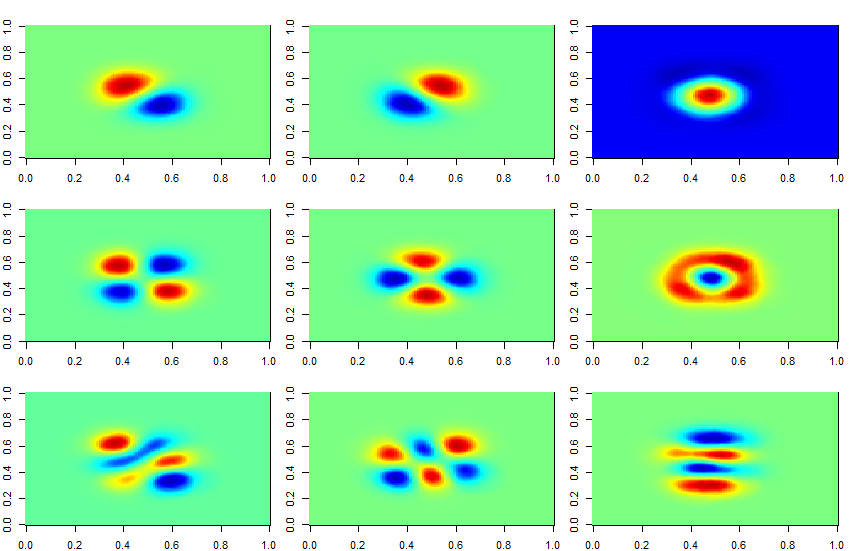}
	\caption{Example \ref{ex:cercle_surparametre}, over-parameterized circle with $d=39$ parameters, 9 first eigenvectors (left to right and top to bottom) when $\phi(\cdot)$ = characteristic function.}
	\label{fig:ex2_image}
\end{figure}

\begin{figure}[h!]
	\centering
	\includegraphics[width=0.6\textwidth]{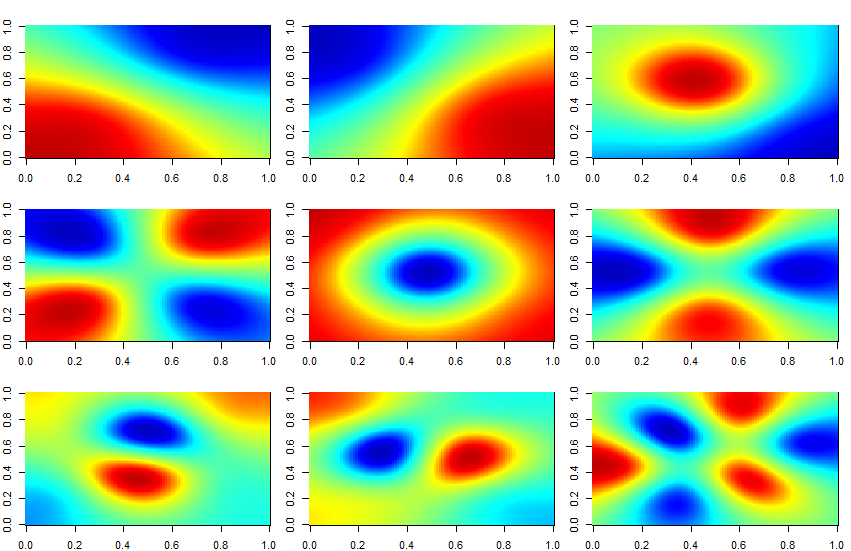}
	\caption{Example \ref{ex:cercle_surparametre}, over-parameterized circle with $d=39$ parameters, 9 first eigenvectors (left to right and top to bottom) when $\phi(\cdot)$ = signed distance.}
	\label{fig:ex2_signed_distance}
\end{figure}

\begin{figure}[h!]
	\centering
	\includegraphics[width=0.4\textwidth]{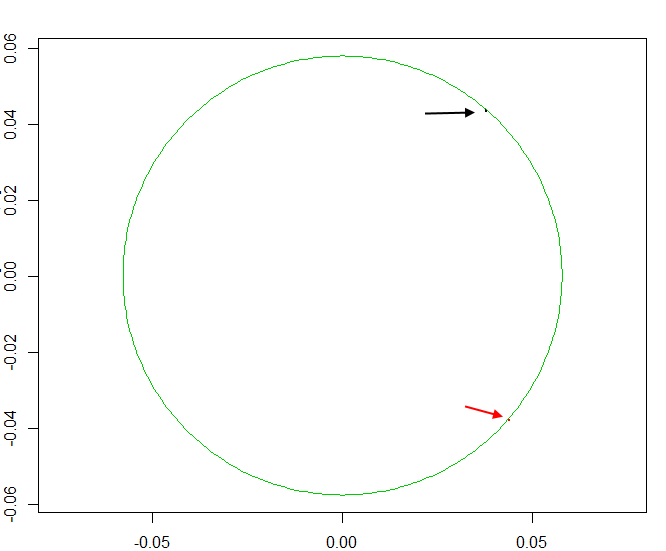}
	\caption{Example \ref{ex:cercle_surparametre}, over-parameterized circle with $d=39$ parameters, 3 first eigenvectors when $\phi(\cdot)$ = contour discretization.}
	\label{fig:ex2_pdm}
\end{figure}

The PCA eigenvalues for this example are given in Table \ref{tab:eigenvalues_d39} and are nearly the same as those in Table \ref{tab:eigenvalues_d3}. 
Apart from the little modification in the uniform distribution for sampling the $\x^{(i)}$'s which might lead to a slightly different $\pmb\Phi$, the over-parameterization is not a concern to retrieve the correct dimension.
Figures \ref{fig:ex2_image}-\ref{fig:ex2_pdm} show the 9 first eigenvectors (if they have a strictly positive eigenvalue) for the three $\phi(\cdot)$'s.

\clearpage

\begin{example}
	Three (non-overlapping) holes in $\R^2$, whose centers and radii are determined by $x_1$, $x_2$, $x_3$ (first circle), $x_4$, $x_5$, $x_6$ (second circle), and $x_7$, $x_8$, $x_9$ (third circle). This problem is more complex since it consists of three elements, and has $d=9$ dimensions. 
For $\phi(\cdot)=\mathcal D$, the discretization vector $\phi(\x)\in\R^D$ is split into 3 parts of size $D/3$ which correspond to the discretization of each circle.
	\label{ex:3cercles}
\end{example}
	
	\clearpage

\begin{table}[!ht]
	\centering
	\makebox[\textwidth][c]{
		\begin{tabu}{|c|c|c|c|c|c|c|}
			\hline
			& \multicolumn{2}{c|}{Characteristic function} & \multicolumn{2}{c|}{Signed Distance} & \multicolumn{2}{c|}{Discretization}\\\hline
			$j$ & Eigenvalue & Cumulative percentage & Eigenvalue & Cumulative percentage & Eigenvalue & Cumulative percentage\\\hline
			1 & 96.67 & 9.52 & 1785.93 & 31.51 & 154.26 & 19.06\\
			2 & 81.57 & 17.56 & 1267.81 & 53.88 & 151.80 & 37.82\\
			3 & 80.07 & 25.45 & 912.40 & 69.98 & 149.81 & 56.33\\
			4 & 66.03 & 31.96 & 588.30 & 80.36 & 148.09 & 74.63\\
			5 & 48.28 & 36.71 & 402.56 & 87.46 & 91.34 & 85.91\\
			6 & 40.66 & 40.72 & 159.38 & 90.27 & 90.53 & 97.10\\
			7 & 39.37 & 44.60 & 144.75 & 92.83 & 8.65 & 98.17\\
			8 & 38.75 & 48.42 & 121.80 & 94.97 & 8.54 & 99.22\\
			9 & 25.07 & 50.89 & 54.63 & 95.94 & 6.29 & 100\\
			10 & 24.45 & 53.30 & 47.36 & 96.77 & 0 & 100\\\hline
		\end{tabu}
	}
	\caption{10 first PCA eigenvalues for the different $\phi(\cdot)$'s, three circles with $d=9$ parameters.}
	\label{tab:eigenvalues_d9}
\end{table}
	
\begin{figure}[h!]
		\centering
		\includegraphics[width=0.6\textwidth]{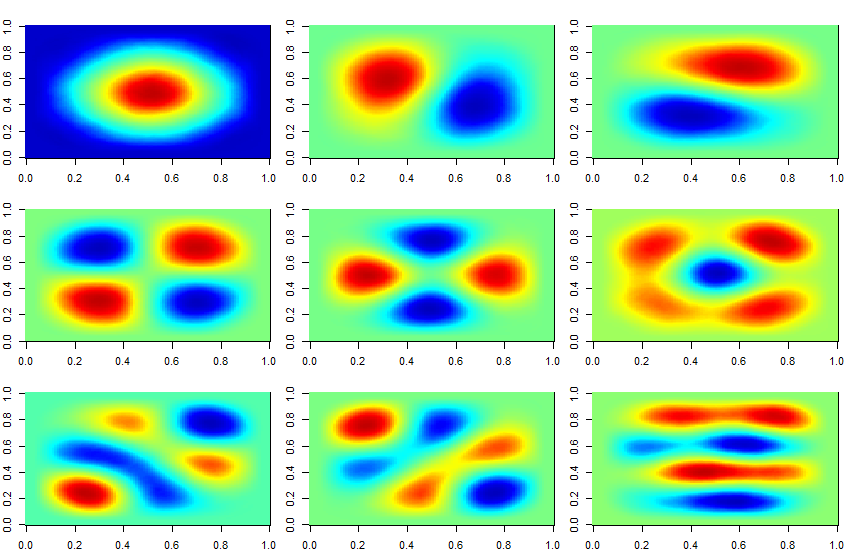}
		\caption{Example \ref{ex:3cercles}, three circles with $d=9$ parameters, 9 first eigenvectors (left to right and top to bottom) when $\phi(\cdot)$ = characteristic function.}
		\label{fig:ex3_image}
	\end{figure}

\begin{figure}[h!]
		\centering
		\includegraphics[width=0.6\textwidth]{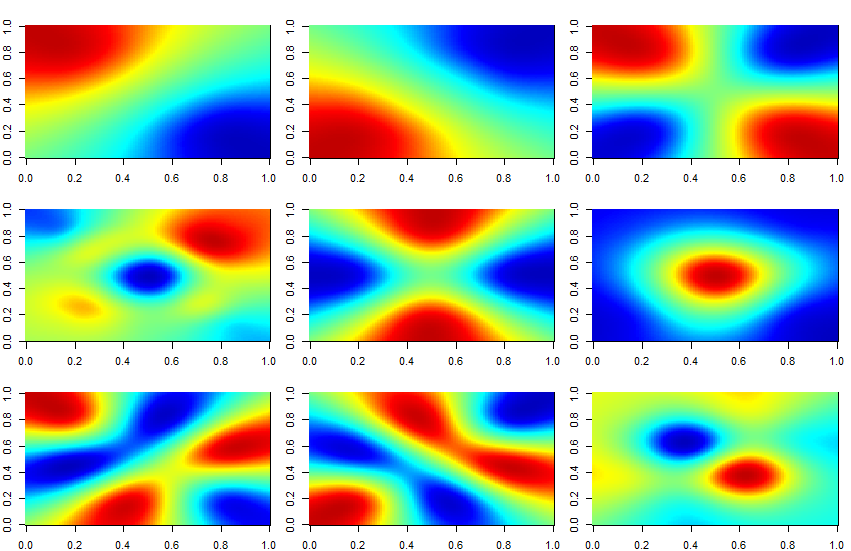}
		\caption{Example \ref{ex:3cercles}, three circles with $d=9$ parameters, 9 first eigenvectors (left to right and top to bottom) when $\phi(\cdot)$ = signed distance.}
		\label{fig:ex3_signed_distance}
	\end{figure}

\begin{figure}[h!]
		\centering
		\includegraphics[width=0.6\textwidth]{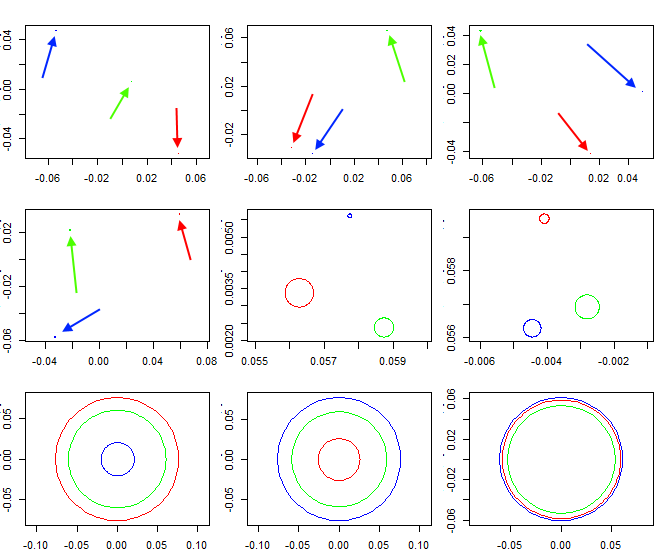}
		\caption{Example \ref{ex:3cercles}, three circles with $d=9$ parameters, 9 first eigenvectors (from left to right, top to bottom) when $\phi(\cdot)$ = discretization. The blue part of each eigenvector acts on the first circle, the red part of each eigenvector modifies the second circle and the green part of each eigenvector applies on the third circle.}
		\label{fig:ex3_pdm}
	\end{figure}

The 9 first eigenvectors are illustrated for the three $\phi(\cdot)$'s in Figures \ref{fig:ex3_image} to \ref{fig:ex3_pdm}.
	
\clearpage
	
In each example, for all $\phi(\cdot)$'s, any shape $\phi(\x^{(i)})$ can be reconstructed via Equation (\ref{eq:sumav}). 
$\pmb\alpha^{(i)}$ is nonetheless $D$-dimensional hence no dimension reduction is obtained. We are therefore interested in low-rank approximations $\pmb\phi_{1:\delta}:=\overline{\pmb\phi}+\sum_{j=1}^\delta\alpha_j\v^j$ which solely consider the $\delta$ first eigenvectors, while guaranteeing a sufficient precision. 
It is known \cite{jolliffe2011principal} that $\Vert\pmb\Phi-\pmb\Phi_{1:\delta}\Vert_F^2=N\sum_{j=\delta+1}^D\lambda_j$ where $\pmb\Phi_{1:\delta}$ is the reconstruction matrix using the $\delta$ first principal axes $\v^j$ only, and whose $i$-th row is $\overline{\pmb\phi}+\sum_{j=1}^\delta\alpha^{(i)}_j\v^j$. $\pmb\Phi_{1:\delta}$ is also known to be the closest (in terms of Frobenius norm) matrix to $\pmb\Phi$ with rank lower or equal to $\delta$. 
The $\lambda_j$'s with $j>\delta$ inform us about the reconstruction loss.
Hence, we look for a mapping $\phi(\cdot)$ for which the $\lambda_j$ quickly go to zero.
In Tables \ref{tab:eigenvalues_d1} to \ref{tab:eigenvalues_d9}, the vanishing of $\lambda_j$ beyond the intrinsic dimension only happens when $\phi(\cdot)=\mathcal D$.
With the other mappings, alternative techniques relying on local PCAs \cite{fukunaga1971algorithm} on the $\pmb\alpha^{(i)}$'s are required to estimate the dimensionality of manifolds such as the ones on the top row of Figure \ref{fig:manifold}. The $d$ first principal components, $\pmb\alpha^{(i)}_{1:d}$ suffice to reconstruct $\phi(\x^{(i)})$ exactly using $\mathcal D$ as the $\phi(\cdot)$ mapping, while more than $d$ components are required for $\phi(\x^{(i)})$ to be recovered using $\chi$ or $\mathbb D$.
With $\mathcal D$, the eigenvectors $\v^j$ (Right plot of Figure \ref{fig:ex1_d1_signed_distance}, Figures \ref{fig:ex1_d2_pdm}, \ref{fig:ex1_d3_pdm}, \ref{fig:ex2_pdm} and \ref{fig:ex3_pdm}) are physically meaningful: 
they can be interpreted as shape discretizations, which, being multiplied by coefficients $\alpha_j$ and added to the mean shape $\overline{\pmb\phi}$, act on the hole's size (Eigenvector 1 in right plot of Figure \ref{fig:ex1_d1_signed_distance}, Eigenvector 2 in Figure \ref{fig:ex1_d2_pdm}, Eigenvector 3 in Figure \ref{fig:ex1_d3_pdm}, Eigenvector 3 in Figure \ref{fig:ex2_pdm}, Eigenvectors 7-9 in Figure \ref{fig:ex3_pdm}), or on the hole's position (Eigenvector 1 in Figure \ref{fig:ex1_d2_pdm}, Eigenvectors 1-2 in Figure \ref{fig:ex1_d3_pdm}, Eigenvectors 1-2 in Figure \ref{fig:ex2_pdm}, Eigenvectors 1-6 in Figure \ref{fig:ex3_pdm}). 
For example, very small eigenvectors such as the first one in Figure \ref{fig:ex1_d2_pdm} displace the shape in the direction specified by the eigenvector's position. In Figure \ref{fig:ex3_pdm}, the first eigenvectors tend to move each circle with respect to each other, while the sizes of the holes are affected by the last eigenvectors. 
Whereas the characteristic function $\chi$ and the signed distance $\mathbb D$ are images, 
the mapping $\mathcal D$ is a discretization of the final object we represent, a contour shape.
Without formal proof, we think that this is related to the observed property that the $d$ (the number of intrinsic dimensions) first eigencomponents $\pmb\alpha_{1:d}^{(i)}$, $i=1,\dotsc,N$ make a convex set as can be seen in Figures \ref{fig:manifold} and \ref{fig:manifold_cercle_surparametre}.

In a solid mechanics analogy, the $\overline{\pmb\phi}+\sum_j\alpha_j\v^j$ reconstruction can be thought as a sum of pressure fields $\v^j$ applied on each node of the Point Distribution Model, and which deform the initial mean shape $\overline{\pmb\phi}$ by a magnitude $\alpha_j$ to obtain $\pmb\phi$. Such an interpretation cannot be conducted with the eigenvectors obtained via the $\chi$ or $\mathbb D$ mapping, shown in the other figures.

Because of its clear pre-eminence, in the following, we will only consider the $\pmb\alpha$'s obtained using the contour discretization as $\phi(\cdot)$ mapping. 

\subsubsection{Hierarchic shape basis for the reduction of high-dimensional designs}

Following these observations, we now deal with slightly more complex and realistic shapes $\Omega_\x$. Even though they are initially described with many parameters, they mainly depend on few intrinsic dimensions.

\begin{figure}[h!]
		\centering
		\includegraphics[width=0.6\textwidth]{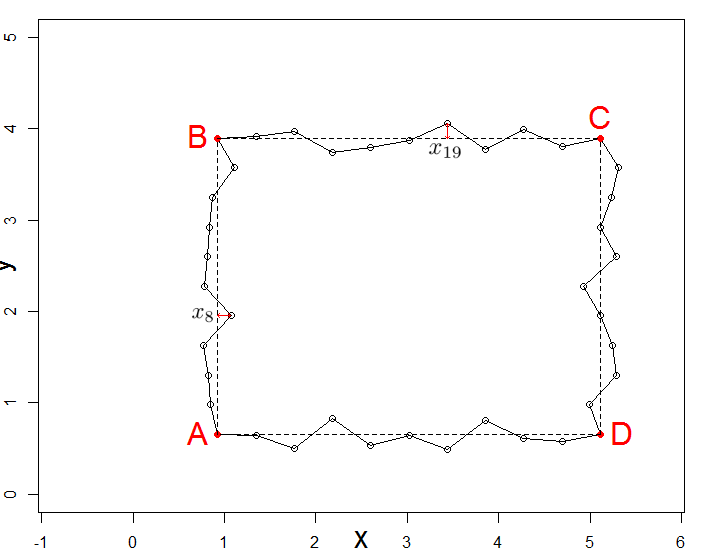}
		\caption{Example \ref{ex:coeur}: a rectangle with varying position, size, and deformation of its sides.}
		\label{fig:rectangle}
	\end{figure}

\begin{example}
	A rectangle ABCD with $\x\in\R^{40}$ whose parameters $x_1$ and $x_2$ are the location of A, $x_3$ and $x_4$ are the width and the height of ABCD, and $\x_{5:13}$, $\x_{14:22}$, $\x_{23:31}$ and $\x_{32:40}$ are small evenly distributed perturbations, on the AB, BC, CD and DA segments, respectively. 
	\label{ex:coeur}
\end{example}
$x_1,\dotsc,x_4$ are of a magnitude larger than the other parameters to ensure a close-to-rectangular shape, as shown in Figure \ref{fig:rectangle}.
	
\begin{table}[!ht]
	\centering
	\makebox[\textwidth][c]{
		\begin{tabular}{|c|c|c|}
			\hline
			$j$ & Eigenvalue & Cumulative percentage\\\hline
			1 & 867.65 & 48.73\\
			2 & 866.90 & 97.42\\
			3 & 21.46 & 98.62\\
			4 & 21.43 & 99.83\\
			5 & 0.13 & 99.83\\
			6 & 0.13 & 99.84\\
			7 & 0.13 & 99.85\\
			8 & 0.13 & 99.86\\
			9 & 0.12 & 99.86\\
			10 & 0.12 & 99.87\\
			\vdots & \vdots & \vdots\\
			39 & 0.04 & 99.99\\
			40 & 0.04 & 100\\
			41 & 0 & 100\\\hline
		\end{tabular}
	}
	\caption{First PCA eigenvalues for $\phi(\cdot)$ = discretization, rectangles with $d=40$ parameters (Example \ref{ex:coeur}).}
	\label{tab:eigenvalues_rectangle}
\end{table}
\clearpage

In this example where 4 parameters (position and sizes) mainly explain the differences among shapes, we see that a reconstruction quality of 99.83\% is attained with the 4 first eigenvectors $\v^j$.

\begin{figure}[h!]
	\centering
	\includegraphics[width=0.6\textwidth]{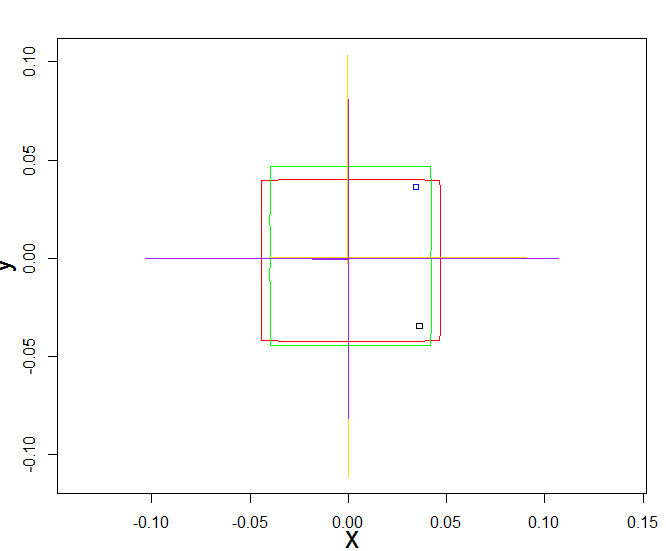}
	\caption{6 first eigenshapes (in the order black, blue, red, green, yellow, purple) of the rectangles in Example \ref{ex:coeur}.}
	\label{fig:coeur_eigenshapes}
\end{figure}

Figure \ref{fig:coeur_eigenshapes} details the eigenvectors.
$\v^{1}$ and $\v^{2}$, the most influencing eigenshapes plotted in black and blue act as translations, 
while $\v^{3}$ and $\v^{4}$ (in red and green) correspond to widening and heightening of the rectangle.
The fluctuations along the segments appear from the 5th eigenshape on.
Any shape is retrieved with the $d=40$ first eigenshapes which corresponds to the total number of parameters.

	\begin{example}
		A straight line joining two fixed points A and B, modified by smooth perturbations $\mathbf r\in\R^{29}$, evenly distributed along [AB] to approximate a smooth curve.
	\label{ex:catenoide}
	\end{example}
The fifth example is inspired by the catenoid problem \cite{Colding11106}. The perturbations $\mathbf r$ are generated by a Gaussian Process with squared exponential kernel and with length-scale 6 times smaller than [AB]. Therefore, in this example, the $N=5000$ $\mathbf r^{(i)}$'s used for building $\pmb\Phi$ are not uniformly distributed in $X$.
	
	\begin{figure}[h!]
		\centering
		\includegraphics[width=0.6\textwidth]{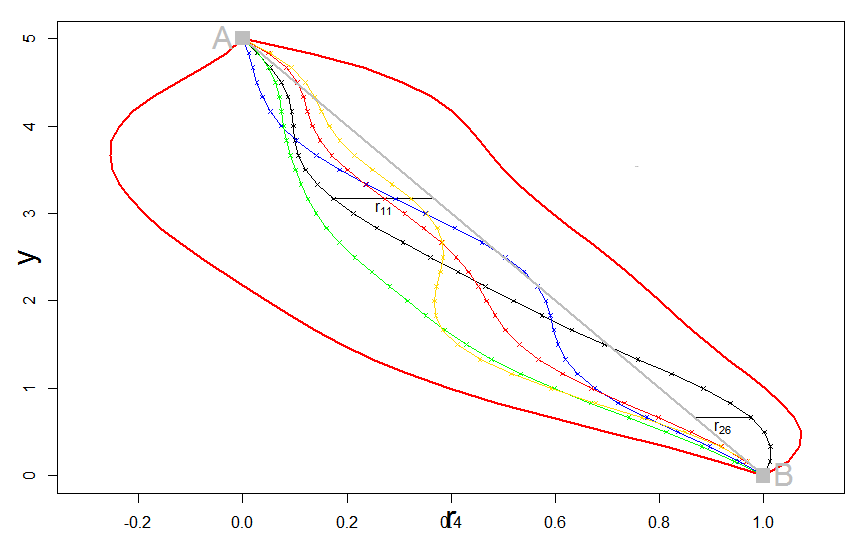}
		\caption{Example \ref{ex:catenoide}: a straight line joining two points, modified by the perturbations $r_j$ to approximate a curve. Gray: the line joining A and B. Blue, red, yellow and green curve: examples of lines with regular $r_j$ perturbations. Red envelope: boundaries for the $r_j$'s.
}
		\label{fig:catenoid_problem_description}
	\end{figure}

\begin{table}[!ht]
	\centering
	\makebox[\textwidth][c]{
		\begin{tabular}{|c|c|c|}
			\hline
			$j$ & Eigenvalue & Cumulative percentage\\\hline
			1 & 2.156 & 50.258\\
			2 & 1.251 & 79.422\\
			3 & 0.590 & 93.181\\
			4 & 0.206 & 97.973\\
			5 & 0.065 & 99.480\\
			6 & 0.017 & 99.882\\
			7 & 0.004 & 99.975\\
			8 & 0.001 & 99.995\\
			9 & $\varepsilon$ & 99.999\\
			10 & $\varepsilon$ & 100\\
			\vdots & \vdots & \vdots\\
			28 & $\varepsilon$ & 100\\
			29 & $\varepsilon$ & 100\\
			30 & 0 & 100\\\hline
		\end{tabular}
	}
	\caption{First PCA eigenvalues for $\phi(\cdot)$ = discretization, curve with $d=29$ parameters. $\varepsilon$ means the quantity is not exactly 0, but smaller than $10^{-3}$, hence less than 0.04\% of the first PCA eigenvalue.}
	\label{tab:eigenvalues_catenoide}
\end{table}
\clearpage

Again, the initial dimension $(d=29)$ is recovered by looking at the strictly positive eigenvalues. 
Furthermore, the manifold is found to mainly lie in a lower dimensional space: $\mathcal A_N$ can approximated in $\delta=7$ dimensions since $\sum_{j=1}^\delta\lambda_j/\sum_{j=1}^D\lambda_j=99.975\%$.

\begin{figure}[h!]
	\centering
	\includegraphics[width=0.6\textwidth]{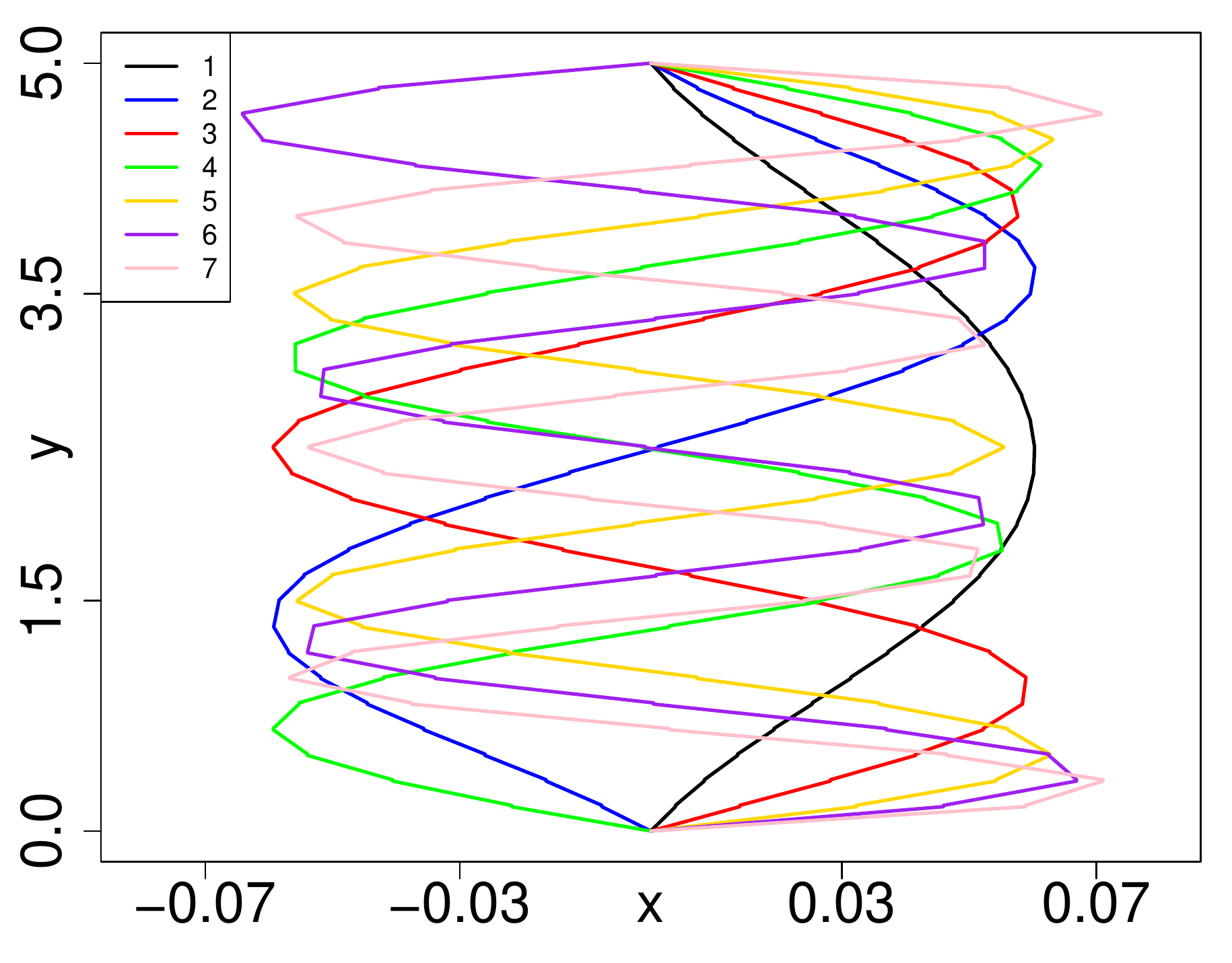}
	\caption{7 first eigenshapes for the curves of Example \ref{ex:catenoide}.}
	\label{fig:catenoide_eigenshapes}
\end{figure}

Figure \ref{fig:catenoide_eigenshapes} shows the corresponding eigenshapes. 
The eigenshapes are similar to the ordered modes of the harmonic series with the associated eigenvalues ordered as the inverse of the frequencies.

\begin{figure}[h!]
	\centering
	\includegraphics[width=0.6\textwidth]{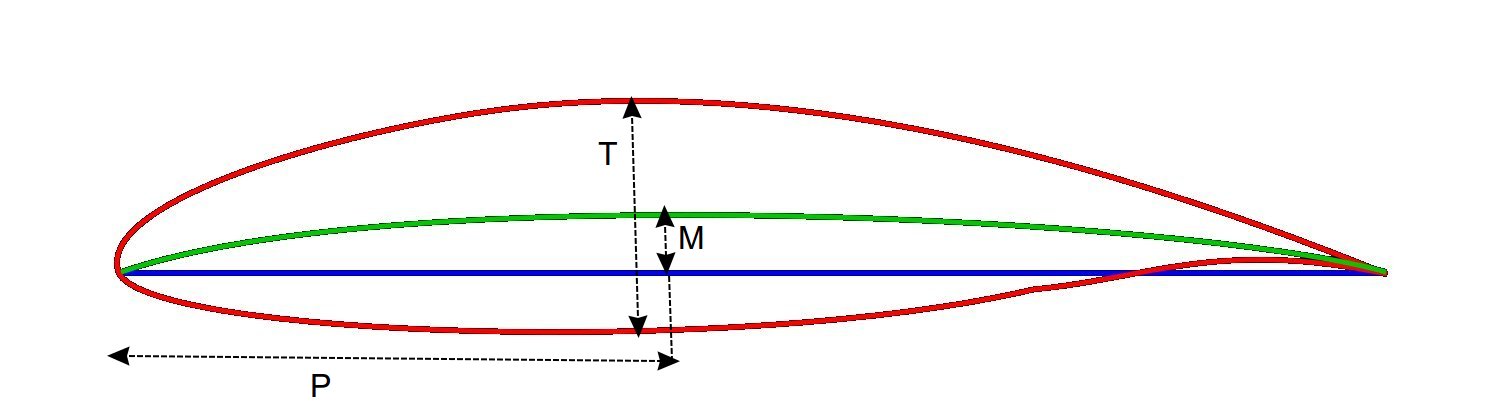}
	\caption{Description of a NACA airfoil with its $M$, $P$, $T$ parameters.}
	\label{fig:naca_description}
\end{figure}

\begin{example}
	A NACA airfoil parameterized by three parameters: $\x=(M,P,T)^\top\in\R^3$ where $M$ is the maximum  camber, $P$ is the position of this maximum, and $T$ is the maximal thickness. Figure \ref{fig:naca_description} describes the airfoil.
	\label{ex:naca3}
\end{example}

\begin{table}[!ht]
	\centering
	\makebox[\textwidth][c]{
		\begin{tabular}{|c|c|c|}
			\hline
			$j$ & Eigenvalue & Cumulative percentage\\\hline
			1 & 0.2819 & 54.619\\
			2 & 0.2203 & 97.318\\
			3 & 0.0129 & 99.814\\
			4 & 0.0008 & 99.959\\
			5 & 0.0001 & 99.983\\
			6 & $\varepsilon$ & 99.991\\
			7 & $\varepsilon$ & 99.996\\
			8 & $\varepsilon$ & 99.997\\
			9 & $\varepsilon$ & 99.999\\
			10 & $\varepsilon$ & 99.999\\\hline
		\end{tabular}
	}
	\caption{First PCA eigenvalues of the NACA airfoil with $d=3$ parameters ($\phi(\cdot)$ is the contour discretization). $\varepsilon$ means the quantity is smaller than $10^{-4}$, hence less than 0.04\% of the first PCA eigenvalue.}
	\label{tab:eigenvalues_naca3}
\end{table}

In this example, a typical noise-truncation criterion such as discussed in Example \ref{ex:catenoide} would retain 3 or 4 axes. In Example \ref{ex:naca3} too, the effective dimension can almost be retrieved from the $\lambda$'s.

\begin{figure}[h!]
	\centering
	\includegraphics[width=0.57\textwidth]{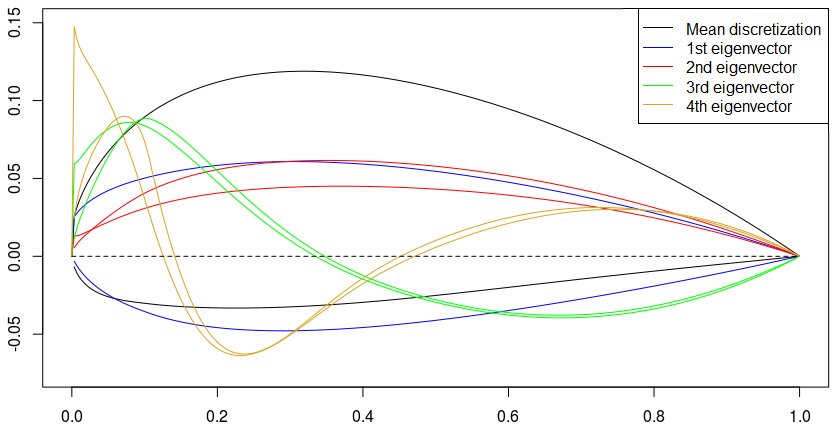}
	\includegraphics[width=0.4\textwidth]{naca22_alpha_manifold.png}
	\caption{NACA airfoil with $d=3$ parameters. Left: mean shape and 4 first eigenshapes (black, blue, red, green, yellow). Right: three first eigencomponents ($\alpha_1,\alpha_2,\alpha_3$) of the $\mathcal A_N$ manifold.}
	\label{fig:naca3_eigenshapes}
\end{figure}

Figure \ref{fig:naca3_eigenshapes} shows the 4 first eigenshapes (left) as well as the $\mathcal A_N$ manifold (right). The eigenvectors can be interpreted as a reformulation of the CAD parameters. 
The first eigenshape (blue) is a symmetric airfoil. Multiplying it by a coefficient (after adding it to the black mean shape) will increase or decrease the thickness of the airfoil, hence it plays a similar role to the $T$ parameter. The second eigenshape is a cambered airfoil, whose role is similar to $M$ (maximum camber). Last, the third airfoil, which has a much smaller eigenvalue $\lambda_3$, is very thin, positive in the first part of the airfoil, and negative in its second part. It balances the camber of the airfoil towards the leading edge or towards the rear and plays a role similar to $P$, the position of the maximum camber. $\v^3$'s effect is complemented by $\v^4$.

The analysis of $\mathcal A_N$ (Figure \ref{fig:naca3_eigenshapes}) is physically meaningful: even though $\x^{(i)}$ are sampled uniformly in $X$, $\mathcal A_N$ resembles a pyramid in the ($\v^1,\v^2,\v^3$) basis. Designs with minimal $\alpha_2$ share the same $\alpha_3$ value. Since negative $\alpha_2$'s correspond to wings with little camber, the position of this maximum camber has very little impact, hence the almost null $\alpha_3$ value. 
By looking at $\mathcal A_N$, it is learned that the parameter $P$ does not matter when $M$ is small, which is intuitive but is not expressed by the $(M,P,T)$ coordinates. 
Distances in $\mathcal A_N$ are therefore more representative of shape differences. 
An additional advantage of analyzing shapes is that correlations in the space of parameters (such as the one between $M$ and $P$ in this example) are discovered and removed, since $\mathcal V$ is an orthonormal basis. Here, orthogonality between eigenshapes is measured by the standard scalar product in $\R^D$. Depending on the application, there may exist natural definitions of the orthogonality between discretized shapes, which could be used by the PCA.

\begin{figure}[h!]
	\centering
	\includegraphics[width=0.5\textwidth]{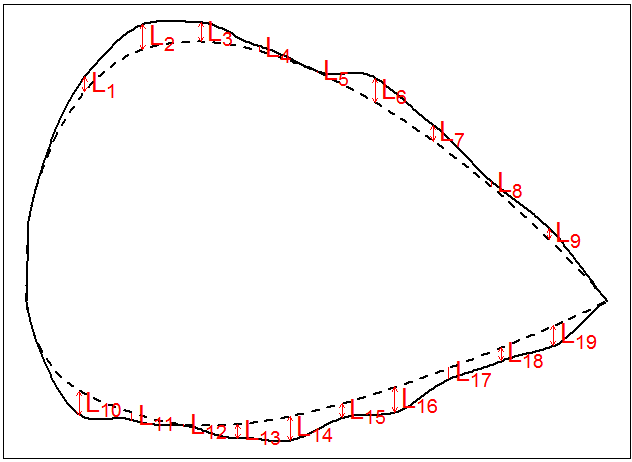}
	\caption{Description of a NACA airfoil in 22 dimensions. It is a standard NACA airfoil whose intrados and extrados have been modified by bumps of size $L_i$.}
	\label{fig:naca22_description}
\end{figure}

\begin{example}
	A modified NACA airfoil which is parameterized by $d=22$ parameters:\\$\x=(M,P,T,L_1,\dotsc,L_{19})^\top\in\R^{22}$ where $M$, $P$, $T$ are the standard NACA parameters (Example \ref{ex:naca3}), and where the $L_i$'s correspond to small bumps along the airfoil. Figure \ref{fig:naca22_description} describes a NACA 22 airfoil.
	\label{ex:naca22}
\end{example}

\clearpage

\begin{table}[!ht]
	\centering
	\makebox[\textwidth][c]{
		\begin{tabular}{|c|c|c|}
			\hline
			$j$ & Eigenvalue & Cumulative percentage\\\hline
			1 & 0.2826 & 53.932\\
			2 & 0.2205 & 96.021\\
			3 & 0.0134 & 98.580\\
			4 & 0.0011 & 98.798\\
			5 & 0.0006 & 98.903\\
			6 & 0.0005 & 99.006\\
			7 & 0.0005 & 99.106\\
			8 & 0.0005 & 99.202\\
			9 & 0.0005 & 99.293\\
			10 & 0.0004 & 99.377\\
			\vdots & \vdots & \vdots\\
			19 & 0.003 & 99.958\\
			20 & 0.002 & 99.992\\
			21 & $\varepsilon$ & 99.995\\
			22 & $\varepsilon$ & 99.998\\
			23 & $\varepsilon$ & 99.999\\\hline
		\end{tabular}
	}
	\caption{First PCA eigenvalues for $\phi(\cdot)$ = discretization, NACA with $d=22$ parameters. $\varepsilon$ means the quantity is not exactly 0, but smaller than $10^{-4}$, hence less than 0.04\% of the first PCA eigenvalue.}
	\label{tab:eigenvalues_naca22}
\end{table}

\begin{figure}[h!]
	\centering
	\includegraphics[width=0.37\textwidth]{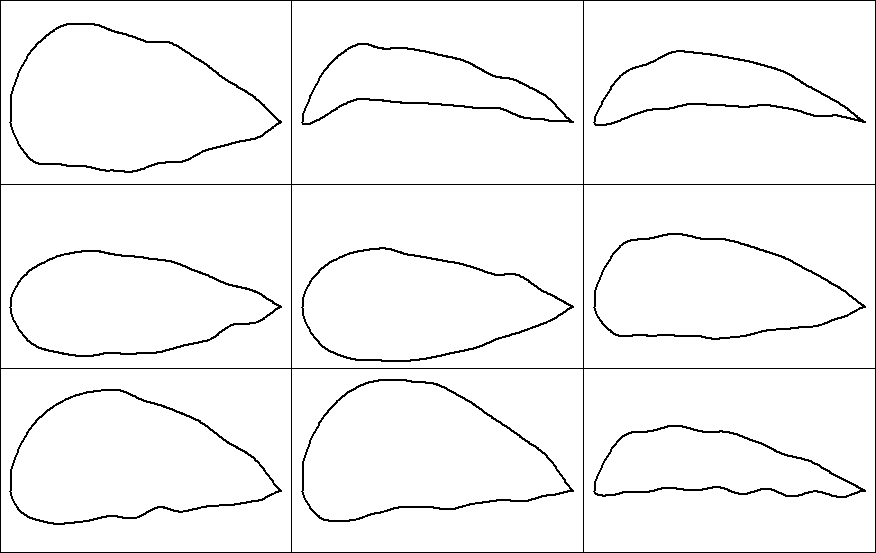}
	\includegraphics[width=0.6\textwidth]{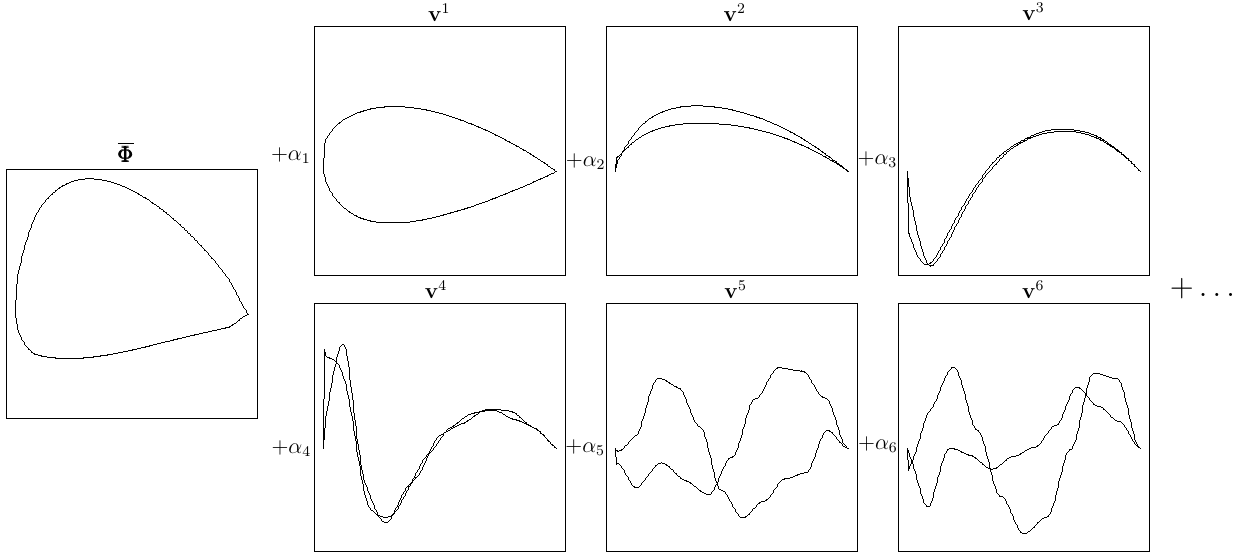}\\
	\includegraphics[width=0.5\textwidth]{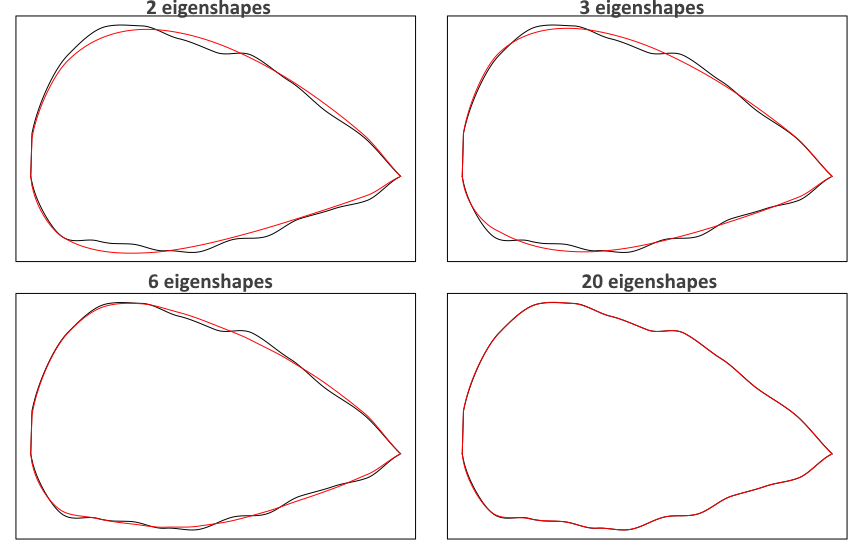}
	\caption{
Left: examples of NACA 22 airfoils. Even though the true dimension is 22, less dimensions may suffice to approximate the shapes well enough. Right: reconstruction scheme of any NACA 22 shape: a weighted deviation from the mean shape $\overline{\pmb\phi}$ in the direction of the eigenshapes. Bottom: example of shape reconstruction (red) using 2, 3, 6 or 20 eigenshapes. The more $\v^j$'s, the better the reconstruction but the larger the dimension of $\pmb\alpha$.}
	\label{fig:exemple_naca22}
\end{figure}

Here, as in the Example \ref{ex:naca3}, the noise-truncation criteria will retain between 6 and 20 dimensions, depending on the reconstruction quality required. 
Indeed, when looking at specimen of NACA 22 airfoils as the one in the upper left part of Figure \ref{fig:exemple_naca22}, less than 22 dimensions are expected to be necessary to retrieve an approximation of sufficient quality.

\begin{figure}[h!]
	\centering
	\includegraphics[width=0.6\textwidth]{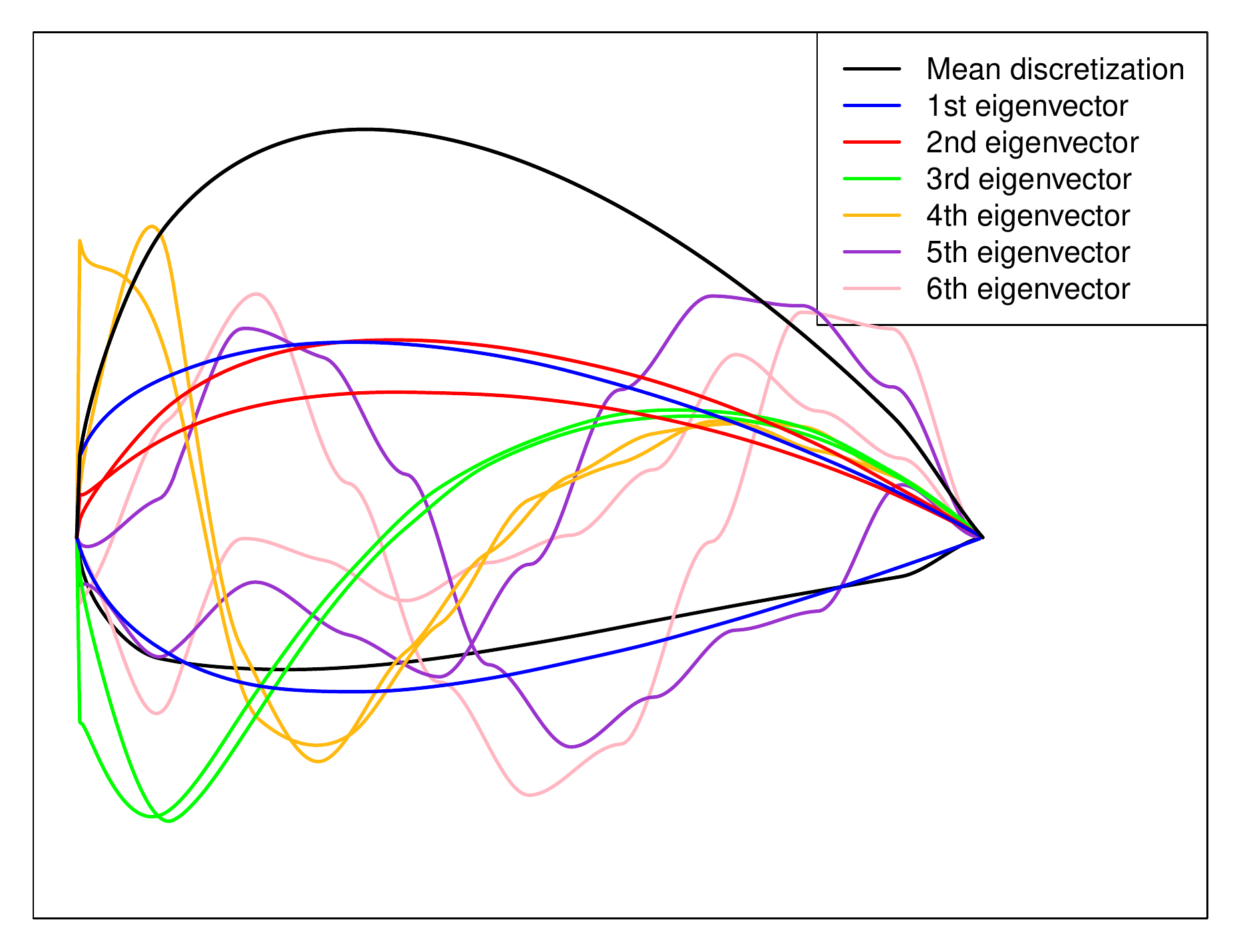}
	\caption{Mean shape (black) and 6 first eigenshapes (blue, red, green, yellow, purple, pink) for the NACA with 22 parameters. 
The three first eigenvectors are similar to those observed on Figure \ref{fig:naca3_eigenshapes} for the original NACA 3. Fluctuations along the eigenshapes are found from the 4th eigenshape on. 
They allow to reconstruct the local refinements (bumps) of the airfoils.}
	\label{fig:naca22_eigenshapes}
\end{figure}

The analysis of eigenshapes, shown in Figure \ref{fig:naca22_eigenshapes}, is similar to the one of Example \ref{ex:naca3}. Small details that act on the airfoil such as the bumps only appear from the 4th eigenshape on. Not taking them into account leads to a weaker reconstruction, as shown in the bottom part of Figure \ref{fig:exemple_naca22}.

\clearpage

According to these experiments, the eigenvectors $\v^j$, $j\in\{d+1,\dotsc,D\}$, can already be discarded without even considering the values of the associated objective functions since the $d$ first shape modes explain the whole variability of the discretized shapes. 
In practice, to filter numerical noise and to remove non-informative modes in shapes that are truly over-parameterized, we only consider the $d'$ first eigenshapes, $d':=\min(d,\tilde d)$ where $\tilde d$ corresponds to the smallest number of axes that explain more than a given level of diversity in $\Phi$ (e.g. 99.9, 99.95 or 99.99\%), measured by $100\times\sum_{j=1}^{\tilde d}\lambda_j/\sum_{j=1}^D\lambda_j$. Another alternative is to define $\tilde d$ according to the dimensions for which $\lambda_j/\lambda_1$ is smaller than a prescribed threshold (e.g. 1/1000). 
Even though the notation $D$ is kept, the eigenvectors $\v^j$ and the principal components $\alpha_j$, are considered to be null $\forall j>d'$ so that in fact $D=d'$ in the following.
	
\section{GP models for reduced eigenspaces}
\label{section:GP_in_eigenbasis}
Building a surrogate model in the space of principal components has already been investigated in the context of reduced order models \cite{berkooz1993proper}.
In most applications, the dimension reduction is carried out in the output space, which has large dimension when it corresponds to values on a finite element mesh. 
The response is approximated by a linear combination of a small number of modes, and the metamodel is a function of the modes coefficients.
The construction of surrogates with inherent dimensionality reduction has also been considered. 
In the active subspace method \cite{constantine2014active}, the dimension reduction comes from a linear combination of the inputs which is carried out by projecting $\x$ onto the hyperplane spanned by the directions of largest $\nabla f(\x)$ variation. The reduced-dimension GP is then $Y(\mathbf W^\top\x)$ with $\mathbf W\in\R^{d\times\delta}$ containing these directions in columns. In \cite{palar2018accuracy}, cross-validation is employed for choosing the number of such axes.
An application to airfoils is given in \cite{li2019surrogate} where the authors take the directions of largest drag and lift gradients as columns of $\mathbf W$, even though this basis is no longer orthogonal. 
Another related technique with a $Y(\mathbf W^\top\x)$ GP which does not require the knowledge of $\nabla f(\x)$ is the Kriging and Partial Least Squares (KPLS) method \cite{bouhlel2016improving}, where $\x$ is projected onto the hyperplane spanned by the first $\delta$ axes of a PLS regression \cite{frank1993statistical}. The dimension reduction is output-driven but $\mathbf W$ is no longer orthogonal, and information may be lost when $n<d'$ because any shape (of effective dimension $d'$) cannot be exactly reconstructed (Equation \ref{eq:sumav}) with these $n$ vectors. Coordinates in the PLS space are therefore incomplete and metamodeling loses precision when $n$ is too small.
In the same spirit, a double maximum-likelihood procedure is developed in \cite{tripathy2016gaussian} to build an output-related and orthogonal matrix $\mathbf W$ for the construction of a Gaussian Process with built-in dimensionality reduction. Rotating the design space through hyperparameters determined by maximum likelihood is also performed in \cite{namura2017kriging}. Table \ref{tab:built_in_gp} summarizes the existing literature for building such GPs as well as the approach introduced in Section \ref{sec:additive_gp} (last column).

\begin{table}[!ht]
	\centering
	\makebox[\textwidth][c]{
		\begin{tabular}{|c|c|c|c|}
			\hline
			Model & $Y(\mathbf W^\top\x)$ & $Y(\mathbf W^\top\phi(\x))$ & $Y^a(\mathbf W_a^\top\phi(\x))+Y^{\overline{a}}(\mathbf W_{\overline{a}}^\top\phi(\x))$\\\hline
			Dimension reduction & Linear in $\x$ & Nonlinear in $\x$ & \makecell{Nonlinear in $\x$;\\group-additive model}\\\hline
			Construction of $\mathbf W$ & \makecell{Active subspaces \cite{constantine2014active,palar2018accuracy,li2019surrogate}\\PLS \cite{bouhlel2016improving}\\GP hyperparameters \cite{tripathy2016gaussian,namura2017kriging}\\Sensitivity analysis \cite{salem2018sequential}} & PLS \cite{li2018data} & \makecell{Selection of mapped variables\\through penalized likelihood\\(Section \ref{sec:eigenshape_selection})}\\\hline
		\end{tabular}
	}
	\caption{An overview of GP models with built-in dimensionality reduction.}
	\label{tab:built_in_gp}
\end{table}

\subsection{Unsupervised dimension reduction}	
\label{sec:unsupervised_dimension_reduction}
Instead of the space of CAD parameters $\x$, we reduce the dimension of the input space by building the surrogate with information from the space of shape representations, $\Phi$, as in \cite{li2018data}. To circumvent the high dimensionality of $\Phi\subset\R^D$, a linear dimension reduction of $\phi(\x)$ is achieved by building the model in the space spanned by $\mathbf W^\top\phi(\x)$. 
A natural candidate for $\mathbf W$ is a restriction to few columns (eigenshapes) of the matrix $\mathbf V$. 
Notice that contrarily to the other dimension reduction techniques which operate a linear dimension reduction of $\x$, 
this approach is nonlinear in $\x$ since it operates linearly on the nonlinear transformation $\phi(\x)$.
Also, it operates on a better suited representation of the designs, their shapes, instead of their parameters.

A first idea to reduce the dimension of the problem is to conserve the $\delta$ first eigenvectors $\v^j$ according to some reconstruction quality criterion measured by the eigenvalues. 
Given a threshold $T$ (e.g., 0.95 or 0.99), only the first $\delta$ modes such that $\frac{\sum_{j=1}^{\delta}\lambda_j}{\sum_{j=1}^{D}\lambda_j}>T$ are retained in $\mathbf V_{1:\delta}\in\R^{D\times\delta}$ because they contribute for $100\times T\%$ of the variance in $\Phi$.
The surrogate model is implemented in the space of the $\delta$ first principal components as
\begin{equation}
Y(\pmb\alpha_{1:\delta})=Y(\mathbf V_{1:\delta}^\top(\phi(\x)-\overline{\pmb\phi})).
\label{eq:Yalpha}
\end{equation}
Using a stationary kernel for the $Y(\pmb\alpha_{1:\delta})$ GP, i.e. $k(\pmb\alpha_{1:\delta},\pmb\alpha'_{1:\delta})=\tilde k(\Vert\pmb\alpha_{1:\delta}-\pmb\alpha'_{1:\delta}\Vert_{\R^\delta})$, the correlation between designs is $k(\pmb\alpha_{1:\delta},\pmb\alpha'_{1:\delta})=\tilde k(\Vert\mathbf V_{1:\delta}^\top(\phi(\x)-\phi(\x'))\Vert_{\R^\delta})=\tilde k(r)$ with $r^2=(\phi(\x)-\phi(\x'))^\top\mathbf M(\phi(\x)-\phi(\x'))$ where $\mathbf M=\mathbf V_{1:\delta}\mathbf V_{1:\delta}^\top$ is a $D\times D$ matrix with low rank ($\delta$). Hence, this model implements a Gaussian Process in the $\Phi$ space with an integrated linear dimensionality reduction step \cite{GPML}.
Note that the kernel is non-stationary in the original $X$ space.
	
The approaches \cite{constantine2014active,bouhlel2016improving,tripathy2016gaussian} mainly differ from that proposed in Equation (\ref{eq:Yalpha}) in the construction of the reduced basis: in Equation (\ref{eq:Yalpha}), dimension reduction is carried out without the need to call the expensive $f(\x)$ (or its gradient): the directions of largest variation of an easy to compute mapping $\phi(\cdot)$ are used instead. 
This also prevents from a spurious or incomplete projection when $n$ is smaller than $D$ and avoids recomputing the basis at each iteration.
	
This is nonetheless a limitation since the $Y(\pmb\alpha_{1:\delta})$ approach relies only on considerations about the shape geometry. 
The output $y$ is not taken into account for the dimension reduction even though some $\v^j$, $j\in\{1,\dotsc,\delta\}$ may influence $y$ or not. 
Two shapes which differ in the $\alpha_j$ components with $j\le\delta$ may behave similarly in terms of output $y$, so that further dimension reduction is possible. Vice versa, eigencomponents that have a small geometrical effect and were neglected may be reintroduced because they matter for $y$.

\begin{figure}[h!]
	\centering
	\includegraphics[width=0.6\textwidth]{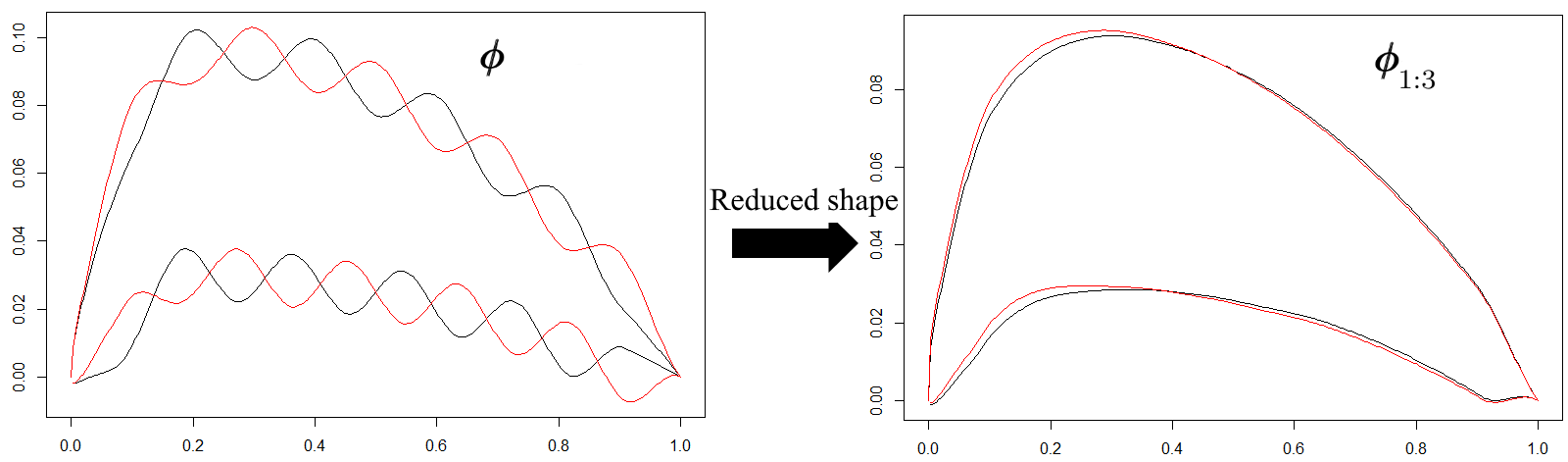}
	\caption{Example of two different shapes (black and red) whose reconstruction in the space of the three first eigenshapes is very similar.}
	\label{fig:naca_proches}
\end{figure}

As an illustration consider the red and black shapes of Figure \ref{fig:naca_proches}. Both are associated to parameters $\x$ and $\x'$ and their discretizations $\phi(\x)$ and $\phi(\x')$ are quite different. Depending on the objective function, $f(\x)$ and $f(\x')$ might differ widely. However, when considering the $\overline{\pmb\phi}+\sum_{j=1}^{\delta}\alpha_j\v^j$ reconstruction with $\delta=3$, they look very similar because $\pmb\alpha_{1:3}\approx\pmb\alpha'_{1:3}$. 
Even though $\mathcal V_{1:3}:=\{\v^1,\v^2,\v^3\}$ is a tempting basis because it explains 98.5\% of the discretizations variance, it is not a good choice if $f(\x)$ and $f(\x')$ are different: because of continuity assumptions a surrogate model would typically suffer from inputs $\pmb\alpha\approx\pmb\alpha'$ with $y\ne y'$.

For this reason, instead of building the surrogate in the space spanned by the most relevant shape modes, we would prefer to build it in the $\mathcal V_a\subset\mathcal V$ basis of the most output-influencing eigenshapes $\pmb\alpha^a$. Additionally, since the remaining ``inactive'' components $\pmb\alpha^{\overline{a}}$ refine the shape and might explain small fluctuations of $y$, instead of omitting them (which is equivalent to stating $\pmb\alpha^{\overline{a}}=\mathbf 0$), we would like to keep them in the surrogate model while prioritizing $\pmb\alpha^a$: a GP $Y^a(\mathbf W_a\phi(\x))+Y^{\overline{a}}(\mathbf W_a\phi(\x))$ is detailed in Sec. \ref{sec:additive_gp}.

\subsection{Supervised dimension reduction}
\label{sec:supervised_dimension_reduction}
\subsubsection{Selection of active eigenshapes}
\label{sec:eigenshape_selection}
To select the eigencomponents that impact $y$ the most, the penalized log-likelihood \cite{yi2011penalized} of a regular, anisotropic GP in the high dimensional space of $\pmb\alpha$'s is considered,
\begin{equation}
\max_\vartheta pl_\lambda(\pmb\alpha^{(1:n)},{\mathbf y_{1:n}};\vartheta) \quad \text{ where } \quad
pl_\lambda(\pmb\alpha^{(1:n)},{\mathbf y_{1:n}};\vartheta):=l(\pmb\alpha^{(1:n)},{\mathbf y}_{1:n};\vartheta)-\lambda\Vert\pmb\theta^{-1}\Vert_1
\label{eq:pmle}
\end{equation}
The $\vartheta$ are the GP's hyper-parameters made of the length-scales $\theta_j$, a constant mean term $\beta$, and the variance of the GP $\sigma^2$. 
$\pmb\alpha^{(1:n)}$ are the eigencomponents of the evaluated designs $\x^{(1)},\dotsc,\x^{(n)}$, and $\mathbf y_{1:n}$ the associated outputs, $\mathbf y_{1:n}=(y_1,\dotsc,y_n)^\top=(f(\x^{(1)}),\dotsc,f(\x^{(n)}))^\top$. 
The mean and the variance terms can be solved for analytically by setting the derivative of the penalized log-likelihood (\ref{eq:pmle}) equal to 0 which yields
\begin{equation}
\widehat\beta:=\frac{\mathbf1^\top\mathbf R^{-1}_{\pmb\theta}\mathbf y_{1:n}}{\mathbf1^\top \mathbf R^{-1}_{\pmb\theta}\mathbf 1}
\quad \text{ and } \quad
\widehat{\sigma}^2:=\frac1n(\mathbf y_{1:n}-\mathbf1\widehat\beta)^\top\mathbf R^{-1}_{\pmb\theta}(\mathbf y_{1:n}-\mathbf1\widehat\beta)
\label{eq:beta_sig}
\end{equation}
where $\mathbf K_\vartheta$ is the covariance matrix with entries ${K_\vartheta}_{ij}=\widehat{\sigma^2}k_{\pmb\theta}(\x^{(i)},\x^{(j)})$, with determinant $\vert\mathbf K_\vartheta\vert$ and $\mathbf R_{\pmb\theta}$ is the correlation matrix, $R_{ij}=k_{\pmb\theta}(\x^{(i)},\x^{(j)})$.
The (concentrated) penalized log-likelihood of this GP is
\begin{equation}
pl_\lambda(\pmb\alpha^{(1:n)},{\mathbf y}_{1:n};\vartheta)=-\frac n2\log(2\pi)-\frac12\log(\vert\mathbf K_\vartheta\vert)-\frac12({\mathbf y}_{1:n}-\mathbf 1\widehat{\beta})^\top\mathbf K_\vartheta^{-1}({\mathbf y}_{1:n}-\mathbf 1\widehat{\beta})-\lambda\Vert\pmb\theta^{-1}\Vert_1
\label{eq:concpenalLL}
\end{equation}

The penalization is applied to $\pmb\theta^{-1}:=(1/\theta_1,\dotsc,1/\theta_D)^\top$, the vector containing the inverse length-scales of the GP.
It is indeed clear \cite{salem2018sequential} that if $\theta_j\rightarrow+\infty$, the direction $\v^j$ has no influence on $y$ as all the points are perfectly correlated together, making the GP flat in this dimension. 
The $L^1$ penalty term applied to the $\theta_j$'s performs variable selection: this Lasso-like procedure promotes zeros in the vector of inverse length-scales, hence sets many $\theta_j$'s to $+\infty$. 
Few directions with small $\theta_j$ are selected and make the active dimensions, $\pmb\alpha^a$ (step \circled{3} in Figure \ref{fig:summary}).
Even if the maximization of $pl_\lambda$ is carried out in a $D$-dimensional space, the problem is tractable since the gradients of $pl_\lambda$ are analytically known \cite{roustant2012dicekriging}, and because the $L^1$ penalty convexifies the problem. 
We solve it using standard gradient-based techniques such as BFGS \cite{liu1989limited} with multistart. 

Numerical experiments not reported here for reasons of brevity have shown that most local optima to this problem solely differ in $\theta_j$'s that are already too large to be relevant and consistently yield the same set of active variables $\pmb\alpha^a$.
Notice that in \cite{yi2011penalized}, a similar approach is undertaken but the penalization was applied on the reciprocal variables $\mathbf w=(w_1,\dotsc,w_D)^\top$ with $w_j=1/\theta_j$. In our work, the inverse length-scales are penalized, the gradient of the penalty is proportional to $1/\theta_j^2$. This might help the optimizer since directions with $\theta_j$'s that are not large yet are given more emphasis. In comparison, the $\mathbf w$ penalty function's gradient is isotropic. 
Since we can restrict the number of variables to $d'\ll D$ with no loss of information (cf. discussion at the end of Section \ref{sec:experiments_reduction}), the dimension of Problem (\ref{eq:pmle}) is substantially reduced which leads to a more efficient resolution. 
Because the $\alpha_j$'s have zero mean and variance $\lambda_j$, they have magnitudes that decrease with $j$. 
When $m<n$, $1/\theta_n$ is typically larger than $1/\theta_m$, meaning that the optimizer is better rewarded by diminishing $1/\theta_n$ than $1/\theta_m$. 
Starting from reasonable $\theta_j$ values\footnote{Typically of the order of range$(\alpha_j)$.} the first $\theta_j$'s are therefore less likely to be increased in comparison with the last ones, i.e. they are less likely to be found inactive. 
This can be seen as a bias which can be removed by scaling all $\alpha_j$'s to the same interval. 
However, we do not normalize the $\pmb\alpha$ variables for two reasons. 
First, since the $\alpha_j$'s correspond to reconstruction coefficients associated to normalized eigenshapes ($\Vert\v^j\Vert_{\R^D}=1$), they share the same physical dimension and can be interpreted in the same manner. 
Second, this bias is equivalent to assuming that the most significant shape variations are responsible for the largest output variations, which is a reasonable prior.
In experiments that are not reported here for the sake of brevity, we have noticed that a BFGS algorithm optimizing Problem (\ref{eq:pmle}) got trapped by weak local optima more frequently when the $\alpha_j$'s were normalized.
	
\begin{definition}[Selection of active dimensions]
Let a GP be indexed by $\alpha_1,\dotsc,\alpha_D \in [\pmb\alpha^{\min},\pmb\alpha^{\max}] \subset \mathbb R^D$ and $\{\pmb\alpha^{(1:n)},\mathbf y_{(1:n)})\}$ be the data to model. The length-scales $\pmb\theta$ of the GP are set by maximizing the $L^1$ penalized concentrated log-likelihood of Equation (\ref{eq:concpenalLL}).
A dimension $j$ is declared \emph{active} if
\begin{equation*}
\frac{\theta_j}{\text{range}(\alpha_j)} \le 10 \times \underset{i=1,\dotsc,D}{\min}~\frac{\theta_i}{\text{range}(\alpha_i)}\text{.}
\end{equation*}
The $\delta$ such active dimensions are denoted $\pmb\alpha^a=(\alpha_{a_1},\dotsc,\alpha_{a_\delta})\in\R^\delta$.
\label{def:selection}
\end{definition}
Since the $\alpha_j$'s have different (decreasing) ranges, the length-scales have to be normalized by the range of $\pmb\alpha_j^{(1:n)}$ to be meaningful during this $\theta_j$ comparison.
Our implementation extends the likelihood maximization of the \texttt{kergp} package \cite{deville2015package} to include the penalization term. 
After a dimensional analysis of $pl_\lambda$, we have chosen to take $\lambda=\frac nD$ 
to balance both terms. Other techniques such as cross-validation or the use of different $\lambda$'s for obtaining a pre-defined number of active components can also be considered.

\begin{figure}[h!]
		\centering
		\includegraphics[width=\textwidth]{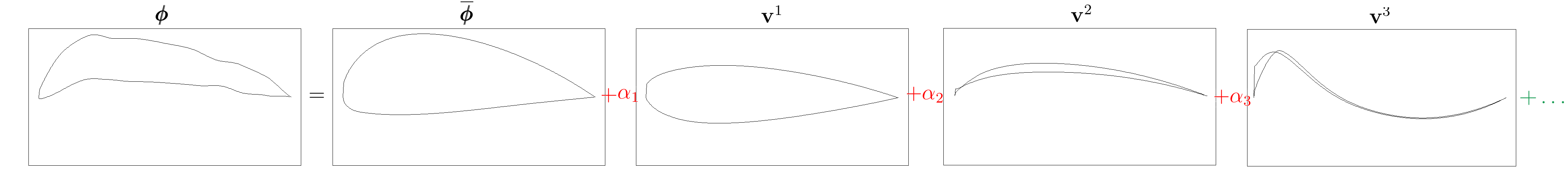}\\[0.5cm]
		\includegraphics[width=\textwidth]{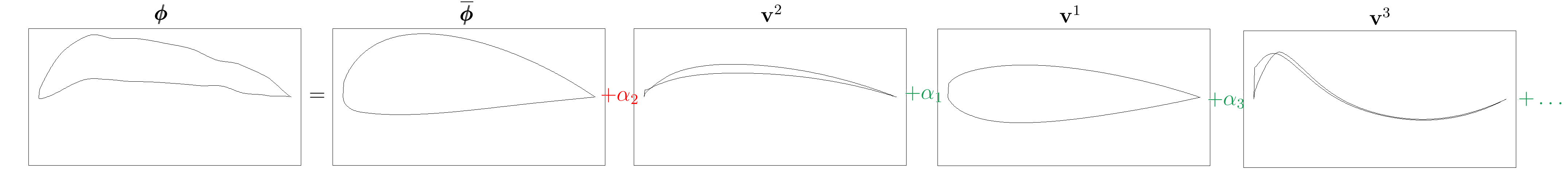}
		\caption{Variable selection on the NACA 22 benchmark by penalized maximum likelihood. 
For the drag (top), the three first eigenshapes that act on the shape, hence on its drag, are selected (red coefficients). For the lift, only the second eigencomponent ($\v^2$) is selected (bottom). Indeed $\v^2$ modifies the camber of the airfoil, hence it plays a major role on the lift. The other eigenbasis vectors (green coefficients) are estimated to be less influential on $y$.}
		\label{fig:naca_selection}
	\end{figure}

On the NACA 22 benchmark with few observations of $f(\cdot)$ ($n=15$ here), Figure \ref{fig:naca_selection} 
gives the only few active components that are selected by the penalized maximum likelihood procedure. 
The three first principal axes, $\v^1$, $\v^2$ and $\v^3$ are retained when considering the drag (top). Indeed, these are the eigenshapes that globally impact the shape the most and change its drag. 
When the output $y$ is the lift (bottom), only the second principal axis is selected. This eigenshape modifies the camber of the shape, which is known to highly impact the lift. The other eigenvectors are detected to be less critical for $y$'s variations.
When $n$ grows, more eigenshapes get selected because they also slightly impact the output. For instance when $n=50$, some eigenshapes that contain bumps (the 4th, the 5th, the 8th, etc.) are selected for modeling the lift. 
They also contribute to changing the camber of the airfoil, hence its lift.
	
\subsubsection{Additive GP between active and inactive eigenshapes}
\label{sec:additive_gp}
\begin{figure}[h!]
	\centering
	\includegraphics[width=0.3\textwidth]{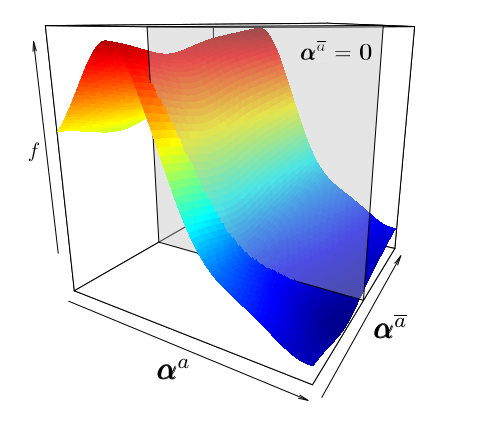}
	\caption{Example of a function that primarily varies along the $\pmb\alpha^a$ direction, and secondarily along $\pmb\alpha^{\overline{a}}$. If $\pmb\alpha^{\overline{a}}$ is omitted, one implicitly considers the restriction of $f(\cdot)$ to the gray plane where $\pmb\alpha^{\overline{a}}=\mathbf0$.}
	\label{fig:illustration_gp_additif}
\end{figure}

Completely omitting the non-active dimensions, $\pmb\alpha^{\overline a}\in\R^{D-\delta}$, and building the surrogate model $Y(\cdot)$ in the sole $\pmb\alpha^a$ space may amount to erasing some geometric patterns of the shapes which contribute to small variations of $y$. 
For this reason, an additive GP \cite{durrande2012additive,duvenaud2011additive} with zonal anisotropy \cite{allard2016anisotropy} between the active and inactive eigenshapes is considered (step \circled{4} in Figure \ref{fig:summary}):
\begin{equation}Y(\pmb\alpha)=\beta+Y^a(\pmb\alpha^a)+Y^{\overline a}(\pmb\alpha^{\overline a})\text{.}\label{eq:modele_additif}\end{equation}
$Y^a(\pmb\alpha^a)$ is the anisotropic main-effect GP which works in the reduced space of active variables. It requires the estimation of $\delta+1$ hyper-parameters (the length-scales $\theta_j$ and a GP variance $\sigma^2_a$) and aims at capturing most of $y$'s variation, related to $\pmb\alpha^a$'s effect.
$Y^{\overline a}(\pmb\alpha^{\overline a})$ is a GP over the large space of inactive components. 
It is a GP which just takes residual effects into account. 
To keep $Y^{\overline a}(\pmb\alpha^{\overline a})$ tractable, it is considered isotropic, i.e., it only has 2 hyper-parameters, a unique length-scale $\theta_{\overline a}$ and a variance $\sigma^2_{\overline a}$. 
In the end, even though $Y(\pmb\alpha)$ operates with $\pmb\alpha$'s $\in\R^D$ and there are fewer observations than dimensions\footnote{Even if pruning the $\alpha_j$ components for $j>d'$ (see comments at the end of Section \ref{sec:experiments_reduction}), $n<d'$ may hold.}, $n\ll D$, it remains tractable since only a
total of $\delta+3\ll n$ hyperparameters have to be learned, which guarantees the identifiability, i.e. the unicity of the hyperparameters solution even when the number of observations is small.
Although the $\alpha_j$'s have different ranges, they are homogeneous in that they all multiply normalized eigenshapes.
Thus, the distances inside the shape manifold, $\mathcal A$, should be relevant and an isotropic model is a possible assumption, which again, tends to emphasize eigenshapes that appear the most within the designs.
This additive model can be interpreted as a GP in the $\pmb\alpha^a$ space, with an inhomogeneous noise fitted by the $Y^{\overline{a}}(\cdot)$ GP \cite{durrande2011etude}. 
It aims at modeling a function that varies primarily along the active dimensions, and fluctuates only marginally along the inactive ones, as illustrated in Figure \ref{fig:illustration_gp_additif}.

Denoting $k_a$ and $k_{\overline{a}}$ the kernels of the GPs, the hyper-parameters $\vartheta_a=(\theta_{a_1},\dotsc,\theta_{a_\delta},\sigma^2_a)$ and $\vartheta_{\overline{a}}=(\theta_{\overline a},\sigma^2_{\overline a})$ are estimated by maximizing the log-likelihood of (\ref{eq:modele_additif}) given the observed data ${\mathbf y}_{1:n}$,
\[l_Y(\pmb\alpha^{(1:n)},{\mathbf y_{1:n}};\vartheta_a,\vartheta_{\overline{a}})=-\frac n2\log2\pi-\frac12\log(\vert\mathbf K\vert)-\frac12(\mathbf y_{1:n}-\mathbf1\widehat{\beta})^\top\mathbf K^{-1}(\mathbf y_{1:n}-\mathbf1\widehat{\beta})\text{,}\]
using the \texttt{kergp} package \cite{deville2015package}. $\mathbf K=\mathbf K_a+\mathbf K_{\overline{a}}$, with ${K_a}_{ij}=\sigma^2_{a}k_a({\pmb\alpha^a}^{(i)},{\pmb\alpha^a}^{(j)})$, and ${K_{\overline{a}}}_{ij}=\sigma^2_{\overline{a}}k_{\overline{a}}({\pmb\alpha^{\overline{a}}}^{(i)},{\pmb\alpha^{\overline{a}}}^{(j)})$, and $\widehat{\beta}$ is 
given by Equation (\ref{eq:beta_sig}).
The correlation between $\pmb\alpha$ and $\pmb\alpha'$ being $k(\pmb\alpha,{\pmb\alpha}')=\sigma^2_ak_a({\pmb\alpha}^a,{{\pmb\alpha}^a}')+\sigma^2_{\overline{a}}k_{\overline{a}}({\pmb\alpha}^{\overline{a}},{\pmb\alpha^{\overline{a}}}')$, the kriging predictor and variance of this additive GP are \cite{GPML}

\begin{align}
\begin{split}
m(\pmb\alpha)=\mathbf1_n\widehat{\beta}+k(\pmb\alpha,\pmb\alpha^{(1:n)})^\top \mathbf K^{-1}({\mathbf y}_{1:n}-\mathbf1_n\widehat{\beta})\\
s^2(\pmb\alpha)=\sigma^2_a+\sigma^2_{\overline{a}}-k(\pmb\alpha,\pmb\alpha^{(1:n)})^\top\mathbf K^{-1}k(\pmb\alpha,\pmb\alpha^{(1:n)})
\label{eq:formules_krigeage_additif}
\end{split}
\end{align}

\subsection{Experiments: Metamodeling in the eigenshape basis}
\label{sec:metamodeling_shape_eigenbasis}
We now study the performance of the variable selection and of the additive GP described in the previous section. 
The different versions of GPs that are compared are the following:
\begin{itemize}
\item \texttt{GP($X$)} is the GP in the original space of parameters $X$; 
\item \texttt{GP($\pmb\alpha_{\text{\_\_}}$)} indicates the GP is built in the space of $\text{\_\_}$ (to be specified) principal components;
\item \texttt{GP($\pmb\alpha^a$)} means the GP works with the active $\pmb\alpha$'s only;
\item \texttt{AddGP($\pmb\alpha^a+\pmb\alpha^{\overline{a}}$)} refers to the additive GP (Section \ref{sec:additive_gp}).
\end{itemize}

We equip the example designs \ref{ex:cercle_surparametre}, \ref{ex:coeur}, \ref{ex:catenoide} and \ref{ex:naca22} (Section \ref{sec:experiments_reduction}) with objective functions $f(\x)$ that are to be modeled by the fitted GPs. 
For each function, the predictive capability of different models is compared on a distinct test set
using the R2 coefficient of determination.
Later, in Section \ref{sec:experiments_optimization}, the objective functions will be optimized.

\begin{itemize}
	\item Example \ref{ex:cercle_surparametre}: $f_2(\x)=r-\pi r^2-\Vert(x,y)^\top-(3,2)^\top\Vert_2$, where $x$, $y$ and $r$ correspond to the position of the center and the radius of the over-parameterized circle (and accessible through $\x$), respectively.
	\item Example \ref{ex:coeur}: $f_4(\x)=\Vert\Omega_{\mathbf t}-\Omega_{\tilde{\mathbf x}}\Vert_2^2$ where $\tilde{\mathbf x}:=\x-(x_1+2.5,x_2+2.5,0,\dotsc,0)^\top$ corresponds to the centered design, and $\Omega_{\mathbf t}$, $\Omega_{\tilde{\mathbf x}}$ are the nodal coordinates of the shapes, see Figure \ref{fig:rectangle}. The goal is to retrieve a target shape $\mathbf t=(t_1,\dotsc,t_{40})^\top$ whose lower left point (A) is set at $t_1=t_2=2.5$ with the flexible rectangle defined by $\x$. The A point of any shape $\x$ is first moved towards (2.5, 2.5) too, and $f_4$ measures the discrepancy. Here, the target $\mathbf t$ is the rectangular heart shown in Figure \ref{fig:heart_design}.

	\begin{figure}[h!]
		\centering
		\includegraphics[width=0.4\textwidth]{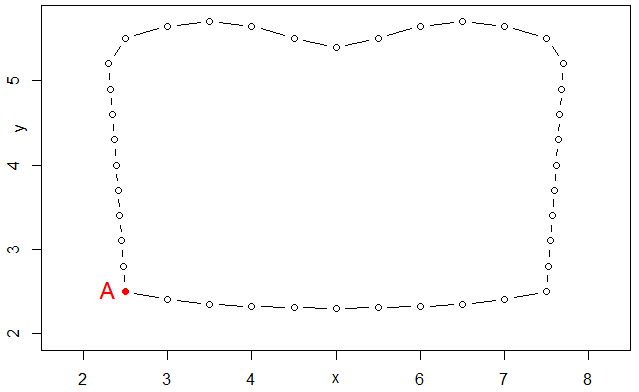}
		\caption{Rectangular heart target shape of Example \ref{ex:coeur}.}
		\label{fig:heart_design}
	\end{figure}
	
	\item Example \ref{ex:catenoide}: $f_5(r)=2\pi\int_{y_A}^{y_B}r(y)\sqrt{1+r'(y)^2}dy$: inspired by the catenoid problem \cite{Colding11106}, we aim at finding a regular curve joining two points $\text{A}=(0,y_A)$ and $\text{B}=(1,y_B)$, with the smallest axisymmetric surface. The curve $r(y)$ is the straight line between A and B, modified by $\mathbf r=(r_1,\dotsc,r_{29})^\top$, see Figure \ref{fig:catenoid_problem_description}.
	
	\item Example \ref{ex:naca22}: the objective functions are the lift coefficient and the drag coefficient of the airfoil, $f_{7L}$, $f_{7D}$. The latter are computed using a commercial Computational Fluid Dynamics (CFD) computer code.
\end{itemize}

\subsubsection*{Over-parameterized circle (Example \ref{ex:cercle_surparametre})}
For the over-parameterized circle, the objective function is
$f_2(\x)=r-\pi r^2-\Vert(x,y)^\top-(3,2)^\top\Vert_2$, where $x$, $y$ and $r$ correspond to the position of the center and the radius of the circle (accessible through $\x$), respectively.
$f_2$ explicitly depends on the parameters that truly define the circle. Three models are compared
\begin{itemize}
	\item A model using the CAD parameters $\x\in\R^{39}$;
	\item A model using the 3 first eigencomponents, $(\alpha_1,\alpha_2,\alpha_3)$;
	\item A model built over the \emph{true} circle parameters $(x,y,r)$.
\end{itemize}

Table \ref{tab:R2_cercle} gives the average R2 over 10 runs with different space-filling DoEs of size $n=20,50,100,200$. Since $d=39>20$, no GP was fitted in the CAD parameter space when $n=20$.

\begin{table}[!ht]
	\centering
	\makebox[\textwidth][c]{
		\begin{tabular}{|c|c|c|c|}
			\hline
			$n$ & \texttt{GP($X$)} & \texttt{GP($\pmb\alpha_{1:3}$)} & \texttt{GP(True)}\\\hline
			20 & - & 0.99741 & 0.99701\\
			50 & 0.78193 & 0.99954 & 0.99951\\
			100 & 0.86254 & 0.99984 & 0.99985\\
			200 & 0.93383 & 0.99992 & 0.99997\\\hline
		\end{tabular}
	}
	\caption{Average R2 over 10 runs for the prediction of $f_2$. 
\texttt{GP($X$)} is the GP in the 39-dimensional CAD parameter space, \texttt{GP($\pmb\alpha_{1:3}$)} corresponds to a GP fitted to the 3 first principal components $\alpha_1,\alpha_2,\alpha_3$, and \texttt{GP(True)} to the GP with the space of minimal circle coordinates.}
	\label{tab:R2_cercle}
\end{table}

$f_2$ is easily learned by the surrogate model as shown by large R2 values. Obviously, the quality of prediction increases with $n$ and the eigenshape GP (\texttt{GP($\pmb\alpha_{1:3}$)}) built in a 3-dimensional space outperforms the GP in the CAD parameters space (\texttt{GP($X$)}, $d=39$). 
Yet, the \texttt{GP($\pmb\alpha_{1:3}$)} performs as well (and even better for small $n$'s) as \texttt{GP(True)}. 

\subsubsection*{Heart target (Example \ref{ex:coeur})}
We turn to the metamodeling of $f_4$. 
It is a 40-dimensional function, $f_4(\x)=\Vert\Omega_{\mathbf t}-\Omega_{\tilde{\mathbf x}}\Vert_2^2$ that explicitly depends on the CAD parameters. Unlike the previous test problem, the shapes do not have superfluous parameters since all $x_j$'s are necessary to retrieve $\mathbf t$.

7 different models detailed through Sections 
\ref{sec:unsupervised_dimension_reduction} and \ref{sec:supervised_dimension_reduction} are investigated. 
\texttt{GP($X$)}, the standard GP carried out in the space of CAD parameters. \texttt{GP($\pmb\alpha_{1:40}$)}, the metamodel built in the space of 40 first principal components. Indeed, Table \ref{tab:eigenvalues_rectangle} informed us that any shape is retrieved via its 40 first eigenshape coefficients. To build surrogates in reduced dimension, considering the cumulative eigenvalue sum in Table \ref{tab:eigenvalues_rectangle}, \texttt{GP($\pmb\alpha_{1:2}$)}, \texttt{GP($\pmb\alpha_{1:4}$)} and \texttt{GP($\pmb\alpha_{1:16}$)} are models that consider the 2, 4 and 16 first principal components only. 
Finally, \texttt{GP($\pmb\alpha^{a}$)} and \texttt{AddGP($\pmb\alpha^a+\pmb\alpha^{\overline{a}}$)} are also compared.

\begin{figure}[h!]
	\centering
	\includegraphics[width=0.6\textwidth]{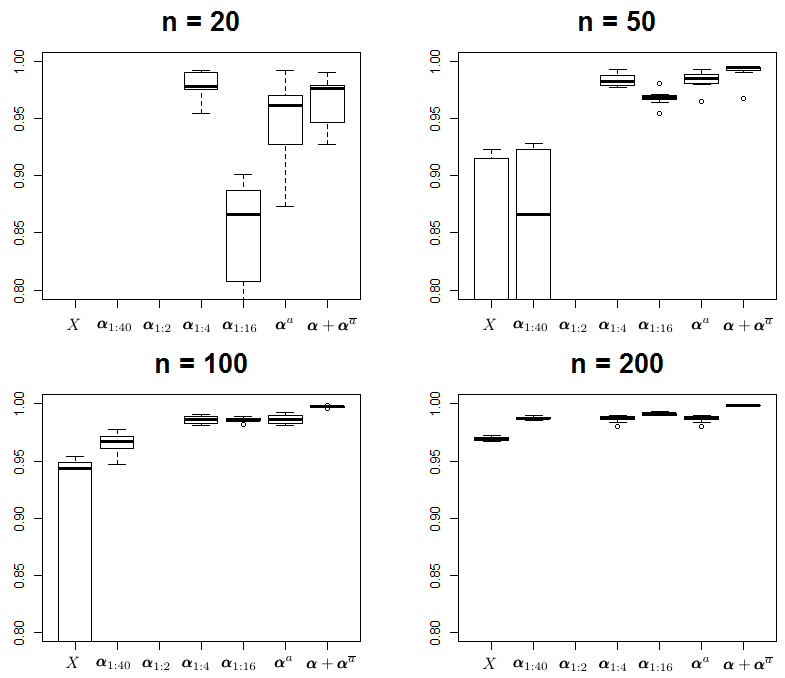}
	\caption{Boxplots of R2 coefficient for the different models, rectangle test case (Example \ref{ex:coeur}).
}
	\label{fig:boxplot_coeur}
\end{figure}

Table \ref{tab:R2_coeur} reports the average R2 indicator over 10 runs starting with space-filling DoEs of size $n=20,50,100,200$. Figure \ref{fig:boxplot_coeur} shows a boxplot of the results (for the sake of clarity, only runs with R2 $\ge$ 0.8 are shown). The input dimension for \texttt{GP($X$)} and for \texttt{GP($\pmb\alpha_{1:40}$)} is too large for coping with $n=20$ observations. \texttt{GP($\pmb\alpha_{1:40}$)} is given beside \texttt{GP($X$)} because both GPs have the same input space dimension.

\begin{table}[!ht]
	\centering
	\makebox[\textwidth][c]{
		\begin{tabular}{|c|c|c|c|c|c|c|c|}
			\hline
			$n$ & \texttt{GP($X$)} & \texttt{GP($\pmb\alpha_{1:40}$)} & \texttt{GP($\pmb\alpha_{1:2}$)} & \texttt{GP($\pmb\alpha_{1:4}$)} & \texttt{GP($\pmb\alpha_{1:16}$)} & \texttt{GP($\pmb\alpha^a$)} & \texttt{AddGP($\pmb\alpha^a+\pmb\alpha^{\overline{a}}$)}\\\hline
			20 & - & - & -0.063 & 0.979 & 0.844 & 0.935 & 0.967\\
			50 & 0.455 & 0.542 & -0.009 & 0.984 & 0.968 & 0.983 & 0.991\\
			100 & 0.662 & 0.868 & 0 & 0.986 & 0.986 & 0.986 & 0.997\\
			200 & 0.873 & 0.988 & 0 & 0.987 & 0.991 & 0.987 & 0.999\\\hline
		\end{tabular}
	}
	\caption{
		Average R2 over 10 runs when metamodeling $f_4$.}
	\label{tab:R2_coeur}
\end{table}

The benefits of the additive GP appear to be threefold. 
First, it ensures sparsity by selecting a small number of eigenshapes for the anisotropic part of the kernel. 
A high-dimensional input space hinders the predictive capabilities when $n$ is small, as confirmed by the weak performance of \texttt{GP($X$)}, \texttt{GP($\pmb\alpha_{1:40}$)} and even \texttt{GP($\pmb\alpha_{1:16}$)} for $n=20$. 
When $n$ increases, higher-dimensional models become more accurate. For $n=100$ and $n=200$, the model with 16 principal components outperforms the one with 4 principal components, even though the latter was more precise with $n=20$ or $n=50$ observations. In the case $n=200$, even \texttt{GP($\pmb\alpha_{1:40}$)} outperforms the 4 dimensional one (\texttt{GP($\pmb\alpha_{1:4}$)}). This is due to the fact that more principal components mean a more realistic shape, hence less ``input space errors''. When few observations are available, these models suffer from the curse of dimensionality, but become accurate as soon as their design space gets infilled enough. With more observations, \texttt{GP($\pmb\alpha_{1:40}$)} may become the best model.

Besides the dimension reduction,
the selection of eigenshapes that truly influence the output is also critical. 
According to Table \ref{tab:eigenvalues_rectangle}, a tempting decision to reduce the dimension would be to retain the two first principal components, i.e. \texttt{GP($\pmb\alpha_{1:2}$)}.
But since the 2 first eigenshapes act on the shape's position (see Figure \ref{fig:coeur_eigenshapes}) to which $f_4$ is insensitive, this is a weak option, as pointed out by the R2 scores which are close to 0 for this model. 
Here, the selected variables are usually the 3rd and the 4th eigenshape which act on the size of the rectangle, hence are of first order importance for $f_4$. 
In about 30\% of the runs, they are accompanied by the first and the second one, and more rarely by other eigenshapes. 

Third, the \texttt{AddGP($\pmb\alpha^a+\pmb\alpha^{\overline{a}}$)} outperforms \texttt{GP($\pmb\alpha^a$)}. 
Indeed, the less important eigenshapes (from a geometric point of view) $\v^5,\dotsc,\v^{40}$ locally modify the rectangle, and allow the final small improvements in $f_4$. This highlights the benefits of taking the remaining eigenshapes which act as local shape refinements into account.

Last, even though their input spaces have the same dimension, \texttt{GP($\pmb\alpha_{1:40}$)} consistently outperforms \texttt{GP($X$)}. 
This confirms our comments about the NACA manifold of Figure \ref{fig:naca3_eigenshapes}: 
the eigenshapes are a better representation than the CAD parameters for statistical prediction.

\subsubsection*{Catenoid shape (Example \ref{ex:catenoide})}
In relation with the catenoid, we introduce the objective function
$f_5(r)=2\pi\int_{y_A}^{y_B}r(y)\sqrt{1+r'(y)^2}dy$.
$f_5$ is an integral related to the surface of the axisymmetric surface given by the rotation of a curve $r(y)$. In our example, $r(y)$ is the line between two points A and B modified by regularly spaced deviations $\mathbf r=(r_1,\dotsc,r_{29})^\top$. 
Only $\mathbf r$'s generated by a GP that lead to a curve inside a prescribed envelope (see Figure \ref{fig:catenoid_problem_description}) are kept in the same spirit as \cite{li2019surrogate} where a smoothing operator is applied to consider realistic airfoils. 
With this, it is expected that less than 29 dimensions suffice to accurately describe all designs. This is confirmed by the eigenvalues in Table \ref{tab:eigenvalues_catenoide} and the true dimensionality detected to be 7.

In this experiment, we compare the predictive capabilities of six models. The first one is the classical \texttt{GP($X$)}. 
The objective function explicitly depends on $\mathbf r$ but its high-dimensionality may be a drawback for metamodeling. 
Even though less dimensions are necessary and many eigenshapes correspond to noise, a GP fitted to all $d=29$ eigenshapes, \texttt{GP($\pmb\alpha_{1:29}$)}, is considered. Along with it, \texttt{GP($\pmb\alpha_{1:4}$)} and \texttt{GP($\pmb\alpha_{1:7}$)} are considered. 
The former is an unsupervised dimension reduction, considering the $\lambda_j$'s, while the latter is the full dimensional eigenshape GP, since the eigenshapes 8 to 29 are non-informative.
Finally, the GPs with variable selection \texttt{GP($\pmb\alpha^a$)} and \texttt{AddGP($\pmb\alpha^a+\pmb\alpha^{\overline{a}}$)}, are also compared.

\begin{figure}[h!]
\centering
\includegraphics[width=0.6\textwidth]{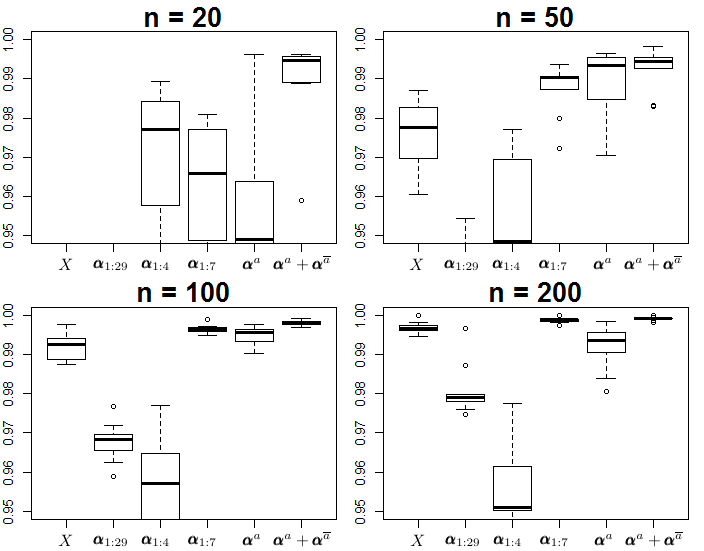}
\caption{Boxplots of R2 coefficient for the different models, catenoid test case (Example \ref{ex:catenoide}).}
\label{fig:boxplot_catenoide}
\end{figure}

Table \ref{tab:R2_catenoide} reports the average R2 indicator over 10 runs starting with space-filling DoEs of size $n=20,50,100,200$. Figure \ref{fig:boxplot_catenoide} shows a boxplot of the results (for the sake of clarity, only runs with R2 $\ge$ 0.95 are shown). The input dimension for \texttt{GP($X$)} and for \texttt{GP($\pmb\alpha_{1:29}$)} is too large for coping with $n=20$ observations. \texttt{GP($\pmb\alpha_{1:29}$)} is given beside \texttt{GP($X$)} because these GPs have the same input space dimension.

\begin{table}[!ht]
	\centering
	\makebox[\textwidth][c]{
		\begin{tabular}{|c|c|c|c|c|c|c|}
			\hline
			$n$ & \texttt{GP($X$)} & \texttt{GP($\pmb\alpha_{1:29}$)} & \texttt{GP($\pmb\alpha_{1:4}$)} & \texttt{GP($\pmb\alpha_{1:7}$)} & \texttt{GP($\pmb\alpha^a$)} & \texttt{AddGP($\pmb\alpha^a+\pmb\alpha^{\overline{a}}$)}\\\hline
			20 & - & - & 0.966 & 0.958 & 0.914 & 0.992\\
			50 & 0.976 & 0.925 & 0.954 & 0.987 & 0.938 & 0.997\\
			100 & 0.992 & 0.968 & 0.958 & 0.997 & 0.957 & 0.999\\
			200 & 0.997 & 0.981 & 0.952 & 0.998 & 0.951 & 0.999\\\hline
		\end{tabular}
	}
	\caption{Average R2 over 10 runs for the metamodeling of $f_5$.}
	\label{tab:R2_catenoide}
\end{table}

These results indicate a better performance of \texttt{AddGP($\pmb\alpha^a+\pmb\alpha^{\overline{a}}$)} which benefits from the prioritization of the most influential eigenshapes in the additive model
and, at the same time, accounts for all the 7 eigenshapes. 
Modeling in the space of the full $\pmb\alpha$'s (\texttt{GP($\pmb\alpha_{1:7}$)}) performs fairly well too because the low true dimensionality (7).
Despite its lower dimensionality, \texttt{GP($\pmb\alpha_{1:4}$)} does not work well. 
This is because the refinements induced by $\v^5$, $\v^6$ and $\v^7$ are disregarded while acting on $f_5$. 
This explanation also stands for the moderate performance of \texttt{GP($\pmb\alpha^a$)} in which mainly the 4 first principal components are selected. Including the remaining components in a coarse GP as is done inside \texttt{AddGP($\pmb\alpha^a+\pmb\alpha^{\overline{a}}$)} increases the performance.

Even though there are $d=29$ CAD parameters, \texttt{GP($X$)} exhibits correct performances: since only smooth curves are considered, they are favorable to GP modeling and the curse of dimensionality is damped. In this example, considering all 29 eigenshapes (\texttt{GP($\pmb\alpha_{1:29}$)}), even though it was assumed that solely 7 were necessary, leads to the worst results, since the non-informative eigenshapes augment the dimension without bringing additional information.

\subsubsection*{NACA 22 airfoil (Example \ref{ex:naca22})}
The last example brings us closer to real world engineering problems. 
The objective functions associated to the NACA airfoil with 22 parameters (Example \ref{ex:naca22}), $f_{7L}$ and $f_{7D}$ are the lift and the drag coefficient of this airfoil. 
$f_{7L}$, $f_{7D}$ depend implicitly and nonlinearly on $\x$ through $\Omega_\x$.

Table \ref{tab:eigenvalues_naca22} shows that only the 20 first eigenvectors are informative. 
Seven metamodeling strategies are compared: \texttt{GP($X$)}; \texttt{GP($\pmb\alpha_{1:20}$)}, the surrogate in the space of all 20 meaningful eigenshapes; \texttt{GP($\pmb\alpha_{1:2}$)}, \texttt{GP($\pmb\alpha_{1:3}$)}, \texttt{GP($\pmb\alpha_{1:6}$)} where fewer eigenshapes are considered; \texttt{GP($\pmb\alpha^a$)}; and \texttt{AddGP($\pmb\alpha^a+\pmb\alpha^{\overline{a}}$)}. \texttt{GP($\pmb\alpha_{1:20}$)} is given beside \texttt{GP($X$)} because these GPs have almost the same input space dimension.

\begin{figure}[h!]
	\centering
	\includegraphics[width=\textwidth]{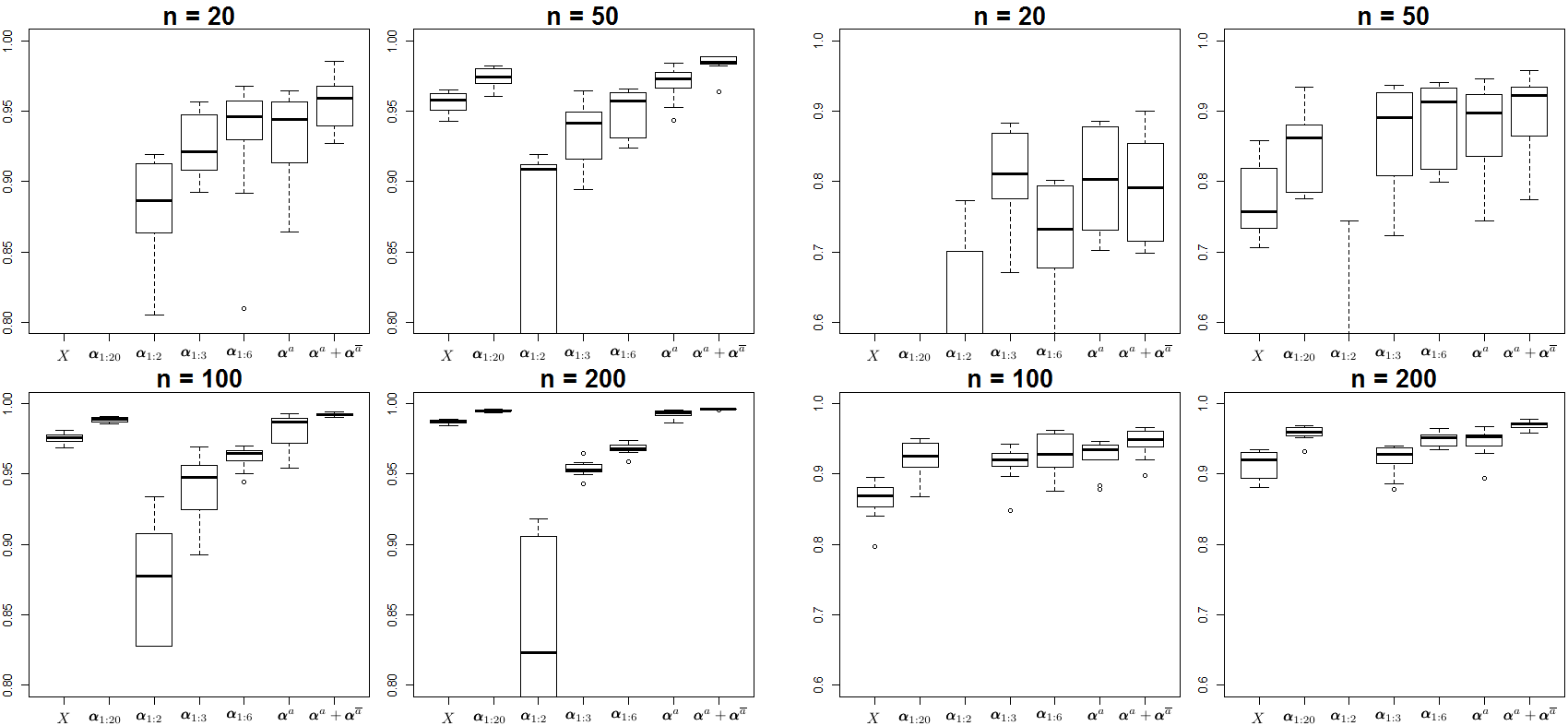}
	\caption{Boxplots of R2 coefficient for the different models, NACA 22 airfoil example. Left: Lift, $f_{7L}$. Right: Drag, $f_{7D}$.
}
	\label{fig:boxplot_naca22}
\end{figure}

Table \ref{tab:R2_naca22} reports the average R2 indicator over 10 runs starting with space-filling DoEs of $n=20,50,100,200$ observations. 
Figure \ref{fig:boxplot_naca22} shows a boxplot of the results (for the sake of clarity, only runs with R2 $\ge$ 0.8 for $f_{7L}$ and $\ge$ 0.6 for $f_{7D}$ are shown). The input dimension for the \texttt{GP($X$)} ($d=22$) and for \texttt{GP($\pmb\alpha_{1:20}$)} is too large for coping with $n=20$ observations.

\begin{table}[!ht]
	\centering
	$f_{7L}$
	\makebox[\textwidth][c]{
		\begin{tabular}{|c|c|c|c|c|c|c|c|}
			\hline
			$n$ & \texttt{GP($X$)} & \texttt{GP($\pmb\alpha_{1:20}$)} & \texttt{GP($\pmb\alpha_{1:2}$)} & \texttt{GP($\pmb\alpha_{1:3}$)} & \texttt{GP($\pmb\alpha_{1:6}$)} & \texttt{GP($\pmb\alpha^a$)} & \texttt{AddGP($\pmb\alpha^a+\pmb\alpha^{\overline{a}}$)}\\\hline
			20 & - & - & 0.857 & 0.907 & 0.930 & 0.935 & 0.957\\
			50 & 0.956 & 0.973 & 0.714 & 0.935 & 0.950 & 0.970 & 0.984\\
			100 & 0.975 & 0.989 & 0.708 & 0.938 & 0.962 & 0.981 & 0.992\\
			200 & 0.987 & 0.995 & 0.515 & 0.954 & 0.968 & 0.993 & 0.996\\\hline
		\end{tabular}
	}
\end{table}
\begin{table}[!ht]
	\centering
	$f_{7D}$
	\makebox[\textwidth][c]{	
	\begin{tabular}{|c|c|c|c|c|c|c|c|}
		\hline
		$n$ & \texttt{GP($X$)} & \texttt{GP($\pmb\alpha_{1:20}$)} & \texttt{GP($\pmb\alpha_{1:2}$)} & \texttt{GP($\pmb\alpha_{1:3}$)} & \texttt{GP($\pmb\alpha_{1:6}$)} & \texttt{GP($\pmb\alpha^a$)} & \texttt{AddGP($\pmb\alpha^a+\pmb\alpha^{\overline{a}}$)}\\\hline
		20 & - & - & 0.443 & 0.806 & 0.720 & 0.800 & 0.796\\
		50 & 0.771 & 0.847 & 0.259 & 0.866 & 0.882 & 0.878 & 0.896\\
		100 & 0.861 & 0.921 & 0.192 & 0.915 & 0.928 & 0.925 & 0.945\\
		200 & 0.915 & 0.958 & -0.008 & 0.920 & 0.950 & 0.946 & 0.969\\\hline
	\end{tabular}
	}
	\caption{Average R2 over 10 runs for the metamodeling of $f_{7L}$ (top) and $f_{7D}$ (bottom).
}
	\label{tab:R2_naca22}
\end{table}

In this example too, \texttt{AddGP($\pmb\alpha^a+\pmb\alpha^{\overline{a}}$)} exhibits the best predictive capabilities. Even though they are coarsely taken into account, the non active eigenshapes which mostly represent bumps, are included in the surrogate model. For the lift, \texttt{GP($\pmb\alpha^a$)} performs quite well too since the $f_{7L}$ relevant dimensions have been selected. 
The variable selection method provides contrasted results between $f_{7L}$ and $f_{7D}$. 
For the lift, the first eigenshape is not always selected. The second and the third one, as well as some higher order eigenshapes get selected, which confirms the effect of the bumps on the lift (see Figures \ref{fig:exemple_naca22} and \ref{fig:naca22_eigenshapes}). For the drag ($f_{7D}$) however, only the 2 or 3 first eigenshapes are usually selected.

We have also noticed that the number of selected components tends to grow with $n$. 
This is a desirable property since with larger samples, an accurate surrogate can be built in a higher dimensional space. 
As already remarked in the previous examples (e.g. Table \ref{tab:R2_coeur}),
it is seen here in Figure \ref{fig:boxplot_naca22} that models with more eigenshapes become more accurate when the number of observations grows.
For $f_{7D}$ (bottom table) for example, when $n$ is small, \texttt{GP($\pmb\alpha_{1:3}$)} is better than \texttt{GP($\pmb\alpha_{1:6}$)} and \texttt{GP($\pmb\alpha_{1:20}$)}, but this changes as $n$ grows, \texttt{GP($\pmb\alpha_{1:6}$)} and \texttt{GP($\pmb\alpha_{1:20}$)} becoming in turn the best eigenshape truncation-based model.
For $f_{7L}$ (top table), in spite of the dimension reduction, very poor results are achieved when retaining only 2 or 3 components, even with small $n$'s. 
When considering only the two first eigenshapes (\texttt{GP($\pmb\alpha_{1:2}$)}), the R2 is weak as the third eigenshape significantly modifies the camber. For this GP, the performance decreases with $n$ because of situations like the one shown in Figure \ref{fig:naca_proches} where shapes that falsely look similar when considering $\pmb\alpha_{1:2}$ only actually differ in lift. Such situations are more likely to occur during the training of the GP as $n$ grows, which degrades performance. 
The example of $f_{7L}$ is informative in the sense that \texttt{GP($\pmb\alpha_{1:20}$)} always outperforms \texttt{GP($\pmb\alpha_{1:6}$)} which outperforms \texttt{GP($\pmb\alpha_{1:3}$)}, for any $n$ (including very little $n$'s), despite the higher dimension. By ignoring second order eigenshapes, \texttt{GP($\pmb\alpha_{1:3}$)} and \texttt{GP($\pmb\alpha_{1:6}$)} provide less reconstruction details. These details are nonetheless important since they change the camber of the airfoil and this is why \texttt{GP($\pmb\alpha_{1:20}$)}, a more precise reconstruction, performs better. Indeed, the remaining $\pmb\alpha$'s mainly reconstruct the bumps of this airfoil as can be seen in Figure \ref{fig:naca22_eigenshapes}, which does influence the lift.

This is also the reason why \texttt{GP($X$)} is better at predicting lift than \texttt{GP($\pmb\alpha_{1:3}$)} and \texttt{GP($\pmb\alpha_{1:6}$)}, which could seem counter-intuitive at first glance since the dimension is reduced. 

Last, let us point out than even though the dimension is almost the same, \texttt{GP($\pmb\alpha_{1:20}$)} consistently outperforms \texttt{GP($X$)} for both the lift and the drag: it confirms that the eigenshape basis $\mathcal V$ is more relevant than the CAD parameters basis for GP surrogate modeling. 

\subsubsection*{GP in reduced dimension: summary of results}
These four examples have proven the worth of the additive GPs: 
they are the models that perform the best because of the selection and prioritization of active variables. 
Models in reduced dimension that exclusively rely on the active eigenshapes provide accurate predictions too, but are slightly outperformed as they disregard smaller effects. 
GPs built in the space of all (informative) eigenshapes always outperform the ones built in the space of CAD parameters, even when both models have the same dimension. 
Among the GPs over the reduced space of $\delta$ first principal axes, further removing dimensions generally produces better predictions when the number of data points $n$ is small.  As $n$ increases, more eigenshapes lead to better metamodeling. 
Models where dimensions have been chosen only from a geometric criterion (the PCA) have a prediction quality that depends on the output: if the first modes do not impact $y$, as the 2 first eigenshapes of the rectangle problem, predictions are poor. Ignoring reconstruction details that affect the output as second-order eigenshapes in $f_{7L}$ also degrades the performance, highlighting the importance of finding the active variables that affect the output.

\section{Optimization in reduced dimension}
\label{section:optim_in_eigenbasis}
We now turn to the problem of finding the shape that minimizes an expensive objective function $f(\cdot)$. 
To this aim, we employ the previous additive GP, which works in the space of eigencomponents $\pmb\alpha$, in an
Efficient Global Optimization procedure \cite{jones1998efficient}:
at each iteration, a new shape is determined given the previous observations $\{(\pmb\alpha^{(1)},y_1),\dotsc,(\pmb\alpha^{(n)},y_n)\}$ by maximizing the Expected Improvement (EI, \cite{ExpectedImprovement,jones1998efficient}) as calculated with the GP $Y(\pmb\alpha)$: \begin{equation}\pmb\alpha^{(n+1)^*} = \arg \max_{\pmb\alpha\in \R^D} \text{EI}(\pmb\alpha;Y(\pmb\alpha))\text{,}\label{eq:EI_maximization}\end{equation}
\noindent where the EI is defined as
\begin{equation}
\text{EI}(\x;Y(\x))=(a-m(\x))\phi_\mathcal N\left(\frac{a-m(\x)}{s(\x)}\right)+s(\x)\varphi_\mathcal N\left(\frac{a-m(\x)}{s(\x)}\right)\text{.}
\label{eq:EI}
\end{equation}
\noindent $m(\cdot)$ and $s(\cdot)$ are the conditional mean and standard deviation of $Y(\cdot)$ (Equation (\ref{eq:formules_krigeage_additif})), respectively, while $\phi_{\mathcal N}$ and $\varphi_{\mathcal N}$ stand for the normal cumulative distribution function and probability density function. The threshold $a$ is usually set as the current minimum, $f_{\min}:=\underset{i=1,\dotsc,n}{\min}~y_i$, while other values have also been investigated \cite{jones2001taxonomy,gaudrie2018budgeted}.

\subsection{Alternative Expected Improvement Maximizations}
\label{sec:alternatives_ei_maximizaton}
\subsubsection*{Maximization in the entire $\pmb\alpha$ space}
The most straightforward way to maximize the EI is to consider its maximization in $\R^D$ as in Equation (\ref{eq:EI_maximization}).
However, this optimization is typically difficult as the EI is a multi-modal and high ($D$) dimensional function\footnote{As explained at the end of Section \ref{sec:CAD_to_eigenbasis}, we can restrict all calculations to $\pmb\alpha$'s $d'$ first coordinates. Even though $d'\ll D$, it has approximately the same dimension as $d$, hence the optimization is still carried out in a high dimensional space.}.

\subsubsection*{Maximization in the $\pmb\alpha^a$ space}
We can however take advantage of the dimension reduction beyond the construction of $Y(\cdot)$: 
$\pmb\alpha^a\in\R^\delta$ are the variables that affect $y$ the most and should be prioritized for the optimization of $f(\cdot)$. A second option is therefore to maximize the EI solely with respect to $Y^a(\pmb\alpha^a)$ in dimension $\delta$. 
This option is nonetheless incomplete as the full GP $Y(\cdot)$ requires the knowledge of $\pmb\alpha=[{\pmb\alpha}^a,{\pmb\alpha}^{\overline a}]$. 

A first simple idea to augment ${\pmb\alpha}^a$ is to set $\pmb\alpha^{\overline{a}}$ equal to its mean, $\mathbf{0}$.
The inactive part of the covariance matrix $\mathbf K_{\overline{a}}$ would be filled with the same scalar and the full covariance matrix $\mathbf K=\mathbf K_a+\mathbf K_{\overline{a}}$ would have a degraded conditioning. A second simple idea is to sample
$\pmb\alpha^{\overline{a}}\sim\mathcal N(\mathbf 0,\pmb\lambda_{\overline{a}})$.
However, $\pmb\alpha^{\overline{a}}$ act as local refinements to the shape that contribute a little to $y$, and should also be optimized. 
In \cite{li2019surrogate}, the authors observed that despite the gain in accuracy of surrogate models in a reduced basis (directions of largest variation of the gradient of the lift and drag in their application), a restriction to too few directions led to poorer optimizations since small effects could not be accounted for.

\subsubsection*{Optimization in $\pmb\alpha^a$ space complemented with a random embedding in $\pmb\alpha^{\overline{a}}$}
This leads to the third proposed EI maximization, which makes step \circled{5} in Figure \ref{fig:summary}: a maximization of the EI with respect to $\pmb\alpha^a$ and the use of a random embedding \cite{wang2013bayesian} to coarsely optimize the components $\pmb\alpha^{\overline a}$: 
EI$([\pmb\alpha^a,\overline\alpha\overline{\mathbf a}])$ 
is maximized, where $\overline\alpha\in\R$ is the coordinate along a random line in the $\pmb\alpha^{\overline a}$ space, $\overline{\mathbf a}=(\overline{\mathbf a}_{1},\dotsc,\overline{\mathbf a}_{D-\delta})^\top$.
Since $\pmb\alpha^{\overline a}$ have been classified as inactive, it is not necessary to make a large effort for their optimization. 
This approach can be viewed as an extension of REMBO \cite{wang2013bayesian}. 
In REMBO, a lower dimensional vector $\mathbf y\in\R^{\delta}$ is embedded in $X$ through a linear random embedding, $\mathbf y\mapsto\mathbf A_{R}\mathbf y$, where $\mathbf A_{R}\in\R^{D\times \delta}$ is a random matrix. 
Instead of choosing a completely random and linear embedding with user-chosen (investigated in \cite{binois2017choice}) effective dimension $\delta$, our embedding is nonlinear (effect of the mapping $\phi(\cdot)$), supervised and semi-random (choice of the active/inactive directions).
The dimension is no longer arbitrarily chosen since it is determined by the number of selected active components (Section \ref{sec:eigenshape_selection}), and the random part of the embedding is only associated to the inactive parts of $\pmb\alpha$: denoting $\pmb\alpha^a=(\alpha_{a_1},\dotsc,\alpha_{a_\delta})^\top$ the selected components (that are not necessarily the $\delta$ first axes) and $\pmb\alpha^a=(\alpha_{\overline{a}_1},\dotsc,\alpha_{\overline{a}_{D-\delta}})^\top$ the inactive ones, our embedding matrix $\mathbf A_{emb}\in\R^{D\times (\delta+1)}$ transforms $[\pmb\alpha^a,\overline\alpha]$ into the $\pmb\alpha$ space to which the $\x$'s are nonlinearly mapped. The $\delta$ first columns of $\mathbf A_{emb}$, $\mathbf A_{emb}^{(i)}$, $i=1,\dotsc,\delta$, correspond to $\pmb\alpha^a$ and contain the $\delta$ first vectors of the canonical basis of $\R^D$, $\mathbf e^{(i)}_D$,  i.e. $\mathbf A_{emb}^{(i)}=\delta_{{a_i}i}$, where $\delta_{ij}$ stands for the Kronecker symbol here, $\delta_{ij}=1$ if $i=j$, 0 else. The $\delta+1$-th column of $\mathbf A_{emb}$ contains $\overline{\mathbf a}$ in the rows which correspond to $\pmb\alpha^{\overline{a}}$, ${\mathbf A_{emb}^{(\delta+1)}}_{\overline{a}_i}=\overline{\mathbf a}_{i}$, $i=1,\dotsc,D-\delta$. Rows corresponding to active $\pmb\alpha$'s equal 0.

Assuming $\phi^{-1}$ exists, the proposed approach is the embedding of a lower dimensional design $[\pmb\alpha^a,\overline\alpha]$ whose dimension $\delta+1$ has carefully be chosen, in $X$, via the nonlinear and problem-related mapping $[\pmb\alpha^a,\overline\alpha]\mapsto\phi^{-1}(\mathbf V\mathbf A_{emb}[\pmb\alpha^a,\overline\alpha]+\overline{\pmb\phi})$. 
The approach can alternatively be considered as an affine mapping of $[\pmb\alpha^a,\overline\alpha]$ to the complete space spanned by the eigenshapes $\mathcal V$, 
\begin{equation}
[\pmb\alpha^a,\overline\alpha] \mapsto \mathbf V_{emb} [\pmb\alpha^a,\overline\alpha]+\overline{\pmb\phi}
\quad \text{ with }\quad
\mathbf V_{emb}:=\mathbf V\mathbf A_{emb}
\label{eq:Vembed}
\end{equation}
The shapes generated by the map of Equation (\ref{eq:Vembed}) are embedded in the space of all discretized shapes.
The columns of $\mathbf V_{emb}\in\R^{D\times(\delta+1)}$ associated to active components are the corresponding eigenshapes, while its last column is sum of the remaining eigenshapes, weighted by random coefficients, namely $\mathbf V \overline{\mathbf a}$, hence a supervised and semi-random embedding. 
Another difference to \cite{wang2013bayesian} is that only the EI maximization is carried out in the REMBO framework; the surrogate model is not built in terms of $[\pmb\alpha^a,\overline\alpha]$ but rather with the full $\pmb\alpha$'s via the additive GP (Section \ref{sec:additive_gp}).
	
In this variant, the EI maximization is carried out in a much more tractable $\delta+1$ -dimensional space and still has analytical gradients (see next section). 
From its optimum $\pmb\alpha^*=[{\pmb\alpha^{a}}^*,{\overline\alpha}^*]\in\R^{\delta+1}$ arises a $D$-dimensional vector, 
$\pmb\alpha^{(n+1)^*}=\mathbf A_{emb} [{\pmb\alpha^a}^*,{\overline\alpha}^*]$
to be evaluated by the true function (this is the pre-image problem discussed in Section \ref{sec:preimage}).

\begin{figure}[h!]
	\centering
	\includegraphics[width=0.6\textwidth]{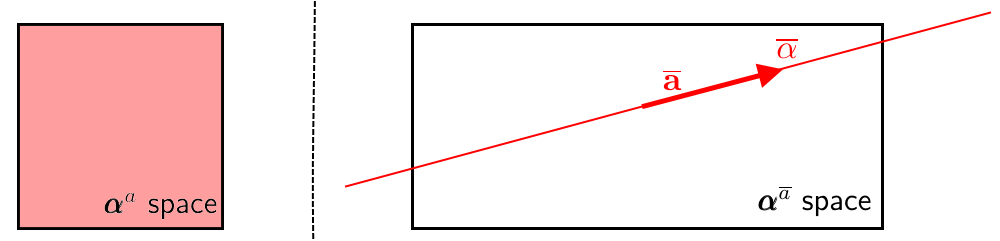}
	\caption{EI maximization in $\pmb\alpha^a$ complemented by the maximization along $\overline{\mathbf a}$, a random line in the $\pmb\alpha^{\overline{a}}$ space.}
	\label{fig:random_line}
\end{figure}

\subsubsection*{EI gradient in $\pmb\alpha$ space}
The Expected Improvement (\ref{eq:EI}) is differentiable and its derivative is known in closed-form \cite{roustant2012dicekriging}:
\begin{equation}
\nabla\text{EI}(\x)=-\nabla m(\x)\times\phi_{\mathcal N}(z(\x))+\nabla s(\x)\times\varphi_{\mathcal N}(z(\x))
\label{eq:gradEI}\text{,}
\end{equation}

where $z(\x)=(f_{\min}-m(\x))/s(\x)$. $\nabla m(\x)$ and $\nabla s(\x)$ require the gradient of $Y(\cdot)$'s kernel $k$ at $\x$, with the past observations $\x^{(1:n)}$, i.e. $\nabla k(\x,\x^{(1:n)})$, which is analytically computable. $\nabla s^2(\x)=2s(\x)\nabla s(\x)$ helps computing $s(\x)$'s gradient.

In the case of the additive GP (\ref{eq:modele_additif}), the mean and variance $m(\pmb\alpha)$ and $s^2(\pmb\alpha)$ are given by (\ref{eq:formules_krigeage_additif}). Using the notations of Section \ref{sec:additive_gp} and exploiting the symmetry of $\mathbf K$, few calculations lead to 
\begin{align}
\begin{split}
\nabla m(\pmb\alpha)=\nabla k(\pmb\alpha,\pmb\alpha^{(1:n)})^\top \mathbf K^{-1}({\mathbf y}_{1:n}-\mathbf1_n\widehat{\beta})\\
\nabla s(\pmb\alpha)=-\frac{\nabla k(\pmb\alpha,\pmb\alpha^{(1:n)})^\top\mathbf K^{-1}k(\pmb\alpha,\pmb\alpha^{(1:n)})}{s(\pmb\alpha)}
\label{eq:formules_gradient_krigeage_additif}
\end{split}
\end{align}

\noindent where $\nabla k(\pmb\alpha,\pmb\alpha^{(1:n)})=\sigma^2_a\nabla k_a(\pmb\alpha^a,\pmb\alpha^{a^{(1:n)}})+\sigma^2_{\overline{a}}\nabla k_{\overline{a}}(\pmb\alpha^{\overline{a}},\pmb\alpha^{{\overline{a}}^{(1:n)}})$, which are plugged in (\ref{eq:formules_gradient_krigeage_additif}) and in (\ref{eq:gradEI}) together with $z(\pmb\alpha)$'s expression to obtain $\nabla\text{EI}(\pmb\alpha;Y(\pmb\alpha))$.
In the alternatives proposed before, given an $\pmb\alpha\in\R^D$, the gradient of the EI can be computed efficiently, accelerating its maximization which is carried out by the genetic algorithm using derivatives \texttt{genoud} \cite{mebane2011genetic}. In the random embedding of $\overline\alpha$ case, the EI of $[\pmb\alpha^a,\overline\alpha]\in\R^{\delta+1}$ is given by EI$(\mathbf A_{emb}[\pmb\alpha^a,\overline\alpha];Y(\pmb\alpha))$, and its gradient by $\mathbf A_{emb}^\top\nabla\text{EI}(\mathbf A_{emb}[\pmb\alpha^a,\overline\alpha];Y(\pmb\alpha))$.

\subsubsection*{Setting bounds on $\pmb\alpha$ for the EI maximization}
As seen in the examples of Section \ref{sec:experiments_reduction}, neither the manifold of $\pmb\alpha$'s, nor its restriction to $\pmb\alpha^a$ need to be hyper-rectangular domains, which is a common assumption made by most optimizers such as \texttt{genoud} \cite{mebane2011genetic}, the algorithm used in our implementation. 
Two strategies were imagined to control the space in which the EI is maximized (\ref{eq:EI_maximization}): the first one is to restrict the EI maximization to $\mathcal A$ by setting it to zero for $\pmb\alpha$'s that are outside of the manifold. 
The benefit of this approach is that only realistic $\pmb\alpha$'s are proposed. 
But it might suffer from an incomplete description of the entire manifold of $\pmb\alpha$'s, $\mathcal A$, which is approximated by $\mathcal A_N$. Additionally, given $\mathcal A_N$, the statement ``being inside/outside the manifold'' has to be clarified. 
We rely on a nearest neighbor strategy in which the 95th quantile of the distances to the nearest neighbor within $\mathcal A_N$, $d_{0.95}$, is computed and used as a membership threshold: a new $\pmb\alpha$ is considered to belong to $\mathcal A$ if and only if the distance to its nearest neighbor within $\mathcal A_N$ is smaller than $d_{0.95}$. In the light of these limitations, a second strategy, in which the EI is maximized in $\mathcal A_N$'s covering hyper-rectangle, is also investigated.
The variant of EI maximization with embedding (random line in $\pmb\alpha^{\overline{a}}$), introduces an $\overline\alpha$ coordinate which has to be bounded too. 
The ${\overline\alpha}_{\min}$ and ${\overline\alpha}_{\max}$ boundaries are computed as the smallest and largest projection of $\mathcal A_N$ on $\overline{\mathbf a}$.
But depending on $\mathcal A_N$ and on $\overline{\mathbf a}$, this may lead to a too large domain since the embedded ${\overline\alpha}\overline{\mathbf a}$ might stay outside the $\pmb\alpha^{\overline{a}}$ covering hyper-rectangle.
In the spirit of \cite{binois2015warped}, to avoid this phenomenon, the largest ${\overline\alpha}_{\min}$ and the smallest ${\overline\alpha}_{\max}$ such that ${\overline\alpha}\overline{\mathbf a}$ belongs to the covering hyper-rectangle $\forall{\overline\alpha}\in[{\overline\alpha}_{\min} ,{\overline\alpha}_{\max} ]$, are chosen.

\subsubsection*{EI maximization via the CAD parameters}
A last option consists in carrying the maximization in the $X$ space through the mapping $\phi(\cdot)$ by $\underset{\x\in X}{\max}\text{ EI}(\x;Y(\underset{\pmb\alpha}{\underbrace{\mathbf V^\top(\phi(\x)-\overline{\pmb\phi})}}))=\text{ EI}(\x;Y(\pmb\alpha(\x)))$. 
This avoids both the aforementioned optimization domain handling and the pre-image search described in the following section. 
However, this optimization might be less efficient since it is a maximization in $d>\delta$ dimensions, and since $\nabla\phi(\x)$ is unknown, the EI loses the closed-form expression of its gradient.

\subsection{From the eigencomponents to the original parameters: the pre-image problem}
\label{sec:preimage}
The (often expensive) numerical simulator underlying the objective function can only take the original (e.g. CAD) parameters as inputs.
When the EI maximization is carried out in the eigencomponents space, 
the $\pmb\alpha$'s need to be translated into $\x$'s. To this aim, the \emph{pre-image} problem consists in finding the CAD parameter vector $\mathbf x$ whose description in the shape representation space $\Phi$
equals $\mathbf V \pmb\alpha^{(n+1)^*}+\overline{\pmb\phi}$.
Because there are more $\pmb\alpha$'s than $\x$'s, $D \gg d$, a strict equality may not hold and the pre-image problem is relaxed into:
\begin{equation}
\x^{(n+1)}=\underset{\x\in X}{\arg\min}\Vert(\phi(\x)-\overline{\pmb\phi})-\mathbf V\pmb\alpha^{(n+1)^*}\Vert^2_{\R^D}\text{.}\label{eq:pre_image}
\end{equation} 
To complete an iteration, the pre-image problem (\ref{eq:pre_image}) is solved and its solution $\x^{(n+1)}$, the parametric shape that resembles $\pmb\alpha^{(n+1)^*}$ the most, is evaluated by the simulator, which returns $y_{n+1}=f(\x^{(n+1)})$. Solving the pre-image problem does not involve calls to the simulator so that it is relatively not costly.
The surrogate model is then updated with $y_{n+1}$ and $\pmb\alpha^{(n+1)}:=\mathbf V^\top(\phi(\x^{(n+1)})-\overline{\pmb\phi})$, the $\x^{(n+1)}$ description in the $\mathcal V$ basis (step \circled{6} in Figure \ref{fig:summary}).

\begin{figure}[h!]
	\centering
	\includegraphics[width=0.6\textwidth]{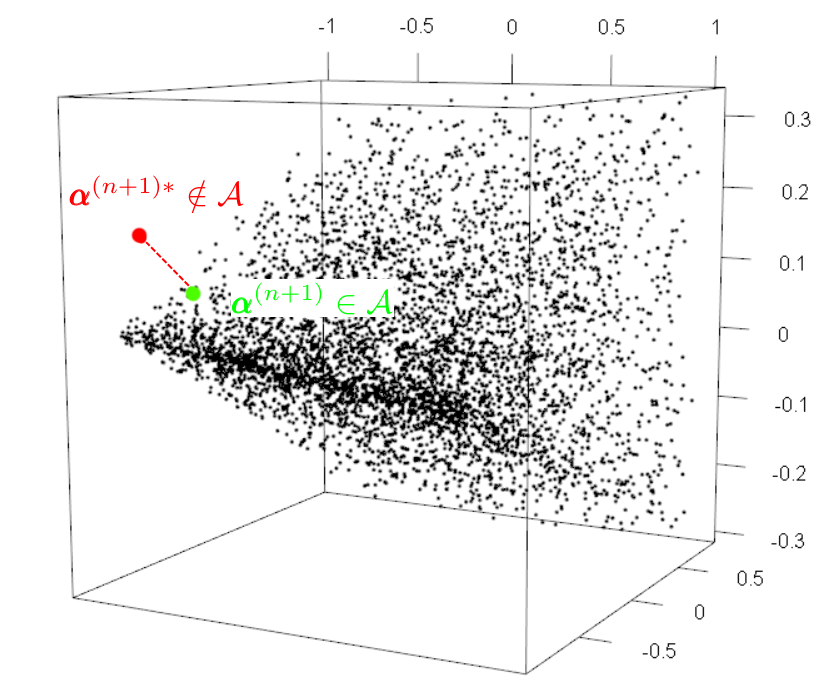}
	\caption{When $\pmb\alpha^{(n+1)^*}\notin\mathcal A$, the solution of the pre-image problem (in the $\pmb\alpha$ space), $\pmb\alpha^{(n+1)}$, is its projection on $\mathcal A$.}
	\label{fig:projection_manifold}
\end{figure}

Depending on the $\pmb\alpha^{(n+1)^*}$ yielded by the EI maximization (remember it may not stay on the manifold $\mathcal A$), 
$\pmb\phi^{(n+1)^*}:=\mathbf V\pmb\alpha^{(n+1)^*}+\overline{\pmb\phi}$ and $\pmb\phi^{(n+1)}:=\mathbf V\pmb\alpha^{(n+1)}+\overline{\pmb\phi}$, the shape representation of the $\pmb\alpha$ promoted by the EI and the shape representation of $\x^{(n+1)}$, respectively, may substantially differ. While it is mandatory to update the GP (\ref{eq:modele_additif}) with the pair $(\pmb\alpha^{(n+1)},y_{n+1})$, it may at first seem unclear what should be done with $\pmb\alpha^{(n+1)^*}$. When $\pmb\alpha^{(n+1)^*}$ does not belong to $\mathcal A$ and does not have a pre-image, it might seem straightforward to ignore it.
However, if $\pmb\alpha^{(n+1)^*}$ was yielded by the EI, it is very likely to be promoted in the following iterations, since its uncertainty, $s^2(\pmb\alpha^{(n+1)^*})$, has not vanished. Therefore, if $\pmb\phi^{(n+1)^*}$ and $\pmb\phi^{(n+1)}$ are substantially different,
the virtual pair $(\pmb\alpha^{(n+1)^*},y_{n+1})$ is included in the GP (\ref{eq:modele_additif}) too in a strategy called \emph{replication}. We define replication in general terms.
\begin{definition}[Replication]
\label{def:replication}
In Bayesian optimization, when the GP is built over coordinates $\pmb\alpha$ that are a mapping\footnote{In this article, the mapping $T(\cdot)$ is the composition of $\phi(\cdot)$ with the projection onto a subspace of $(\v^1,\dotsc,\v^D)$.} of the original 
coordinates $\x$, $\pmb\alpha = T(\x)$, at the end of each iteration a pre-image problem such as (\ref{eq:pre_image}) must be solved to translate the new acquisition criterion maximizer $\pmb\alpha^{(n+1)^*}$ into the next point to evaluate $\x^{(n+1)}$ and the associated iterate $\pmb\alpha^{(n+1)} = T(\x^{(n+1)})$. 
The \emph{replication} strategy consists in updating the GP with both $\left(\pmb\alpha^{(n+1)},f(\x^{(n+1)})\right)$ and $\left(\pmb\alpha^{(n+1)^*},f(\x^{(n+1)})\right)$ provided $\pmb\alpha^{(n+1)^*}$ and $\pmb\alpha^{(n+1)}$ are sufficiently different.
\end{definition}
Here, the difference between $\pmb\alpha^{(n+1)^*}$ and $\pmb\alpha^{(n+1)}$ is calculated as the distance between the associated shapes $\pmb\phi^{(n+1)^*}$ and $\pmb\phi^{(n+1)}$.
Since the database $\pmb\Phi$ contains the shape representation of $N$ distinct designs, $d_0:=\underset{\substack{i,j=1,\dotsc,N\\i\ne j}}{\min}~\Vert\pmb\Phi_i-\pmb\Phi_j\Vert_{\R^D}$, the minimal distance between two different designs in $\pmb\Phi$ is used as a threshold beyond which $\pmb\phi^{(n+1)}$ and $\pmb\phi^{(n+1)^*}$ are considered to be different. 
The replication strategy is further motivated by the fact that since $\x^{(n+1)}=\underset{\x\in X}{\arg\min}\Vert(\phi(\x)-\overline{\pmb\phi})-\mathbf V\pmb\alpha^{(n+1)^*}\Vert^2_{\R^D}=\underset{\x\in X}{\arg\min}\Vert\underset{\pmb\alpha(\x)}{\underbrace{\mathbf V^\top(\phi(\x)-\overline{\pmb\phi})}}-\pmb\alpha^{(n+1)^*}\Vert^2_{\R^D}$, 
where the last equality expresses just a change of basis since $\mathbf V$ is orthogonal, $\pmb\alpha^{(n+1)}$ is an orthogonal projection\footnote{Since we do not know the convexity of $\mathcal A$, the projection might not be unique.} of $\pmb\alpha^{(n+1)^*}$ on $\mathcal A$, see Figure \ref{fig:projection_manifold}. 

This is somehow similar to \cite{raghavan2013towards,raghavan2014numerical} where the authors project non realistic shapes on a smooth surface built via diffuse approximation or a local polynomial fitting, using the points of $\mathcal A_N$, to retrieve a realistic design. 
In our approach, unrealistic shape representations are directly projected onto $\mathcal A$ through the resolution of (\ref{eq:pre_image}). 
Incorporating the non physical $(\pmb\alpha^{(n+1)^*},y_{n+1})$ in the surrogate model can be viewed as an extension of the surrogate model outside its domain \cite{shahriari2016unbounded} (outside the manifold $\mathcal A$ in our case) by constant prolongation.

\subsection{Experiments}
The ideas developed in Section \ref{sec:CAD_to_eigenbasis}, \ref{section:GP_in_eigenbasis}, \ref{section:optim_in_eigenbasis} when put together make the method already sketched in Figure \ref{fig:summary} and more detailed in the following pseudo-code:

\begin{algorithm}[ht]	
	\caption{Pseudo-code of the Bayesian optimization in reduced eigencomponents, \texttt{AddGP($\pmb\alpha^a$+$\pmb\alpha^{\overline{a}}$)-EI embed with replication}.}
	\circled{1} Sample $N$ designs $\mathbf x^{(i)}$ and discretize them to form the matrix $\pmb\Phi$\tcc*[r]{see Section \ref{section:PCA}}
	
	\circled{2} Eigendecomposition of $\mathbf C_{\pmb\Phi}:=\frac1N(\pmb\Phi-\mathbf1_N\overline{\pmb\phi}^\top)^\top(\pmb\Phi-\mathbf1_N\overline{\pmb\phi}^\top)$ $\Rightarrow$ eigenvector basis $\mathcal V=\{\mathbf v^1,\dotsc,\mathbf v^D\}$ and principal components $\pmb\alpha=\mathbf V^\top(\phi(\mathbf x)-\overline{\pmb\phi})$;
	
	Evaluate $n$ designs $\mathbf x^{(1:n)}$ $\Rightarrow\mathbf y_{1:n}$, and compute their eigencomponents $\pmb\alpha^{(1:n)}$;

	\While{$n<$ computational budget}{
		
		\circled{3} Maximize $pl_\lambda(\pmb\alpha^{(1:n)},\mathbf y_{1:n},\vartheta)$ $\Rightarrow$ active and inactive eigencomponents, $\pmb\alpha=[\pmb\alpha^a,\pmb\alpha^{\overline{a}}]$\tcc*[r]{see Section \ref{sec:eigenshape_selection}}
		
		\circled{4} Build the additive GP $Y(\pmb\alpha)=\widehat{\beta}+Y^a(\pmb\alpha^a)+Y^{\overline{a}}(\pmb\alpha^{\overline{a}})$\tcc*[r]{see Section \ref{sec:additive_gp}}
		
		Randomly draw a vector of $D-\delta$ components, $\overline{\mathbf a}$
		
		\circled{5} Maximize the EI with respect to $[\pmb\alpha^a,\overline{\alpha}\overline{\mathbf a}]$ $\Rightarrow \pmb\alpha^{(n+1)^*}$ next shape to be evaluated\tcc*[r]{sharp maximization w.r.t. $\pmb\alpha^a$ and coarse maximization w.r.t. $\pmb\alpha^{\overline{a}}$, see Section \ref{sec:alternatives_ei_maximizaton}}
		
		\circled{6} Solve pre-image problem $\Rightarrow$ $\mathbf x^{(n+1)}$ to be evaluated $\Rightarrow y_{n+1}=f(\mathbf x^{(n+1)})$, and $\pmb\alpha^{(n+1)}=\mathbf V^\top(\phi(\mathbf x^{(n+1)})-\overline{\pmb\phi})$, associated eigencomponents\tcc*[r]{see Section \ref{sec:preimage}}
		
		\circled{6} \uIf{$\pmb\alpha^{(n+1)}$ and $\pmb\alpha^{(n+1)^*}$ too different}{Update the GP with $(\pmb\alpha^{(n+1)},y_{n+1})$ and $(\pmb\alpha^{(n+1)^*},y_{n+1})$\tcc*[r]{Replication, see Definition \ref{def:replication}}}
		\Else{Update the GP with $(\pmb\alpha^{(n+1)},y_{n+1})$;}
		
		$n\leftarrow n+1$;
	}
\end{algorithm}

Many algorithms result from the combination of versions of the GP metamodel and the EI maximization. 
They are related to the space in which these operations are performed (the initial $X$ or the eigencomponents $\mathcal A$ with the retained number of dimensions), the classical or additive GP, and the use of embedding or not. 
Before further explaining and testing them, we introduce a shorthand notation. The algorithms names are made of two parts separated by a dash, \texttt{GP version-EI version}. The GP part may either be an anisotropic GP with Matérn kernel, in which case it is noted \texttt{GP}, or an additive GP made of an anisotropic plus an isotropic kernel noted \texttt{AddGP}. The spaces on which they operate are specified in parentheses. 
For example, \texttt{GP($\pmb\alpha_{1:3}$)} is an anisotropic GP in the space spanned by $(\alpha_1,\alpha_2,\alpha_3)$, 
\texttt{AddGP($X_{1:3}+X_{4:40}$)} is an additive GP where the kernel is the sum of an anisotropic kernel in $(x_1,x_2,x_3)$ and an isotropic kernel in $(x_4,\dotsc,x_{40})$. The space over which the EI maximization is carried out is specified in the same way. Unspecified dimensions in the EI have their value set to the middle of their defining interval, e.g., \texttt{GP($X$)-EI($X_{1:2}$)} means that the EI maximization is done on the 2 first components of $\x$, the other ones being fixed to 0 if the interval is centered. The EI descriptor can also be a keyword characterizing the EI alternative employed (see Section \ref{sec:alternatives_ei_maximizaton}). For example,  \texttt{AddGP($\pmb\alpha_{1:2}+\pmb\alpha_{3:20}$)-EI embed} means that the EI is maximized in a 3 dimensional space made of $\alpha_1$, $\alpha_2$ and the embedding $\overline\alpha$.

\subsubsection{Optimization of a function with low effective dimension}
A set of experiments is now carried out that aims at comparing the three optimization alternatives involving GPs which have been introduced in Section \ref{sec:alternatives_ei_maximizaton} when a subset of active variables has been identified: EI maximization in the space of active variables, in the space of active variables with an embedding in the inactive space, and in the entire space. In order to test the EI maximization separately from the space reduction method (the mapping, PCA and regularized likelihood), we start by assuming that the effective variables are known. Complete experiments will be given later.

We minimize a function depending on a small number of parameters, the following modified version of the Griewank function \cite{molga2005test},
\begin{equation}
f_{\text{MG}}(\x)=f_{\text{Griewank}}(\x)+f_{\text{Sph}}(\x),~\x \in [-600,600]^{d}
\label{eq:griewank_40d}
\end{equation}
where $f_{\text{Griewank}}(\x)$ is the classical Griewank function in dimension 2,
\[f_{\text{Griewank}}(\x)=\frac{1}{4000}\sum_{j=1}^2x_j^2-\prod_{j=1}^{2}\cos(\frac{x_j}{\sqrt j})+1 \]
defined in $[-600,600]^2$ and whose optimum, located in $(0,0)^\top$, is 0. To create a high-dimensional 
function where only few variables act on the output, 
the $f_{\text{Sph}}$ function is added to $f_{\text{Griewank}}$, where $f_{\text{Sph}}$ is a sphere centered in $\mathbf c$, with smaller magnitude than $f_{\text{Griewank}}$, and which only depends on the variables $x_3,\dotsc,x_{10}$:
\[f_{\text{Sph}}(\x)=\frac{1}{400,000}\sum_{j=3}^{10}(x_j-c_{j-2})^2\text{.}\]
$f_{\text{Sph}}(\x)$ is the squared Euclidean distance between $(x_3,\dotsc,x_{10})^\top$ and $\mathbf c$ which is set to\\\noindent$\mathbf c=(-140,-100,-60,-20,20,60,100,140)^\top$ in our experiments. Completely ignoring $(x_3,\dotsc,x_{10})^\top$ therefore does not lead to the optimum of $f_{\text{MG}}$. 
We define $f_\text{MG}$ in $[-600,600]^d$, $d\ge10$: the variables $x_{11},\dotsc,x_d$ do not have any influence on $f_\text{MG}$ but augment the dimension. In the following experiments, we take $d=40$.

The additive GP described in Section \ref{sec:additive_gp} operates between the active space composed of $x_1$ and $x_2$, and the inactive space of $X_{3:d}$.
With the additive GP, three ways to optimize the EI are investigated: \texttt{AddGP($X_{1:2}+X_{3:40}$)-EI($X_{1:2}$)} where the EI is optimized along the active space only and $x_3,\dotsc, x_d$ are set to the middle of their intervals ($\mathbf 0$), 
\texttt{AddGP($X_{1:2}+X_{3:40}$)-EI embed} where the EI is optimized in the active space completed by the embedding in the inactive space, 
and \texttt{AddGP($X_{1:2}+X_{3:40}$)-EI($X$)} where the EI is optimized in the entire $X$.
These Bayesian optimization algorithms with additive GPs are compared to three classical optimizers: 
one based on the GP built in the entire space (\texttt{GP($X$)-EI($X$)}), 
another based on the building of the GP in the $X_{1:2}:=(x_1,x_2)$ space (\texttt{GP($X_{1:2}$)-EI($X_{1:2}$)}), 
and one working in the $X_{1:10}:=(x_1,\dotsc,x_{10})$ space (\texttt{GP($X_{1:10}$)-EI($X_{1:10}$)}).

We start the experiments with an initial DoE of $n=20$ points, which is space-filling in $X$ (or in $X_{1:2}$ or $X_{1:10}$ for the variants where the metamodel is built in these spaces). We then try to find the minimum of $f_{\text{MG}}$, $\x^*:=(0,0,\mathbf c,\pmb *)$ in the limit of $p=80$ iterations\footnote{that is to say EI maximizations, whose optima are evaluated by $f_\text{MG}$.}. For the instance where the metamodel is built in $X\subset\R^{40}$, we cannot start with an initial DoE of $n=20$ points, and the experiments are initialized with $n=50$ designs, only $p=50$ iterations being allowed. The EI being maximized by the genetic algorithm \texttt{genoud} \cite{mebane2011genetic}, we use the same population and number of generations in each variant for fair comparison.

The lowest objective function values obtained by the algorithms are reported in Table \ref{tab:resu_EI_griewank}. 
They are averaged over 10 runs with different initial designs, and standard deviations are given in brackets. The left-hand side columns correspond to standard GPs carried out in different spaces, and the right-hand side columns correspond to runs using the additive GP of Section \ref{sec:additive_gp} together with different EI maximization strategies.

\begin{table}[!ht]
	\centering
	\makebox[\textwidth][c]{
		\begin{tabu}{|c|c|c|c|c|c|c|}
			\hline
			\multirow{2}{*}{Metamodel} & \multicolumn{3}{c|}{Standard GP} & \multicolumn{3}{c|}{\multirow{1}{*}{Additive GP }}\\
			 & \multicolumn{1}{c}{\texttt{GP($X_{1:2}$)-}} & \multicolumn{1}{c}{\texttt{GP($X_{1:10}$)-}} & \multicolumn{1}{c|}{\texttt{GP($X$)-}} & \multicolumn{3}{c|}{\texttt{AddGP($X_{1:2}+X_{3:40}$)-}}
			\\\hline
			EI maximization & \texttt{EI($X_{1:2}$)} & \texttt{EI($X_{1:10}$)} & \texttt{EI($X$)} & \texttt{EI($X_{1:2}$)} & \texttt{EI embed} & \texttt{EI($X$)}\\\hline
			Optimum (sd) & 0.776 (0.221) & 1.127 (0.214) & 0.669 (0.280) & 0.545 (0.210) & 0.481 (0.185) & 0.986 (0.366)\\\hline
		\end{tabu}
	}
	\caption{Objective function values obtained within 100 (20+80 or 50+50 for the third column) evaluations of the 40-dimensional $f_\text{MG}$, with different metamodels and varying EI maximization strategies.
	}
	\label{tab:resu_EI_griewank}
\end{table}

The results in Table \ref{tab:resu_EI_griewank} show that the methods using the additive GP usually outperform those where the GP is built in a more or less truncated $X$ space. 
The results of \texttt{GP$(X_{1:10})$-EI($X_{1:10}$)} are surprisingly bad. Additional experiments have shown that they seem to be linked with a too small initial DoE. 
Notice that with another version of $f_\text{MG}$ (where $\mathbf c$ is closer to the boundaries of $X_{3:10}$, not reported here), \texttt{GP$(X_{1:10})$-EI($X_{1:10}$)} outperforms \texttt{GP$(X_{1:2})$-EI($X_{1:2}$)} and the classical GP$(X)$-EI($X$), 
which is normal since in this situation $X_{3:10}$ become active.
However, the \texttt{AddGP-EI embed} and \texttt{AddGP-EI($X$)} versions with the additive GP remain better.

The maximization of the EI for the additive GP between the active and inactive components performs the best when the maximization strategy combines the advantage of a low-dimensional active space with a rough maximization in the larger inactive subspace, the \texttt{AddGP-EI embed} strategy. 
It is also worth mentioning that it is the variant with lowest standard deviation. \texttt{AddGP-EI($X$)}, searching in a 40 dimensional space, is not able to attain the optimum as well. Even though it is carried out in a very small dimension, \texttt{AddGP-EI($X_{1:2}$)} is also slightly outperformed by \texttt{AddGP-EI embed}, because it cannot optimize the $\x^{\overline{a}}$'s. 
In this instance of $f_{\text{MG}}$ where $\mathbf c$ is relatively close to $\mathbf 0$, \texttt{AddGP-EI($X_{1:2}$)} does not suffer to much from disregarding $\x^{\overline{a}}$'s.
However, in the additional experiment where $\mathbf c$ is close to the boundaries of $X_{3:10}$, \texttt{AddGP-EI($X_{1:2}$)} exhibits poor results, while \texttt{AddGP-EI embed} still performs well. 
In this case \texttt{AddGP-EI($X$)} performs slightly better than \texttt{AddGP-EI embed}, because it benefits from the maximization over the complete $X$ while the restriction on $\overline x$ hinders \texttt{AddGP-EI embed} to get as close to the solution, but \texttt{AddGP-EI embed} still performs reasonably well and has a smaller standard deviation than \texttt{AddGP-EI($X$)}. 
For all these reasons, the additive GP with random embedding (\texttt{AddGP-EI embed}) strategy is assessed as the safest one.

\subsubsection{Experiments with shape optimization}
\label{sec:experiments_optimization}

We now turn to the shape optimization of the designs introduced in Section \ref{sec:experiments_reduction} whose objective functions were defined in Section \ref{sec:metamodeling_shape_eigenbasis}. 
We compare the standard approach where the designs are optimized in the CAD parameters space with the methodologies where the surrogate model is built in the eigenshape basis (all variants described in Section \ref{sec:alternatives_ei_maximizaton}). 
For fair comparison, the same computational effort is put on the internal EI maximization.

\subsubsection*{Catenoid shape}
We want to find a curve $r(y)$ which minimizes the associated axisymmetric surface as expressed by the integral making $f_5(\x)$ in the catenoid problem (Example \ref{ex:catenoide}).

The different versions of Bayesian optimizers that are now tested are the following:
\begin{itemize}
		\item the standard \texttt{GP($X$)-EI($X$)} where both the GP and the EI work with the original $x$'s, i.e. CAD parameters;
		\item \texttt{GP($\pmb\alpha_{\text{\_\_}}$)-EI($\pmb\alpha_{\text{\_\_}}$)} indicates the GP is built in the space of $\text{\_\_}$ (to be specified) principal components over which the EI is maximized;
		\_\_ are taken equal to 1:4 and 1:7 because, as seen in Table \ref{tab:eigenvalues_catenoide}, 4 and 7 eigencomponents account for 98\% and all of the shape variance, respectively.
		\item \texttt{GP($\pmb\alpha_{\text{\_\_}}$)-EI($X$)} indicates the GP is built in the space of $\text{\_\_}$ principal components but the EI is maximized in the $X$ space;
		\item \texttt{AddGP($\pmb\alpha^a$+$\pmb\alpha^{\overline{a}}$)} refers to the additive GP, for which three EI maximizations have been described (Section \ref{sec:alternatives_ei_maximizaton}): \texttt{EI embed} where $\pmb\alpha^a$ and an embedding in the $\pmb\alpha^{\overline{a}}$ space is maximized, \texttt{EI($\pmb\alpha^a$)} where only the actives $\pmb\alpha$'s are maximized (the remaining ones being set to their mean value in $\mathcal A_N$, $\mathbf 0$), and \texttt{EI($\pmb\alpha$)} where all $\pmb\alpha$'s are maximized;
		\item \texttt{GP($\pmb\alpha^a$)-EI($\pmb\alpha^a$)} means the GP is built over the space of active $\pmb\alpha$'s, over which the EI maximization is carried out.
\end{itemize}
Regarding the EI maximization in $\mathcal A$, \texttt{on manifold} states that the search is restricted to $\pmb\alpha$'s close to $\mathcal A_N$. If not, the maximization is carried out in $\mathcal A_N$'s covering hyper-rectangle, and \texttt{with replication} indicates that both $\pmb\alpha^{(n+1)}$ and ${\pmb\alpha^{(n+1)}}^*\notin\mathcal A$ are used for the metamodel update, while \texttt{no replication} indicates that only the $\pmb\alpha^{(n+1)}$'s are considered by the surrogate.

\vskip\baselineskip

The best objective function values obtained by the algorithms are reported in Table \ref{tab:optim_catenoide}. 
They are averaged over 10 runs with different initial DoEs, and standard deviations are given in brackets. The algorithms start with a space-filling DoE of 20 individuals and are run for 60 additional iterations. 
In the case of the CAD parameters, since $d=29>20$, the initial DoE contains 40 designs and the algorithm is run for 40 iterations. 
The number of function evaluations to reach certain levels is also reported, to compare the ability of the algorithms to quickly attain near-optimal values. 
When at least one run has not reached the target, a rough estimator of the empirical runtime \cite{auger2005performance}, $\overline{T_s}/p_s$, is provided in red, the number of runs achieving the target value being reported in brackets. $\overline{T_s}$ and $p_s$ correspond to the average number of function evaluations of runs that reach the target and the proportion of runs attaining it. 

\begin{table}[!ht]
	\centering
	\makebox[\textwidth][c]{	
		\begin{tabular}{|c|c|c|c|c|}
			\hline
			Method & Best value & Time to 27 & Time to 30 & Time to 35\\\hline
			\texttt{GP($X$)-EI($X$)} & 31.83 (2.10) & $\times$ & {\color{red}570.0 [1]} & 68.5 (9.9)\\
			\texttt{GP($\pmb\alpha_{1:7}$)-EI($\pmb\alpha_{1:7}$) on manifold} & 26.93 (0.18) & {\color{red}86.9 [7]} & 40.2 (10.5) & 40.2 (10.5)\\
			\texttt{GP($\pmb\alpha_{1:7}$)-EI($\pmb\alpha_{1:7}$) with replication} & 26.16 (0.10) & 30.5 (2.8) & 24.3 (0.8) & 23.4 (0.5)\\
			\texttt{GP($\pmb\alpha_{1:7}$)-EI($\pmb\alpha_{1:7}$) no replication} & 27.62 (0.72) & {\color{red}147.5 [2]} & 25.4 (2.5) & 23.5 (0.5)\\
			\texttt{GP($\pmb\alpha_{1:7}$)-EI($X$)} & 40.57 (11.61) & {\color{red}370.0 [1]} & {\color{red}163.3 [3]} & {\color{red}120.0 [4]}\\
			\texttt{AddGP($\pmb\alpha^a$+$\pmb\alpha^{\overline{a}}$)-EI embed on manifold} & 50.67 (0.05) & $\times$ & $\times$ & $\times$\\
			\texttt{AddGP($\pmb\alpha^a$+$\pmb\alpha^{\overline{a}}$)-EI embed no replication} & 27.58 (0.53) & {\color{red}172.5 [2]} & 23.6 (1.4) & 22.3 (0.7)\\
			\texttt{AddGP($\pmb\alpha^a$+$\pmb\alpha^{\overline{a}}$)-EI embed with replication} & 26.19 (0.16) & 28.4 (4.1) & 24.2 (3.1) & 22.8 (1.9)\\
			\texttt{GP($\pmb\alpha_{1:4}$)-EI($\pmb\alpha_{1:4}$) with replication} & 27.12 (0.13) & {\color{red}550.0 [1]} & 27.0 (3.9) & 25.4 (3.8)\\\hline
		\end{tabular}
	}
	\caption{Best objective function values found and number of iterations required to attain a fixed target (average over 10 runs, standard deviations in brackets) for different metamodels and optimization strategies, on the catenoid problem (Example \ref{ex:catenoide}). Red figures correspond the empirical runtime, with the number of runs which attained the target in brackets, and '$\times$' signifies that no run was able to attain it within the limited budget.
}
	\label{tab:optim_catenoide}
\end{table}

Comparing the results in Table \ref{tab:optim_catenoide} of the algorithms that stay on the manifold with the others indicates that restricting the search of EI maximizers to the vicinity of $\mathcal A_N$ worsens the convergence. Indeed, promising $\pmb\alpha$'s are difficult to attain or are even falsely considered as outside $\mathcal A$. 
This observation gets even worse with the additive GP: staying in the neighborhood of $\mathcal A_N$ has even stronger consequences because of the restriction to the random line $\overline{\mathbf a}$. The EI should therefore be optimized in the covering hyper-rectangle of $\mathcal A_N$.

\begin{figure}[h!]
	\centering
	\includegraphics[width=0.6\textwidth]{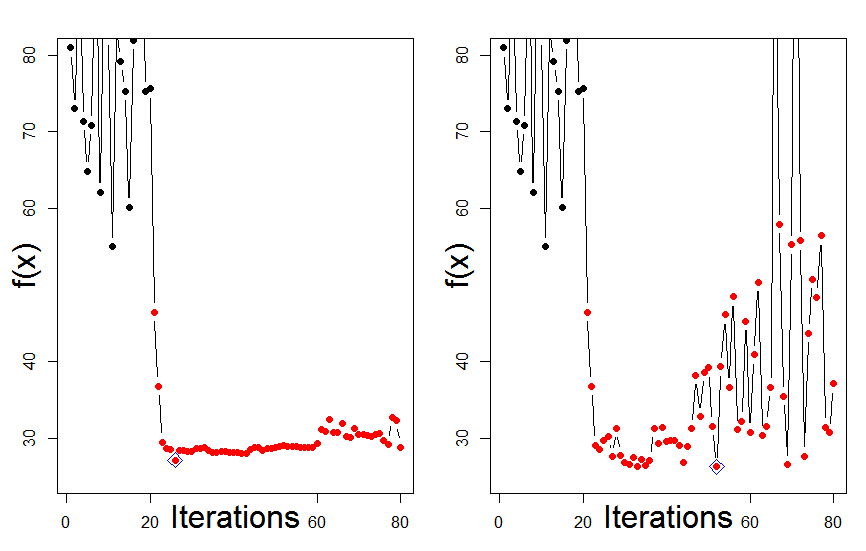}
	\caption{Optimization with EI maximization in the covering hyper-rectangle of $\mathcal A_N$ without (left) or with (right) replication strategy.}
	\label{fig:comparison_replication_strategy}
\end{figure}

For tackling the issue of EI maximizers $\pmb\alpha^{(n+1)*}\notin\mathcal A$, the replication strategy exhibits better performance than the strategy where only the projection, $\pmb\alpha^{(n+1)}$, is used for updating the GP. 
Figure \ref{fig:comparison_replication_strategy} shows the typical effect of the replication strategy. On the left, the inner EI maximization is carried out in the covering hyper-rectangle of $\mathcal A_N$ but only the $\pmb\alpha\in\mathcal A$ obtained through the pre-image problem solving are used to construct the surrogate model. 
On the right, all EI maximizers have been used for the GP, including $\pmb\alpha\notin\mathcal A$. 
Without replication, since the variance of the GP at previous EI maximizers has not vanished, the EI continues promoting the same $\pmb\alpha$'s, which have approximately the same pre-image. 
The same part of the $\pmb\alpha$ space is sampled, which not only leads to a premature convergence (the best observed value has already been attained after 6 iterations), but also increases the risk of getting a singular covariance matrix.
With the replication, the GP variance vanishes for all EI maximizers, even those outside $\mathcal A$, removing any further EI from these $\pmb\alpha$'s. The $\pmb\alpha$ space is better explored with benefits on the objective value (26.26 against 27.13 here).

The EI strategy which consists in maximizing via the $X$ space of CAD parameters avoids the $\pmb\alpha$ manifold issues.
However, it does not perform well, because of the higher dimensional space where the criterion is maximized. An additional drawback for efficient maximization is that $\nabla$EI is not known analytically in this case.

In this catenoid example, the additive GP and the GP in the space of (all) 7 principal components achieve comparable results, both in terms of best value, and of function evaluations to attain the targets. Indeed, the true dimension (7) is relatively low, and we have noticed that the 5, 6 or even 7 first eigenshapes often got classified as active for the additive GP.

\subsubsection*{Heart rectangle}
We now consider Example \ref{ex:coeur} and the minimization of $f_4(\x)$ that expresses the distance from a shape to a rectangle deformed as an heart.

As before, different metamodeling and EI maximization options are benchmarked. 
They include: the standard approach of doing the process in the space of CAD parameters (in dimension $d=40$); the optimization in the space of 2, 4, 16 or 40 first principal components, where 100\% of the shapes variability is recovered with 40 eigencomponents as seen in Table \ref{tab:eigenvalues_rectangle}. 
Supervised eigenshape selection methods (Section \ref{sec:supervised_dimension_reduction}) are also used: the GP built over $\pmb\alpha^a$ only, and the additive model over $\pmb\alpha^a$ and $\pmb\alpha^{\overline{a}}$. 
For the latter, the 4 EI maximization options of Section \ref{sec:alternatives_ei_maximizaton} are compared. 
In light of the above optimization results on the catenoid, the three EI maximization strategies are carried out in the covering hyper-rectangle of $\mathcal A_N$ (as opposed to restricted to the neighborhood of $\mathcal A_N$), and EI maximizers which do not belong to $\mathcal A$ are nonetheless used for the GP update. Henceforth, the \texttt{with replication} strategy becomes the new default in all algorithms carrying out EI maximizations in $\pmb\alpha$'s and it will no longer be specified in the algorithms names.

The statistics on the solutions proposed by the algorithms are reported in Table \ref{tab:optim_heart}. 
They consist in the best objective function values averaged over 10 runs with different initial designs, with standard deviations given in brackets. 
The average and standard deviation of the number of function evaluations to reach certain levels is also given, to compare the ability of the algorithms to quickly attain near-optimal values. When at least one run failed in attaining the target, it is replaced by a rough estimator of the empirical runtime.
The algorithms start with a space-filling DoE of 20 individuals and are run for 80 supplementary iterations. 
In the case of the CAD parameters \texttt{GP($X$)-EI($X$)}
and of \texttt{GP($\pmb\alpha_{1:40}$)-EI($\pmb\alpha_{1:40}$)}, since $d=40>20$, the initial DoE contains 50 designs and the algorithm is run for 50 iterations. 

\begin{table}[!ht]
	\centering
	\makebox[\textwidth][c]{	
		\begin{tabular}{|c|c|c|c|c|}
			\hline
			Method & Best value & Time to 0.5 & Time to 1 & Time to 3\\\hline
			\texttt{GP($X$)-EI($X$)} & 1.18 (0.45) & $\times$ & {\color{red}166.9 [4]} & 42.1 (26.5)\\
			\texttt{GP($\pmb\alpha_{1:2}$)-EI($\pmb\alpha_{1:2}$)} & 9.21 (0.80) & $\times$ & $\times$ & $\times$\\
			\texttt{GP($\pmb\alpha_{1:4}$)-EI($\pmb\alpha_{1:4}$)} & 0.33 (0.07) & 48.8 (21.8) & 21.8 (2.2) & 21.0 (0.0)\\
			\texttt{GP($\pmb\alpha_{1:16}$)-EI($\pmb\alpha_{1:16}$)} & 0.59 (0.15) & {\color{red}197.8 [3]} & 50 (15.4) & 35.0 (9.7)\\
			\texttt{GP($\pmb\alpha_{1:40}$)-EI($\pmb\alpha_{1:40}$)} & 2.95 (0.97) & $\times$ & $\times$ & {\color{red}194.4 [5]}\\
			\texttt{GP($\pmb\alpha^a$)-EI($\pmb\alpha^a$)} & 0.32 (0.09) & 33.7 (9.4) & 24.5 (3.7) & 21.8 (1.3)\\
			\texttt{AddGP($\pmb\alpha^a$+$\pmb\alpha^{\overline{a}}$)-EI($X$)} & 0.54 (0.19) & {\color{red}199.4 [4]} & 40.2 (12.3) & 30.2 (10.5)\\
			\texttt{AddGP($\pmb\alpha^a$+$\pmb\alpha^{\overline{a}}$)-EI embed} & 0.37 (0.08) & 49.0 (21.4) & 26.1 (5.6) & 22.2 (1.9)\\
			\texttt{AddGP($\pmb\alpha^a$+$\pmb\alpha^{\overline{a}}$)-EI($\pmb\alpha^a$)} & 0.37 (0.09) & 33.3 (14.6) & 22.7 (2.6) &  21.4 (0.7)\\
			\texttt{AddGP($\pmb\alpha^a$+$\pmb\alpha^{\overline{a}}$)-EI($\pmb\alpha$)} & 0.60 (0.26) & {\color{red}106.7 [6]} & {\color{red}41.2 [9]} & 21.5 (0.5)\\\hline
		\end{tabular}
	}
	\caption{Minimum objective function values found and number of function evaluations required to attain a fixed target (average over 10 runs, standard deviations in brackets) for different metamodels and optimization strategies, rectangular heart problem (Example \ref{ex:coeur}). The red figures correspond the empirical runtime, with the number of runs which attained the target in brackets, and '$\times$' signifies that no run was able to attain it within the limited budget. All algorithms performing an EI search in $\pmb\alpha$'s do it \texttt{with replication}, the henceforth default.}
	\label{tab:optim_heart}
\end{table}

In this test case, as shown in Figure \ref{fig:coeur_eigenshapes}, the 2 first eigenshapes modify the shape's position, to which $f_4$ is insensitive. 
Poor results are therefore obtained by \texttt{GP($\pmb\alpha_{1:2}$)-EI($\pmb\alpha_{1:2}$)} even though $\v^1$ and $\v^2$ account for 80\% of shape reconstruction, highlighting the benefits of the determination of active eigenshapes. 
In a first order approximation, $\mathbf v^3$ and $\mathbf v^4$ are the most influential eigenshapes with regard to $f_4$, which measures the nodal difference between $\Omega_{\x}$ and the target $\Omega_{\mathbf t}$. 
\texttt{GP($\pmb\alpha_{1:4}$)-EI($\pmb\alpha_{1:4}$)} exhibits very good results, as well as \texttt{GP($\pmb\alpha^a$)-EI($\pmb\alpha^a$)}, which mainly selects $\mathbf v^3$ and $\mathbf v^4$ ($\mathbf v^1$, $\mathbf v^2$ and other eigenshapes are sometimes selected too). 
Even though the shape reconstruction is enhanced, \texttt{GP($\pmb\alpha_{1:16}$)-EI($\pmb\alpha_{1:16}$)} and \texttt{GP($\pmb\alpha_{1:40}$)-EI($\pmb\alpha_{1:40}$)} have poor results because of the increase in dimension which is not accompanied by additional information, as already pointed out during the comparison of the predictive capability of these GPs for small budgets, see Table \ref{tab:R2_coeur}. 
\texttt{GP($\pmb\alpha_{1:40}$)-EI($\pmb\alpha_{1:40}$)} performed better than \texttt{GP($X$)-EI($X$)} in Table \ref{tab:R2_coeur}, yet its optimization performance is decreased. 
This is certainly due to the initial DoE: both DoEs are space-filling in their respective input space ($X$ or the hyper-rectangle of $\pmb\alpha\in\mathcal A$ containing $\mathcal A_N$). 
However, there is a significant difference between the minima in these DoEs: 
the average minimum over the 10 runs was 2.57  for \texttt{GP($X$)-EI($X$)} (hence better than the eventual average best value for \texttt{GP($\pmb\alpha_{1:40}$)-EI($\pmb\alpha_{1:40}$)}), and 9.22 for \texttt{GP($\pmb\alpha_{1:40}$)-EI($\pmb\alpha_{1:40}$)}. While GPs built over the entire $\pmb\alpha$ space (e.g. the additive one) suffer from the same drawback, the selection of variables identifies the dimensions to focus on to rapidly decrease the objective function. 
This remark applies only to the rectangular heart test case and one may wonder what level of generality it contains.
Contrarily to the previous example where building the GP in the space of all (informative) eigenshapes led to the best results, this strategy (\texttt{GP($\pmb\alpha_{1:40}$)-EI($\pmb\alpha_{1:40}$)}) performs weakly here because of the higher dimension.

The variants of the additive GP perform well too but they are slightly outperformed by \texttt{GP($\pmb\alpha_{1:4}$)-EI($\pmb\alpha_{1:4}$)}.
As the objective function mainly depends on $\v^3$ and $\v^4$, always classified as active, strategies that do not put too much emphasis or that neglect $\pmb\alpha^{\overline{a}}$ (namely, \texttt{AddGP($\pmb\alpha^a+\pmb\alpha^{\overline{a}}$)-EI embed} and \texttt{AddGP($\pmb\alpha^a+\pmb\alpha^{\overline{a}}$)-EI($\pmb\alpha^a$)}) perform the best. This explains the good performance of \texttt{GP($\pmb\alpha_{1:4}$)-EI($\pmb\alpha_{1:4}$)}, which disregards $\alpha_5,\dotsc,\alpha_{40}$. The maximization of the EI with respect to the full $\pmb\alpha$ is hindered by the high dimension. Again, the performance decreases when the EI is maximized via the $X$ space. 
\texttt{AddGP($\pmb\alpha^a+\pmb\alpha^{\overline{a}}$)-EI embed} and \texttt{GP($\pmb\alpha_{1:4}$)-EI($\pmb\alpha_{1:4}$)} need more iterations to attain good values (smaller than 0.5) than \texttt{GP($\pmb\alpha^a$)-EI($\pmb\alpha^a$)} and \texttt{AddGP($\pmb\alpha^a+\pmb\alpha^{\overline{a}}$)-EI($\pmb\alpha^a$)} which are early starters. This might be due to the additional though less critical components ($\overline{\alpha}$ or $\alpha_1,\alpha_2$, respectively) considered by these methods.

\subsubsection*{NACA 22 optimization}
In this last test case, we compare two of the aforementioned algorithms by optimizing the lift coefficient and the drag coefficient of a NACA 22 airfoil ($f_{7L}$ and $f_{7D}$).
The simulation is made with a computational fluids dynamic code that solves the Reynolds Averaged Navier-Stokes (RANS) equations with $k-\varepsilon$ turbulence model.
Since a single call to the simulator (one calculation of $f_7$) takes about 20 minutes on a standard personal computer, only two runs are compared for each objective. 
The first algorithm is the classical Bayesian optimizer where the GP is built in CAD parameter space, \texttt{GP($X$)-EI($X$)}. 
In the second algorithm, \texttt{AddGP($\pmb\alpha^a+\pmb\alpha^{\overline{a}}$)-EI embed}, the GP is built in the $\mathcal V$ basis of eigenshapes, while prioritizing the active dimensions, $\pmb\alpha^a$, via the additive GP and the EI random embedding method with the replication option, see Section \ref{sec:preimage}.
The optimization in the eigenshape basis starts with a DoE of $n=10$ designs and is run for $p=90$ additional iterations while, because there are 22 $x_i$'s, the optimization in the CAD parameters space starts using $n=50$ designs and is run for $p=50$ iterations.

\begin{figure}[h!]
	\centering
	\includegraphics[width=0.33\textwidth]{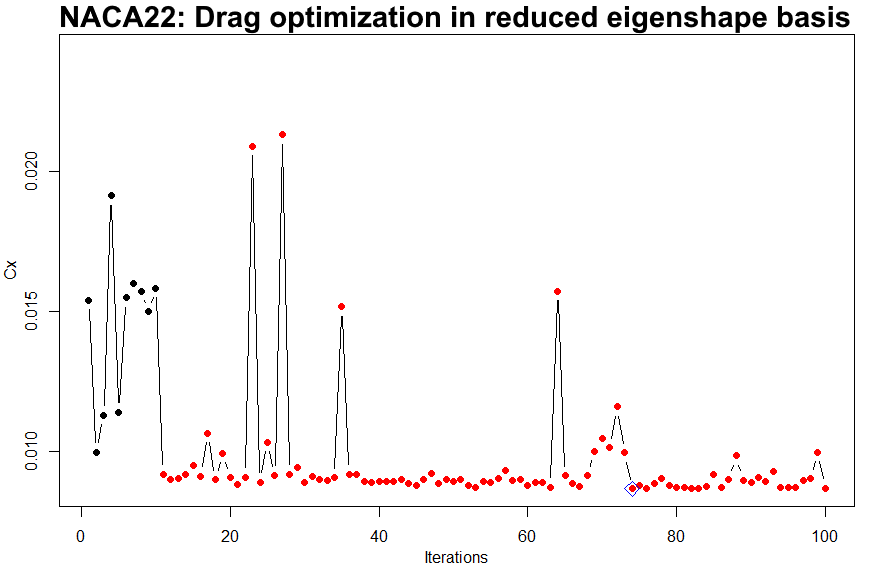}
	\includegraphics[width=0.33\textwidth]{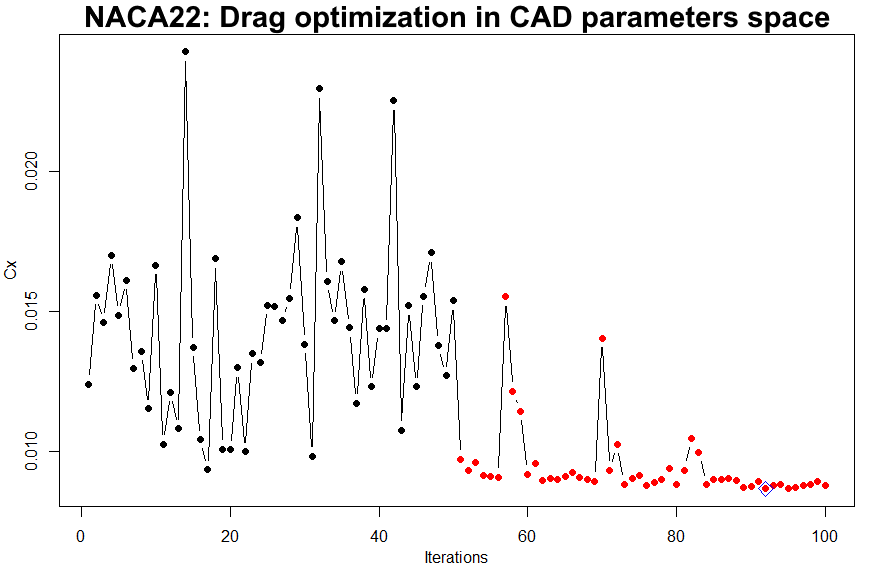}
	\includegraphics[width=0.32\textwidth]{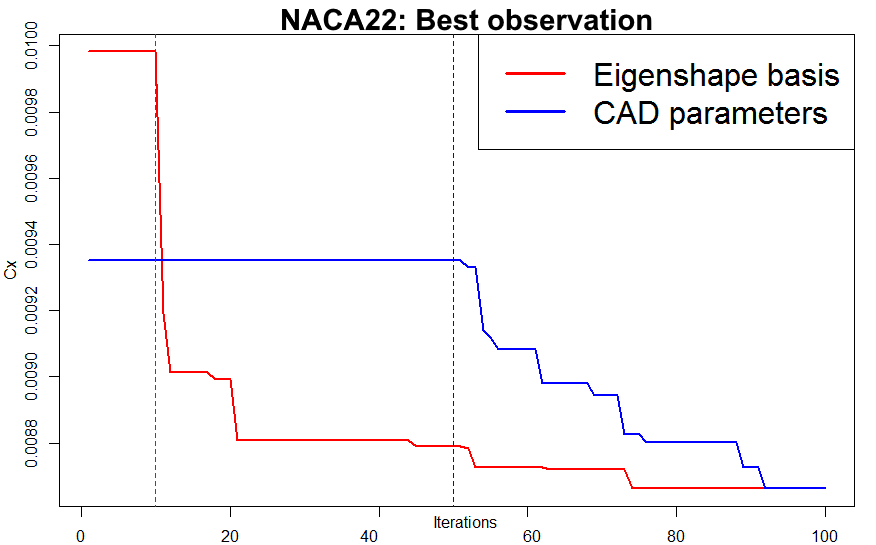}\\
	\includegraphics[width=0.33\textwidth]{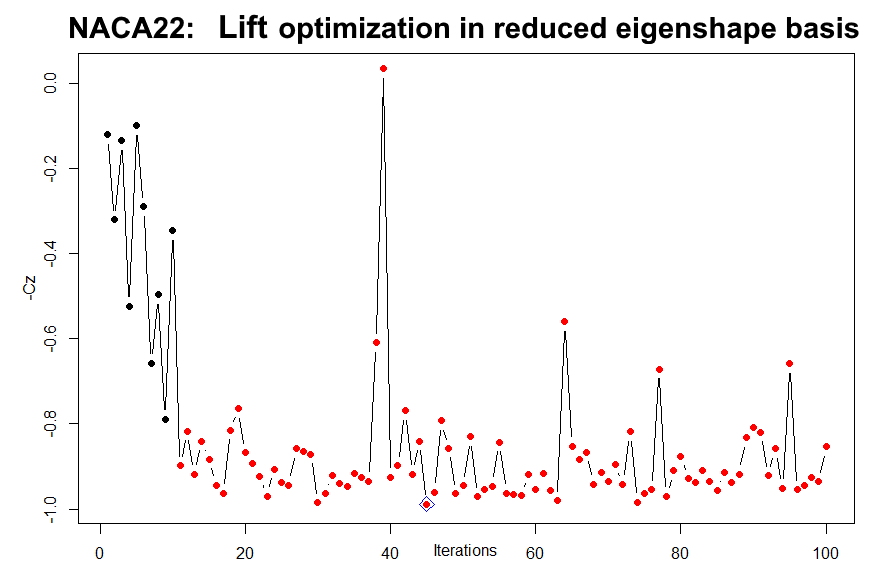}
	\includegraphics[width=0.33\textwidth]{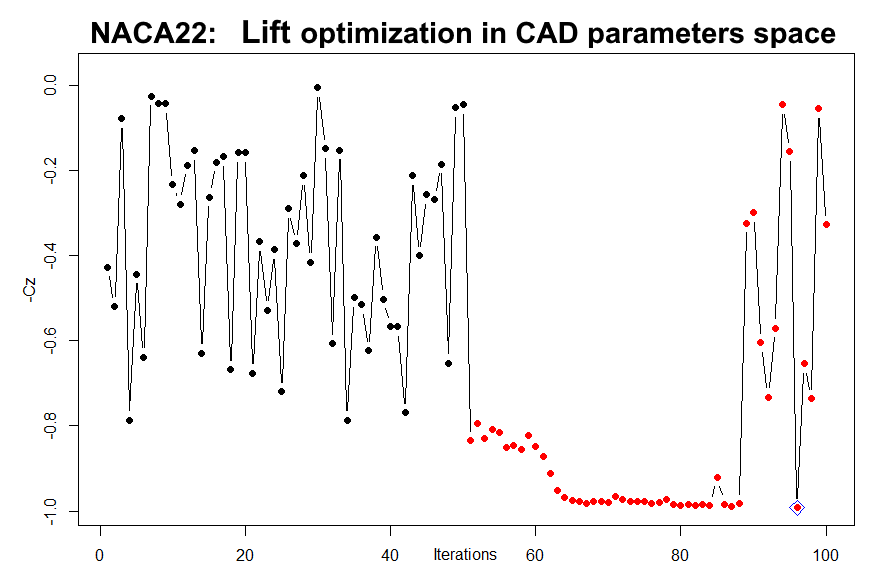}
	\includegraphics[width=0.32\textwidth]{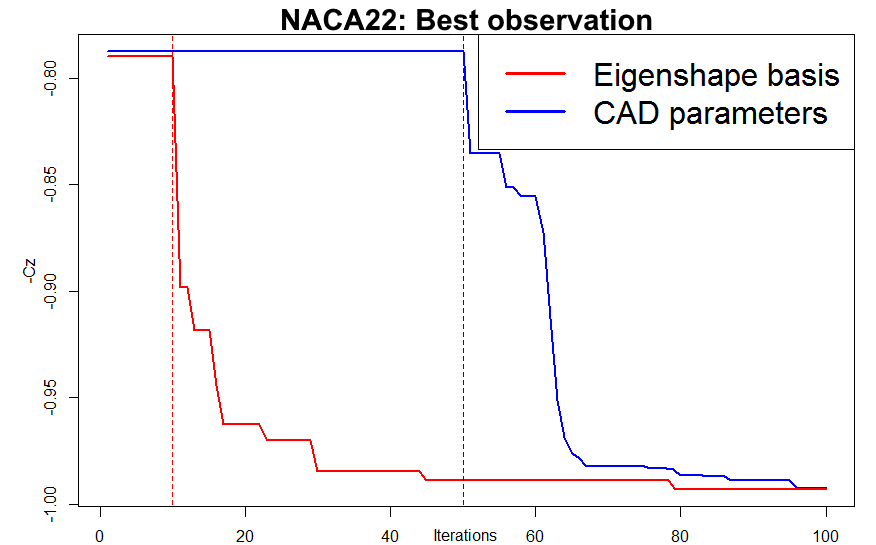}
	\caption{
		Top row: drag optimization of the NACA 22 airfoil in the reduced eigenbasis with \texttt{AddGP($\pmb\alpha^a+\pmb\alpha^{\overline{a}}$)-EI embed} (left) or carried out in the CAD parameters space with \texttt{GP($X$)-EI($X$)} (center).  Low drag airfoils are found with \texttt{AddGP($\pmb\alpha^a+\pmb\alpha^{\overline{a}}$)-EI embed} while the classical method still evaluates the airfoils of the initial design of experiments (right). Bottom row: lift optimization of the NACA 22 airfoil in the reduced eigenbasis with \texttt{AddGP($\pmb\alpha^a+\pmb\alpha^{\overline{a}}$)-EI embed} (left) or carried out in the CAD parameters space with \texttt{GP($X$)-EI($X$)} (center). High lift airfoils are found while the classical method still evaluates the airfoils of the initial design of experiments (right), i.e., lower objective functions are obtained faster.
	}
	\label{fig:optimization}
\end{figure}

\begin{figure}[h!]
	\centering
	Drag optimization\\
	\begin{subfigure}{0.35\textwidth}
		\centering
		\includegraphics[width=\textwidth]{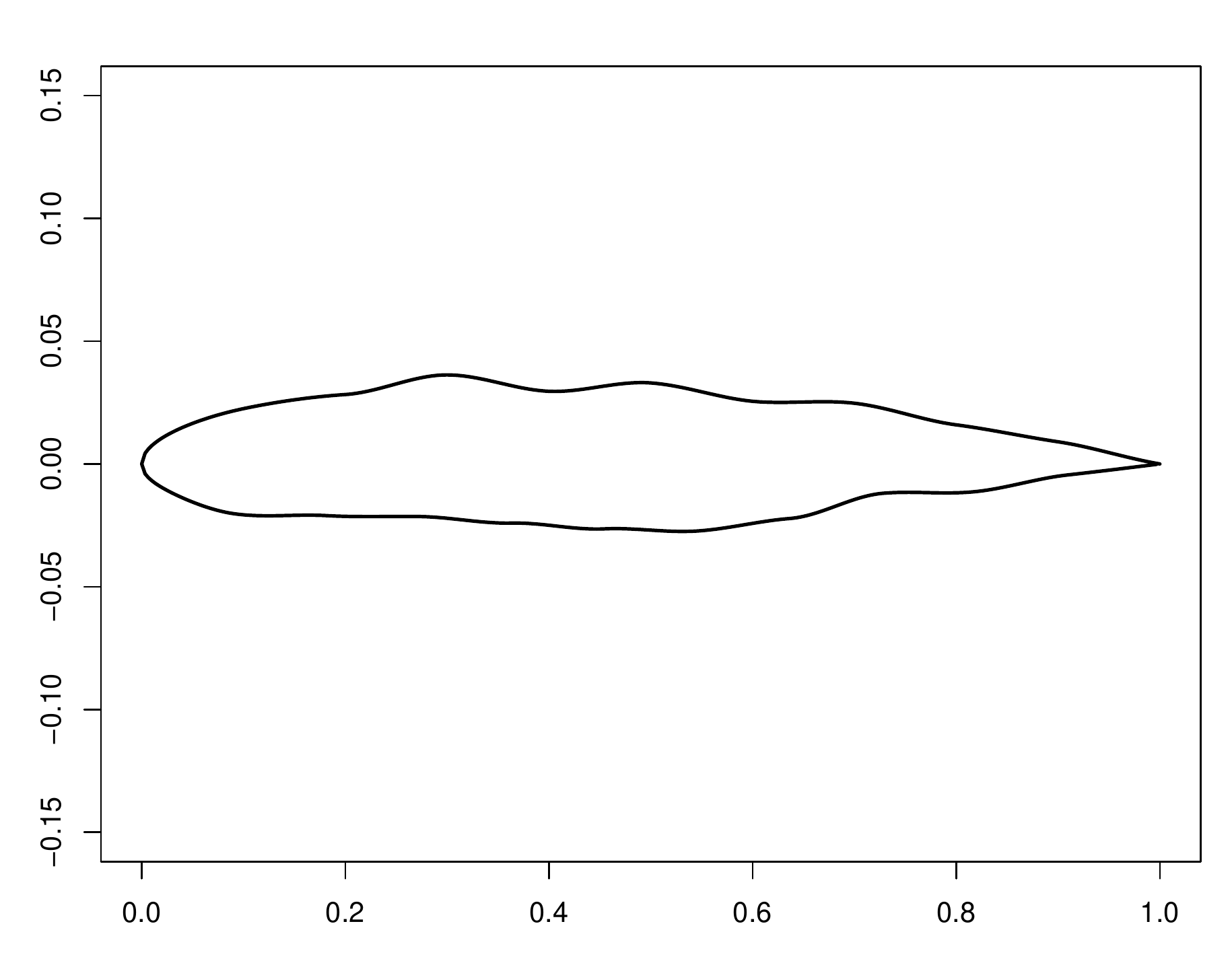}
		\caption*{\texttt{GP($X$)-EI($X$)}}
	\end{subfigure}
	\begin{subfigure}{0.35\textwidth}
		\centering
		\includegraphics[width=\textwidth]{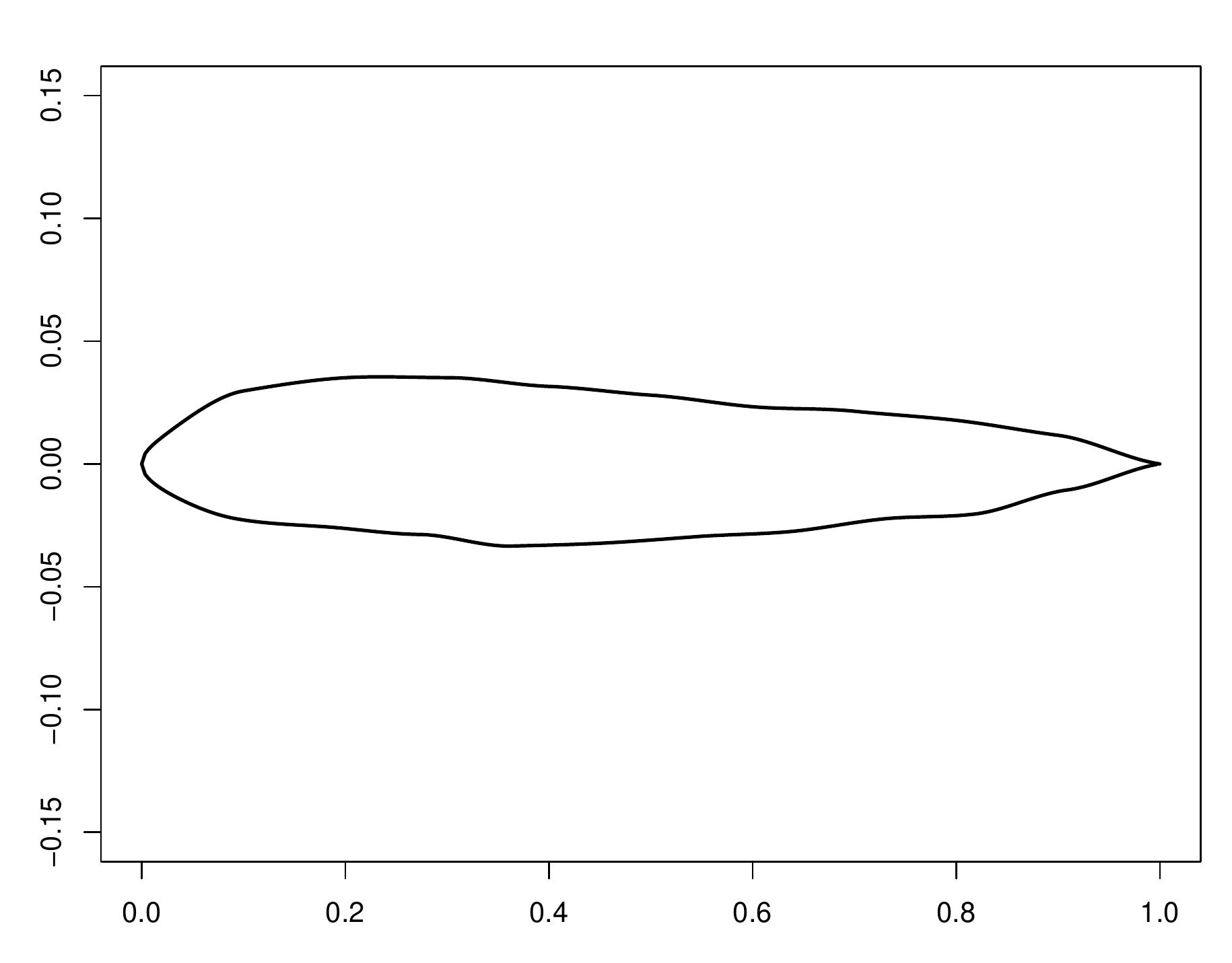}
		\caption*{\texttt{AddGP($\pmb\alpha^a+\pmb\alpha^{\overline{a}}$)-EI embed}}
	\end{subfigure}\\\vspace{5pt}
	Lift optimization\\
	\begin{subfigure}{0.35\textwidth}
		\centering
		\includegraphics[width=\textwidth]{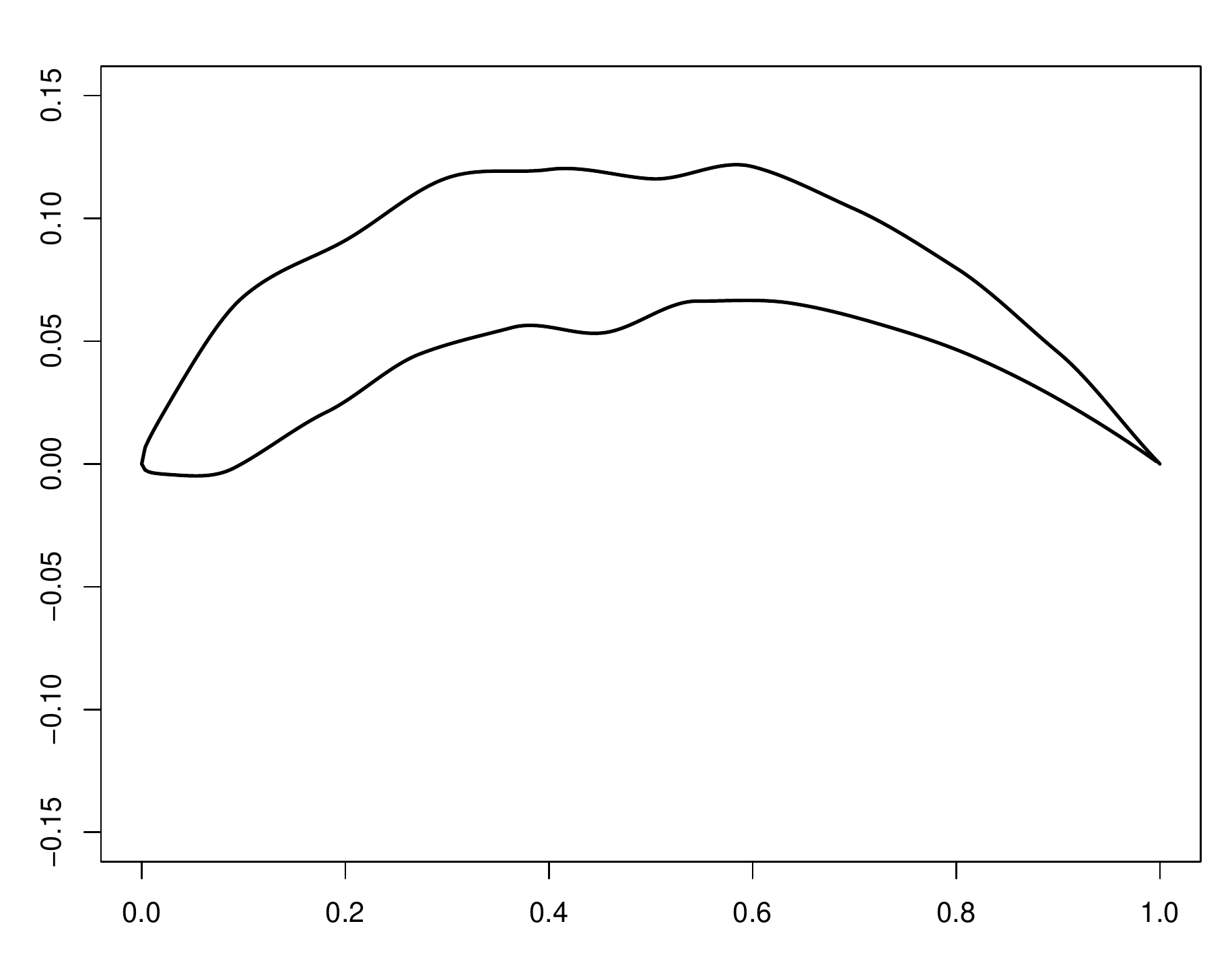}
		\caption*{\texttt{GP($X$)-EI($X$)}}
	\end{subfigure}
	\begin{subfigure}{0.35\textwidth}
		\centering
		\includegraphics[width=\textwidth]{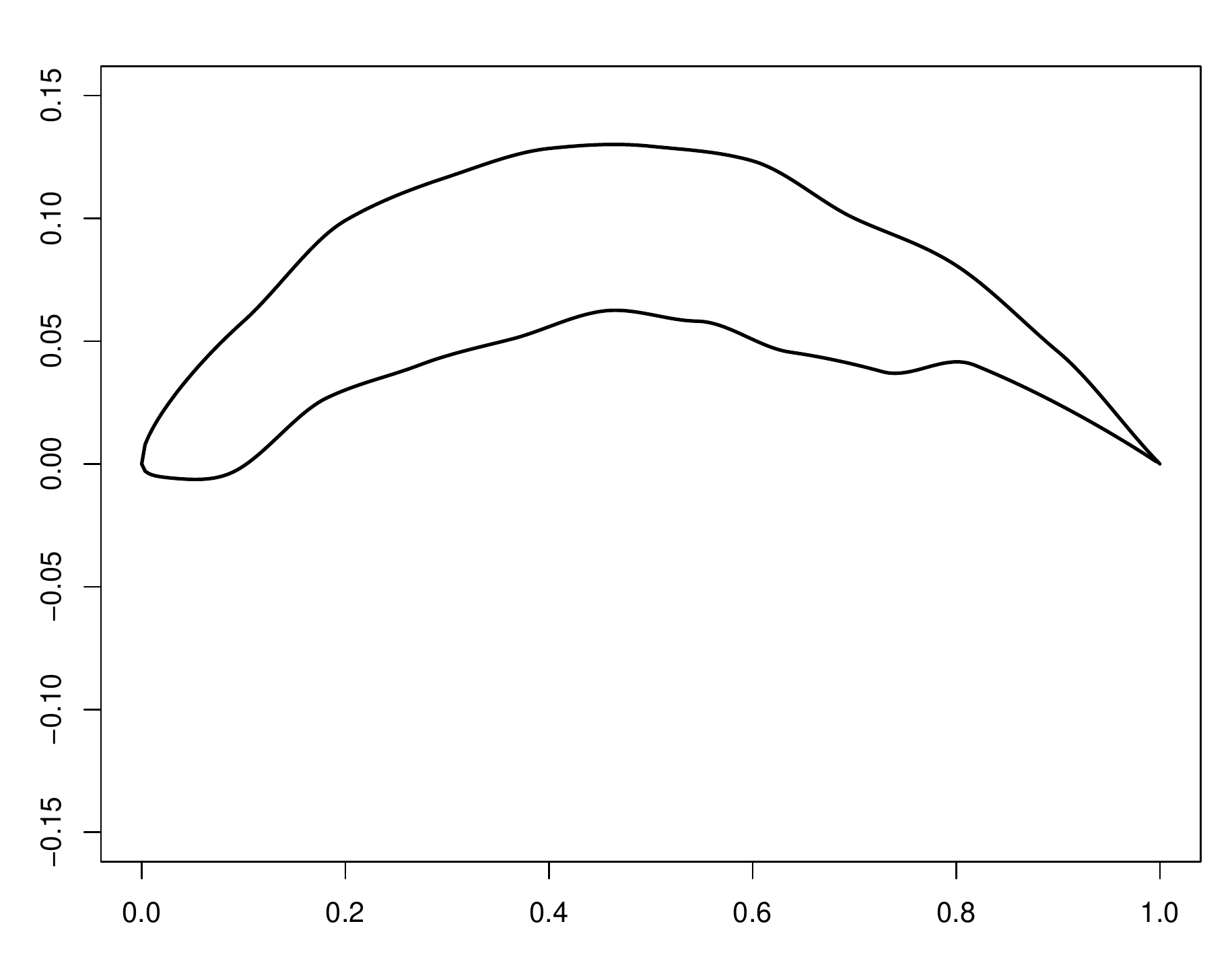}
		\caption*{\texttt{AddGP($\pmb\alpha^a+\pmb\alpha^{\overline{a}}$)-EI embed}}
	\end{subfigure}
	\caption{Airfoils found by the compared optimization algorithms. Top: drag minimization, bottom: lift maximization. Left: optimization with the \texttt{GP($X$)-EI($X$)} algorithm, right: optimization with the \texttt{AddGP($\pmb\alpha^a+\pmb\alpha^{\overline{a}}$)-EI embed} algorithm.}
	\label{fig:airfoils_optimization}
\end{figure}

Figure \ref{fig:optimization} shows the optimization runs of both algorithms for the minimization of the NACA 22's drag (top) and lift (bottom), and Figure \ref{fig:airfoils_optimization} the resulting airfoils.

In this application, the main advantage of the \texttt{AddGP($\pmb\alpha^a+\pmb\alpha^{\overline{a}}$)-EI embed} (Figure \ref{fig:optimization}, top left and bottom left) over the standard Bayesian optimizer (top center and bottom center) is that it enables an early search for low drag, respectively high lift airfoils, at a time when the standard approach is still computing its initial DoE. Indeed, the classical method needs much more function evaluations for building the initial surrogate model (black dots) because the inputs live in a space of higher dimension. 
The approach introduced in this paper would further gain in relevance in problems with more than $d=22$ CAD parameters, where it would almost be impossible to build a large enough initial design of experiments (whose size is typically of the order of $10 \times$ dimension \cite{loeppky2009choosing}).

It is observed in Figure \ref{fig:airfoils_optimization} that smoother airfoils are obtained with \texttt{AddGP($\pmb\alpha^a+\pmb\alpha^{\overline{a}}$)-EI embed} (right column), because it uses a shape coordinate system instead of treating the $L_i$'s (i.e., $x_i$'s with local influences on the airfoil, see Figure \ref{fig:naca22_description}) separately, as is done by \texttt{GP($X$)-EI($X$)} (left column). 
When the optimization aims at minimizing the drag, the \texttt{AddGP($\pmb\alpha^a+\pmb\alpha^{\overline{a}}$)-EI embed} airfoil (top right) is smoother than the \texttt{GP($X$)-EI($X$)} one (top left). And when the objective is to maximize the lift, the camber of the \texttt{AddGP($\pmb\alpha^a+\pmb\alpha^{\overline{a}}$)-EI embed} airfoil (bottom right) is increased in comparison with the design yielded by \texttt{GP($X$)-EI($X$)} (bottom left).

\section{Conclusions}
In this paper we have proposed a new methodology to apply Bayesian optimization techniques to parametric shapes and other problems where a pre-existing set of relevant points and a fast auxiliary mapping exist. 
Instead of working directly with the CAD parameters, which are too numerous for an efficient optimization and may not be the best representation of the underlying shape, we unveil the lower dimensional manifold of shapes through the auxiliary mapping and PCA. 
The dimensions of this manifold that contribute the most to the variation of the output are identified through an $L^1$ penalized likelihood and then used for building an additive Gaussian Process with a zonal anisotropy on the selected variables and isotropy on the other variables. This GP is then utilized for Bayesian optimization. 

The construction of the reduced space of variables opens the way to several strategies for the maximization of the acquisition criterion, in particular the restriction or not to the manifold and the replication.
The different variants for the construction of the surrogate model and for the EI maximization have been compared on 7 examples, 6 of them being analytical and easily reproducible, the last one being a realistic airfoil design.

Even though specific variants are more or less adapted to features of specific test problems, the supervised dimension reduction approach
and the construction of an additive GP between active and inactive components have given the most reliable results. 

Regarding the EI maximization our experiments highlight the efficiency of the random embedding in the space of inactive variables in addition to the detailed optimization of the active variables. It is a trade-off between optimizing the active variables only, and optimizing all variables.
Benefits have been observed for not restricting this inner maximization to the current approximation of $\mathcal A$ as well as for the virtual replication of points outside $\mathcal A$ when $\pmb\alpha\notin\mathcal A$ is promoted by the EI. 

Further research should consider shapes made of multiple elements such as the one in Example \ref{ex:3cercles}. 
This is of practical importance and it brings a new theoretical feature, the presence of symmetries in $\Phi$. The knowledge about symmetries has to be propagated to the eigenshape space to enhance the surrogate model.
\label{section:conclusions}
	
\section*{Acknowledgments}
This research was partly funded by a CIFRE grant (convention \#2016/0690) established between the ANRT and the Groupe PSA for the doctoral work of David Gaudrie.
	
\bibliographystyle{plain}
\bibliography{biblio}
\end{document}